\documentclass[10pt,twocolumn,letterpaper]{article}

\usepackage{3dv}
\usepackage{times}
\usepackage{epsfig}
\usepackage{graphicx}
\usepackage{amsmath}
\usepackage{amssymb}
\usepackage{cite}

\usepackage[pagebackref=false,breaklinks=true,letterpaper=true,colorlinks,bookmarks=false]{hyperref}

\newcommand*{\TS}[1]{}
\newcommand*{\GCS}[1]{}
\newcommand*{\MH}[1]{}

\usepackage{xspace}

\newcommand{\PAR}[1]{\vskip4pt \noindent{\bf #1.~}}

\threedvfinalcopy % *** Uncomment this line for the final submission

\def\threedvPaperID{178} % *** Enter the 3DV Paper ID here

\ifthreedvfinal\fi
\begin{document}

%%%%%%%%% TITLE
\title{Benchmarking Image Retrieval for Visual Localization}

% \author{Noé Pion\\
% NAVER LABS Europe\\
% %TODO \\
% {\tt\small noe.pion@gmail.com}
% \and
% Martin Humenberger\\
% NAVER LABS Europe\\
% %TODO\\
% {\tt\small martin.humenberger@naverlabs.com}
% \and
% Gabriela Csurka\\
% NAVER LABS Europe\\
% %TODO\\
% {\tt\small gabriela.csurka@naverlabs.com}
% \and
% Yohann Cabon \\
% NAVER LABS Europe\\
% {\tt\small yohann.cabon@naverlabs.com}
% \and
% Torsten Sattler\\
% Czech Technical University in Prague\\
% %TODO\\
% {\tt\small torsten.sattler@cvut.cz}
% }

\author{Noé Pion$^1$ \quad Martin Humenberger$^1$ \quad Gabriela Csurka$^1$ \quad Yohann Cabon$^1$ \quad Torsten Sattler$^2$\\
$^1$NAVER LABS Europe \quad $^2$Czech Technical University in Prague\\
%TODO \\
{\tt\small noe.pion@gmail.com, \{firstname.lastname\}@naverlabs.com, torsten.sattler@cvut.cz}}
% \and
% Martin Humenberger\\
% NAVER LABS Europe\\
% %TODO\\
% {\tt\small martin.humenberger@naverlabs.com}
% \and
% Gabriela Csurka\\
% NAVER LABS Europe\\
% %TODO\\
% {\tt\small gabriela.csurka@naverlabs.com}
% \and
% Yohann Cabon \\
% NAVER LABS Europe\\
% {\tt\small yohann.cabon@naverlabs.com}
% \and
% Torsten Sattler\\
% Czech Technical University in Prague\\
% %TODO\\
% {\tt\small torsten.sattler@cvut.cz}
% }

\maketitle
\thispagestyle{empty}

%%%%%%%%% ABSTRACT
\begin{abstract}
Visual localization, \ie, camera pose estimation in a known scene, is a core component of technologies such as autonomous driving and augmented reality. State-of-the-art localization approaches often rely on image retrieval techniques for one of two tasks: (1) provide an approximate pose estimate or (2) determine which parts of the scene are potentially visible in a given query image. It is common practice to use state-of-the-art image retrieval algorithms for these tasks. These algorithms are often trained for the goal of retrieving the same landmark under a large range of viewpoint changes. However, robustness to viewpoint changes is not necessarily desirable in the context of visual localization. This paper focuses on understanding the role of image retrieval for multiple visual localization tasks. We introduce a benchmark setup and compare state-of-the-art retrieval representations on multiple datasets. We show that retrieval performance on classical landmark retrieval/recognition tasks correlates only for some but not all tasks to localization performance. This indicates a need for retrieval approaches specifically designed for localization tasks. 
Our benchmark and evaluation protocols are available at \url{https://github.com/naver/kapture-localization}.

% Existing benchmarks already propose studying Image Retrieval performances given retrieval metrics. Yet, Image Retrieval is also an important component of modern Visual Localization pipelines, where the downstream task is to give accurate positions of query images. This position can be obtained through techniques using global or local reconstructions, or even structure free, and should require different types of retrieval. Many articles propose new visual localization pipelines without controlling for different retrieval methods. In this paper, we propose a benchmark of Image Retrieval on downstream tasks to study what makes for a good retrieval for visual localization.
% 
% We benchmark 4 different state-of-the-art global features on three datasets and show visual localization can be used to benchmark Image Retrieval with no/few human annotation. Moreover, we show that indoor image retrieval remains a challenging issue.
\end{abstract}

\section{Introduction}
\label{sec:introduction}

Visual localization is the problem of estimating the exact camera pose for a given image in a known scene, \ie, the exact position and orientation from which the image was taken. 
Localization algorithms are core components of systems such as % relying on visual position in a previously visited scene, \eg, 
self-driving cars~\cite{Heng2019ICRA}, autonomous robots~\cite{LimIJRR15RealTimeMonocularImageBased6DoFLocalization}, and mixed reality applications~\cite{ArthISMAR09WideAreaLocalizationMobilePhones,MiddelbergECCV14Scalable6DOFLocalization,Ventura2014TVCG,LynenRSSC15GetOutVisualInertialLocalization,Castle08ISWC}.

Traditionally, visual localization algorithms rely on a 3D scene representation of the target area~\cite{Se2002IROS,Li-ECCV-2012worldwide,IrscharaCVPR09FromSFMLocationRecognition,LiECCV10LocationRecPriorFeatureMatching,SattlerPAMI17EfficientPrioritizedMatching}, constructed from reference/database images with known poses. 
They use 2D-3D matches between a query image and the 3D representation for pose estimation. 
This representation can be an explicit 3D model, often obtained via Structure-from-Motion (SFM)~\cite{SchonbergerCVPR16StructureFromMotionRevisited,Snavely08IJCV,Heinly2015CVPR} using local features for 2D-3D matching, or an implicit representation through a machine learning algorithm~\cite{MassicetiICRA17RandomForestVersusNNCamLoc,BrachmannCVPR18LearningLessIsMore6DLocalization,ShottonCVPR13SceneCoordinateRegression}. In the latter case, the learning algorithm is trained to regress 2D-3D matches. % on loc model reconstructed from a set of database images.  
These structure-based methods can be scaled to large scenes through an intermediate image retrieval step~\cite{Taira2019TPAMI,Taira2018CVPR,SarlinCVPR19FromCoarsetoFineHierarchicalLocalization,SattlerBMVC12ImRetLocalizationRevisited,Sattler2015ICCV,Germain20193DV,Brachmann2019ICCVa,Cui3DVGraphMatch}. 
The intuition is that the top retrieved images provide hypotheses about which parts of the scene are likely visible in a query image. %image retrieval can be used to determine which part of the scene is likely visible in a query image. 
2D-3D matching can then be restricted to these parts. % regions. 
% In contrast, scene coordinate regression techniques~\cite{MassicetiICRA17RandomForestVersusNNCamLoc,BrachmannCVPR18LearningLessIsMore6DLocalization,ShottonCVPR13SceneCoordinateRegression} train machine learning algorithms such as random forests or CNNs to predict correspondences between 3D point coordinate and 2D pixel locations. While they achieve high pose accuracy on small datasets, yet they do not scale well to larger and more complex scenes~\cite{Taira2019TPAMI,Li2010ECCV,SattlerCVPR18Benchmarking6DoFOutdoorLoc,WeinzaepfelCVPR19VisualLocObjectsOfInterestDenseMatchRegression}.
% establish these 2D-3D correspondences using a scene-specifically trained deep neural network.
%Structure-based methods are able to scale to large scenes through an intermediate image retrieval  step which identifies visually similar images to match against~\cite{IrscharaCVPR09FromSFMLocationRecognition,Taira2019TPAMI,SarlinCVPR19FromCoarsetoFineHierarchicalLocalization,SattlerBMVC12ImRetLocalizationRevisited,Sattler2015ICCV,Germain20193DV}.

%Local matching against a part of the model instead of the full 3D map also improves performance by removing inconsistent local similarities using global context ~\cite{SarlinCVPR19FromCoarsetoFineHierarchicalLocalization,Taira2019TPAMI,Germain20193DV}, particularly present while matching \eg day against night images.

The pre-processing step of building a 3D scene representation is not strictly necessary. 
Instead, the camera pose of a query image can be computed using the known poses of the top database images found, again using image retrieval. 
This can be achieved via relative pose estimation between query and retrieved images~\cite{ZhouICRA20ToLearnLocalizationFromEssentialMatrices,ZhangS3DPVT06ImageBasedLocUrbanEnvironments}, by estimating the absolute pose from 2D-2D matches~\cite{Zheng2015ICCV}, via relative pose regression~\cite{BalntasECCV18RelocNetMetricLearningRelocalisation,DingICCV19CamNetRetrievalForReLocalization} or by building local 3D models on demand~\cite{Torii2019TPAMI}. 
If high pose accuracy is not required, the query pose can be approximated very efficiently %it is also possible to very efficiently obtain an approximation of the query pose %is required, it is also possible to e
%
% If the 6 degrees of freedom (DOF) camera poses of the database images are known, it is possible to approximate the query pose 
via a combination of the poses of the top retrieved database images~\cite{ToriiICCVWS11VisualLocalizationByLinearCombination,ZamirECCV10AccurateImageLocalization,Torii2019TPAMI}. 
% Even without using a global 3D map of the scene, more accurate pose estimates can be obtained from the results of the retrieval stage, \eg by estimating relative pose from matching and then deriving  the query pose by triangulation~\cite{ZhouICRA20ToLearnLocalizationFromEssentialMatrices,ZhangS3DPVT06ImageBasedLocUrbanEnvironments} by estimating the absolute pose from 2D-2D matches~\cite{Zheng2015ICCV}, via relative pose regression~\cite{BalntasECCV18RelocNetMetricLearningRelocalisation,DingICCV19CamNetRetrievalForReLocalization}, or by building local 3D models on demand~\cite{Torii2019TPAMI}. 
%
% Recent works on absolute camera pose regression~\cite{KendallICCV15PoseNetCameraRelocalization,KendallCVPR17GeometricLossCameraPoseRegression,WalchICCV17ImagebasedLocalizationUsingLSTMs,BrahmbhattCVPR18GeometryAwareLocalization} aim to eliminate the intermediate retrieval step by training CNNs to directly regress the pose of an input image. 
% However, such approaches are currently not (significantly) more accurate than simple retrieval baselines~\cite{SattlerCVPR19UnderstandingLimitationsPoseRegression} and significantly less scalable~\cite{SattlerCVPR18Benchmarking6DoFOutdoorLoc}. %Furthermore, \cite{SattlerCVPR19UnderstandingLimitationsPoseRegression} have shown that they are inherently closer related to image retrieval than to structure-based methods.

\begin{figure*}[t]
    \centering
    \includegraphics[width=0.8\textwidth]{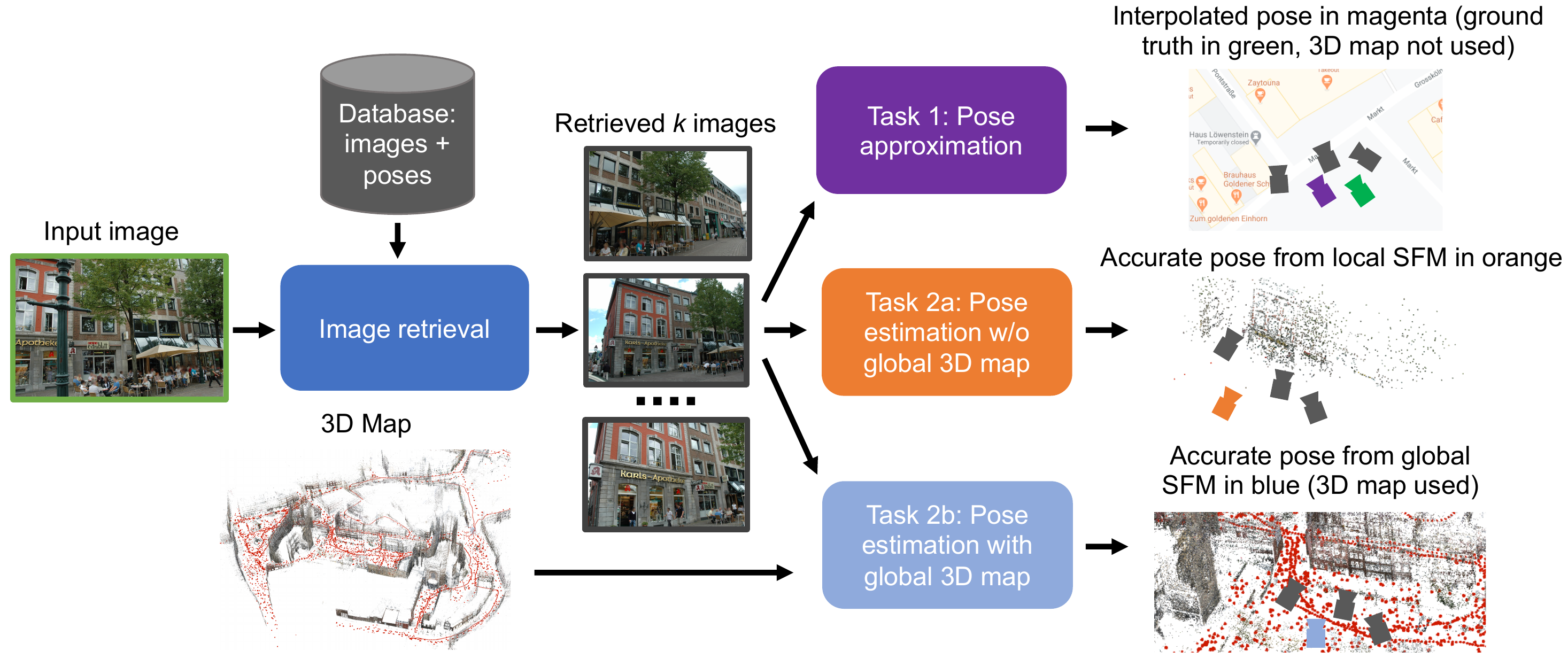}
    \caption{This paper analyzes the role of image retrieval in three visual localization tasks through extensive experiments.}
    \label{fig:overview}
\end{figure*}

As illustrated in Fig.~\ref{fig:overview}, %depending on the target task, 
there are various roles that image retrieval can play in visual localization systems.  
\textbf{Efficient pose approximation} by representing the pose of a query image by a (linear) combination of the poses of retrieved database images~\cite{ToriiICCVWS11VisualLocalizationByLinearCombination,ZamirECCV10AccurateImageLocalization,Torii2019TPAMI} \textbf{(Task 1)}. 
\textbf{Accurate pose estimation without a global 3D map} by computing the pose of the query image relative to the known poses of retrieved database images~\cite{ZhangS3DPVT06ImageBasedLocUrbanEnvironments,ZhouICRA20ToLearnLocalizationFromEssentialMatrices,LaskarICCVWS17CameraRelocalizationRelativePosesCNN,BalntasECCV18RelocNetMetricLearningRelocalisation,DingICCV19CamNetRetrievalForReLocalization,Torii2019TPAMI} \textbf{(Task 2a)}. 
\textbf{Accurate pose estimation with a global 3D map} by estimating 2D-3D matches between features in a query image and the 3D points visible in the retrieved images~\cite{IrscharaCVPR09FromSFMLocationRecognition,CaoCVPR13GraphBasedLocationRecognition,SattlerBMVC12ImRetLocalizationRevisited,SarlinCVPR19FromCoarsetoFineHierarchicalLocalization,Germain20193DV,Taira2019TPAMI} \textbf{(Task 2b)}.
%Given the current state of the art,

These three tasks have differing requirements on the results of the retrieval stage: 
{Task 1} requires the retrieval step to find images taken from poses as similar as possible to the query, \ie, the image representation should not be too robust or invariant to changes in viewpoint. 
Tasks 2a and 2b require the retrieval stage to find images depicting the same part of the scene as the query image. 
However, the retrieved images do not need to be taken from a similar pose as the query as long as local feature matching succeeds. 
In fact, Task 2a usually requires retrieving multiple images from a diverse set of viewpoints that differ from the query pose~\cite{LaskarICCVWS17CameraRelocalizationRelativePosesCNN,ZhouICRA20ToLearnLocalizationFromEssentialMatrices}.  
% Still, the poses of the retrieved images cannot be too different for subsequent which becomes harder for larger viewpoint changes. 
Task 2b benefits from retrieving images of high visual overlap with the query image and (in theory) requires only one relevant database image. % to be retrieved. % as to make the subsequent feature matching stage easier. 
% is interesting if a rough pose approximation is enough for the target application, while \textbf{Task 2b} estimates the most precise poses but comes with the burden of creating and maintaining a 3D reconstruction. The accuracy of \textbf{Task 2a} lies between \textbf{Task 1} and \textbf{Task 2b} but there is no 3D reconstruction needed.
% The three tasks place different requirements on the image representation used for image retrieval:
% \textbf{Task 1} requires the retrieval step to find images taken from poses as similar as possible to the query, \ie, the image representation should not be too robust or invariant to changes in viewpoint. 
% \textbf{Task 2a} (pose estimation without a global map) 
% Similarly, \textbf{Task 2b} benefits from retrieving images taken from similar viewpoints as to make the subsequent feature matching stage easier. 
% However, it is not necessary that the top retrieved image is the most similar pose to the query. 
% For both tasks, retrieving one relevant database image is theoretically sufficient. 
% In contrast, \textbf{Task 2a} requires retrieving multiple images from a diverse set of viewpoints~\cite{LaskarICCVWS17CameraRelocalizationRelativePosesCNN,ZhouICRA20ToLearnLocalizationFromEssentialMatrices}. 
% Still, the poses of the retrieved images cannot be too different as algorithms for \textbf{Task 2a} still depend on local features between images, which becomes harder for larger viewpoint changes. 

Despite differing requirements, modern %image retrieval-based 
localization methods~\cite{Torii2019TPAMI,Taira2019TPAMI,SarlinCVPR19FromCoarsetoFineHierarchicalLocalization,DusmanuCVPR19D2NetDeepLocalFeatures,Germain20193DV,ZhouICRA20ToLearnLocalizationFromEssentialMatrices} indiscriminately use the same representations based on compact image-level descriptors~\cite{ToriiTPAMI2018,ArandjelovicCVPR16NetVLADPlaceRecognition}. 
These descriptors are typically trained for landmark retrieval/place recognition tasks with the goal to produce similar descriptors for images showing the same building or place independently of the pose or other viewing conditions~\cite{LiuICCV19StochasticAttractionRepulsionEmbedding,RadenovicPAMI19FineTuningCNNImRet}. 
Interestingly, to the best of our knowledge, there is no work analyzing the suitability of such descriptors on the three visual localization tasks. % visual localization performance. 

In order to close this gap in the literature, this paper investigates the role of image retrieval for visual localization. 
We design a benchmark to measure the correlation between localization and retrieval/recognition performance for each task. 
Our benchmark enables a fair comparison of different retrieval approaches by fixing the remaining parts of the localization pipeline. %\footnote{In contrast, prior work on retrieval-based localization~\cite{IrscharaCVPR09FromSFMLocationRecognition,SattlerBMVC12ImRetLocalizationRevisited,SarlinCVPR19FromCoarsetoFineHierarchicalLocalization,Sattler2015ICCV} typically focuses on the pose estimation that follows the retrieval step.} % (Sec.~\ref{sec:framework:localization}); 
% (ii) enables us to correlate performances based on landmark retrieval / place recognition metrics with performance under visual localization metrics. 
Our main contributions are a set of extensive experiments and the conclusions we draw from them:
(\textbf{1}) there is no correlation between landmark retrieval/place recognition performance and Task 1. 
(\textbf{2}) Similarly, retrieval/recognition performance is not a good indicator for performance on Task 2a. 
(\textbf{3}) Task 2b correlates with the classical place recognition task. 
(\textbf{4}) Our results clearly show that there is a need to design image retrieval approaches specifically tailored to the requirements of some of the localization tasks.
To foster such research, our benchmark and evaluation protocols are publicly available. % at \url{https://github.com/naver/kapture-localization}. 

\section{Related Work}
\label{sec:related}

% The following section provides a concise review on work on three relevant topics : landmark retrieval (Sec \ref{sec:LMretrieval}), visual localization (Sec. \ref{sec:VisLoc}), and visual geo-localization
%  (Sec. \ref{sec:GeoLoc}). 
% Finally, in Sec \ref{sec:imret4Loc} we discuss previous work on using image retrieval in visual localization pipelines, which is the focus of this paper.

%\subsection{Landmark retrieval}
%\label{sec:LMretrieval}

%-------------------------------------------------------------------------

\PAR{Landmark retrieval}
%The goal of landmark retrieval is to retrieve images which contain an instance of a given object of interest such as a particular landmark. 
Landmark retrieval is the task of identifying all relevant database images depicting the same landmark as a query image. 
Early methods relying on global image statistics were significantly outperformed by %introduction of 
methods based on aggregating local features, most notably the bag of visual words representations for images~\cite{SivicICCV03VideoGoogle,CsurkaECCVWS04VisualCategorizationBagsKeypoints} and its extensions such as Fisher Vectors~\cite{PerronninCVPR07FisherKernels} and the Vector of Locally Aggregated Descriptors (VLAD)~\cite{JegouCVPR10AggregatingLocalDescriptors}. 
More recently, deep representation learning has led to further improvements. % landmark retrieval. 
They apply various {pooling mechanisms}~\cite{RazavianTMTA15VisualInstanceRetCNN,ToliasICLR16ParticularObjRetIntegralMaxpoolingCNN,BabenkoICCV15AggregatingDeepConvolutionalFeature,KalantidisECCVWS16CrossDimensionalWeightingAggregatedDeepFeatures,ToliasICLR16ParticularObjRetIntegralMaxpoolingCNN,RadenovicPAMI19FineTuningCNNImRet,ArandjelovicCVPR16NetVLADPlaceRecognition} on activations in the last {convolutional feature map} of CNNs in order to construct a global image descriptor. 
They learn the similarity metric by using ranking losses such as contrastive, triplet, or average precision (AP)~\cite{BabenkoECCV14NeuralCodesImRet,RadenovicPAMI19FineTuningCNNImRet,GordoIJCV17EndToEndDeepVisualReprImRet,RevaudICCV19LearningwithAPTrainingImgRetrievalListwiseLoss}. 
%
%
%\subsection{Landmark Retrieval Evaluation}
Several benchmark papers compare such image representations on the task of instance-level landmark retrieval~\cite{PhilbinCVPR08LostInQuantization,ArandjelovicCVPR13AllAboutVLAD,LongX16GoodPracticeCNNFeatureTransfer,NohICCV17LargeScaleAttentiveDeepLocalFeatures,RadenovicCVPR18RevisitingOxfordParisImRetBenchmarking,WeyandX20GoogleLandmarksDatasetv2BenchmarkInstanceRecRet}.
% Based on the results, we selected  AP-GeM~\cite{RevaudICCV19LearningwithAPTrainingImgRetrievalListwiseLoss} trained for landmark retrieval as one of the global image representation considered for our experiments.
In contrast, this paper explores how state-of-the-art landmark retrieval approaches perform in the context of visual localization.

%\subsection{Visual localization}
%\label{sec:VisLoc}

\PAR{Visual localization}
%Visual localization is the problem of estimating the 6DOF pose of a camera within a known 3D space representation using query images. 
%Visual localization has been seen an increasing research interest in the last decade as it enables autonomous vehicles to navigate in their surroundings and augmented reality applications to link virtual to real worlds. 
%Recent survey papers cover different aspects of visual localization~\cite{GarciaFidalgoRAS15VisionBasedTopologicalMapLocSurvey,LowryTROB16VisualPlaceRecognitionSurvey,ZamirB16LargeScaleVisualGeoLocalization,BrejchaPAA17StateSOAVisualGeoLocalization,PiascoPR18ASurveyVisualBasedLocHeterogeneousData} and benchmark these methods~\cite{Torii2019TPAMI,SattlerCVPR18Benchmarking6DoFOutdoorLoc,SattlerCVPR19UnderstandingLimitationsPoseRegression}.
%Here, we intend to summarize the literature of the field most relevant to our work.
%The following only briefly recalls the main tendencies. % and refer the interest reader to the literature. 
Traditionally, structure-based methods establish 2D-3D correspondences between a query image and a 3D map, typically via matching local feature descriptors~\cite{Schoenberger2017CVPR,CsurkaX18FromHandcraftedToDeepLocalFeatures} and use them
%such as SIFT~\cite{LoweIJCV04DistinctiveImageFeaturesScaleInvariantKeypoints} or its learned variants~\cite{RevaudNIPS19R2D2ReliableRepeatableDetectorsDescriptors,DusmanuCVPR19D2NetDeepLocalFeatures,Yang2020ARXIV,Ono2018NIPS}. These 2D-3D matches are then used 
to compute the camera pose by solving a perspective-n-point (PNP)  problem~\cite{KneipCVPR11ANovelParametrizationAbsoluteCamPose,Kukelova13ICCV,Larsson2017ICCV} robustly inside a RANSAC~\cite{Fischler81CACM,Chum08PAMI,Lebeda2012BMVC} loop.  
%
%Instead 
%of explicitly computing 2D-3D correspondences via feature matching,
More recently, scene coordinate regression techniques determine these correspondences using random forests~\cite{MassicetiICRA17RandomForestVersusNNCamLoc,ShottonCVPR13SceneCoordinateRegression} or CNNs~\cite{BrachmannCVPR18LearningLessIsMore6DLocalization,MassicetiICRA17RandomForestVersusNNCamLoc,Brachmann2019ICCVa}.
% achieving high pose accuracy at small scale while performing worse when scaled up to larger scenes~\cite{Taira2019TPAMI,BrachmannCVPR18LearningLessIsMore6DLocalization,WeinzaepfelCVPR19VisualLocObjectsOfInterestDenseMatchRegression}. 
Earlier methods trained a regressor specifically for each scene while recent models are able to adapt the trained model on-the-fly to new scenes~\cite{Cavallari2019TPAMI,Cavallari20193DV,YangICCV19SANetSceneAgnosticLocalization}. 
Even if scene coordinate regression methods achieve high pose accuracy on small datasets, they currently do not scale up well to larger and more complex scenes~\cite{Taira2019TPAMI,Taira2018CVPR,Li2010ECCV,SattlerCVPR18Benchmarking6DoFOutdoorLoc,WeinzaepfelCVPR19VisualLocObjectsOfInterestDenseMatchRegression,Brachmann2019ICCVa}. 
This is why we focus on feature-based localization methods that use image retrieval to cope with this problem~\cite{RevaudNIPS19R2D2ReliableRepeatableDetectorsDescriptors,RevaudARXIV2019,SarlinCVPR19FromCoarsetoFineHierarchicalLocalization,DusmanuCVPR19D2NetDeepLocalFeatures}. 

Absolute pose regression methods forego 2D-3D matching and train a CNN to directly predict the full camera pose from an image for a given scene~\cite{KendallICCV15PoseNetCameraRelocalization,KendallCVPR17GeometricLossCameraPoseRegression,WalchICCV17ImagebasedLocalizationUsingLSTMs,BrahmbhattCVPR18GeometryAwareLocalization}. 
However, they are significantly less accurate than structure-based methods~\cite{SattlerCVPR19UnderstandingLimitationsPoseRegression}  
% Recent works on absolute camera pose regression~\cite{KendallICCV15PoseNetCameraRelocalization,KendallCVPR17GeometricLossCameraPoseRegression,WalchICCV17ImagebasedLocalizationUsingLSTMs,BrahmbhattCVPR18GeometryAwareLocalization} aim to eliminate the intermediate retrieval step by training CNNs to directly regress the pose of an input image. 
and currently not (significantly) more accurate than simple retrieval baselines~\cite{SattlerCVPR19UnderstandingLimitationsPoseRegression} but significantly less scalable~\cite{SattlerCVPR18Benchmarking6DoFOutdoorLoc}. 
This is why we focus on image retrieval for efficient and scalable pose approximation instead. 
%While absolute pose regression algorithms require scene-dependent training, 
% In contrast, 
% relative camera pose regression methods are scene agnostic
% as they train a CNN to predict the pose of a test image relative to one or more training images found by image retrieval~\cite{LaskarICCVWS17CameraRelocalizationRelativePosesCNN,BalntasECCV18RelocNetMetricLearningRelocalisation}.
%or implicitly represented  in the ~\cite{SahaBMVC18ImprovedVisualRelocAnchorPoints}.
 %or with a coarse-to-fine retrieval-based deep learning framework where the retrieval is  pose-based refined.

Accurate real-world visual localization needs to be robust to a variety of conditions, including day-night,  weather and seasonal variations.
%, while providing highly accurate 6DOF camera pose estimates.
\cite{SattlerCVPR18Benchmarking6DoFOutdoorLoc} introduces several benchmark datasets specifically designed for analyzing the impact of such factors on visual localization using %carefully created ground truth poses for 
query and training images taken under varying conditions. For our benchmark, we use the Aachen Day-Night-v1.1~\cite{SattlerCVPR18Benchmarking6DoFOutdoorLoc,Zhang2020ARXIV}, the RobotCar Seasons~\cite{Maddern2017IJRR}, 
%These datasets were used in the long-term visual localization challenge\footnote{ \url{https://www.visuallocalization.net}}   allowing to compare several localization methods. We consider 
%for our benchmark  
and the Baidu shopping mall dataset~\cite{SunCVPR17DatasetBenchmarkingLocalization}.

For more details see recent survey papers that cover different aspects of visual localization~\cite{GarciaFidalgoRAS15VisionBasedTopologicalMapLocSurvey,LowryTROB16VisualPlaceRecognitionSurvey,ZamirB16LargeScaleVisualGeoLocalization,BrejchaPAA17StateSOAVisualGeoLocalization,PiascoPR18ASurveyVisualBasedLocHeterogeneousData} and benchmark these methods~\cite{Torii2019TPAMI,SattlerCVPR18Benchmarking6DoFOutdoorLoc,SattlerCVPR19UnderstandingLimitationsPoseRegression}.

%\subsection{Visual geo-localization}
%\label{sec:GeoLoc}

\PAR{Place recognition}
Place recognition, also referred to as visual geo-localization~\cite{ZamirB16LargeScaleVisualGeoLocalization}, lies between landmark retrieval and visual localization. 
While, similarly to the latter, its goal is to estimate the camera location, a coarse geographic position of the image is considered sufficient\cite{SchindlerCVPR07CityScaleLocationRecognition,HaysCVPR08IM2GPSGeographicInformation,ZamirECCV10AccurateImageLocalization,VoICCV17RevisitingIM2GPS}.
%  Even if off-the-shelf image retrieval methods are not the optimal solution for geo-localization, many place recognition methods heavily rely on them~\cite{SchindlerCVPR07CityScaleLocationRecognition,HaysCVPR08IM2GPSGeographicInformation,ZamirECCV10AccurateImageLocalization,VoICCV17RevisitingIM2GPS}. 
It is often important to explicitly handle confusing~\cite{SchindlerCVPR07CityScaleLocationRecognition,Knopp10ECCV} and repetitive scene elements~\cite{ToriiPAMI15VisualPlaceRecognRepetitiveStructures,SattlerCVPR16LargeScaleLocationRecognitionGeometricBurstiness,ArandjelovicACCV14DislocationDistinctivenessForLocation}, especially in large urban scenes. 
To improve scalability, a popular strategy is to perform visual and geo-clustering~\cite{CrandallWWW09MappingTheWorldsPhotos,LIICCV09LandmarkClassificationLargeScaleImageCollections,CaoCVPR13GraphBasedLocationRecognition,AvrithisACMMM10RetrievingLandmarkCommunityPhoto,KalantidisMTA11VIRaLVisualImgRetLocalization}.
%building scene graphs used for indexing and retrieval. 
%This strategy improves scalability compared to handling each image individually. 

As image matching and retrieval are key ingredients of place recognition methods, several papers proposed improved image representations using GPS and geometric information as a form of weak supervision~\cite{ArandjelovicCVPR16NetVLADPlaceRecognition,VoICCV17RevisitingIM2GPS,KimCVPR17LearnedContextualFeatureReweightingGeoLoc,RadenovicPAMI19FineTuningCNNImRet}.
%to define the set of places.  
In this paper, among others, we use NetVLAD~\cite{ArandjelovicCVPR16NetVLADPlaceRecognition} which is probably the most popular representation trained this way, as well as DenseVLAD~\cite{ToriiTPAMI2018}, its handcrafted counterpart.  

%We will compare them with representations designed for landmark retrieval considering various task including geo-localization  in a visual localization   Furthermore, considering our visual localization datasets, we we will design 

%\section{Image Retrieval for Visual Localization}
%\label{sec:imret4Loc}

\section{The proposed benchmark}
\label{sec:framework}

Modern localization algorithms tend to only use state-of-the-art landmark retrieval and place recognition representations. 
However, different localization tasks have different requirements on the retrieved images and thus on the used retrieval representations. 
In this paper, we are interested in understanding how landmark retrieval/place recognition performance relates to visual localization performance. 
In particular, we are interested in determining whether current state-of-the-art retrieval/recognition representations are sufficient or whether specialized (task-dependent) representations for localization are needed. 

This section presents an evaluation framework designed to answer this question. 
Our framework enables a fair comparison of different retrieval approaches for each of the three localization tasks by fixing the remaining parts of the localization pipeline. %\footnote{In contrast, prior work on retrieval-based localization~\cite{IrscharaCVPR09FromSFMLocationRecognition,SattlerBMVC12ImRetLocalizationRevisited,SarlinCVPR19FromCoarsetoFineHierarchicalLocalization,Sattler2015ICCV} typically focuses on the pose estimation that follows the retrieval step.}  
% , we design
It consists of two parts, one measuring localization performance for the three tasks identified above (Sec.~\ref{sec:framework:localization}) and the other measuring landmark retrieval/place recognition performance (Sec.~\ref{sec:framework:retrieval}), both on the same datasets. 
Relating the performance of state-of-the-art retrieval representations on all these tasks thus enables us to understand the relation between image retrieval in visual localization and landmark retrieval/recognition tasks.

\subsection{Visual localization tasks}
\label{sec:framework:localization}
As outlined in Sec.~\ref{sec:introduction}, we consider two roles for image retrieval in the context of visual localization: 
identifying reference images taken from a similar pose as the query image (Task 1) and retrieving database images depicting the same part of the scene as the query image but not necessarily from similar poses (Tasks 2a and 2b). 
%%%%%%%%
%%
%%%%%%%%
\PAR{Task 1: Pose approximation} 
%The task of estimating the pose of the query image via the known poses of the top $k$ retrieved database images is called pose approximation. 
Methods falling into the first category are inspired by place recognition~\cite{ZamirECCV10AccurateImageLocalization,Torii2019TPAMI,ToriiICCVWS11VisualLocalizationByLinearCombination,SattlerCVPR19UnderstandingLimitationsPoseRegression} 
and aim to efficiently approximate the query pose from the poses of the top $k$ retrieved database images.
%They take their inspiration from visual geo-localization techniques. 
%After retrieval, they 
%either directly inferring the pose of the top retrieved image~\cite{Torii2019TPAMI}
%or by weighted interpolation of several top retrieved image poses~\cite{ToriiICCVWS11VisualLocalizationByLinearCombination,SattlerCVPR19UnderstandingLimitationsPoseRegression}.
%While conceptually simple,  computationally efficiency with
%such  approaches are surprisingly competitive if there are limited viewpoint changes between the query and the database images such as in autonomous driving scenarios \cite{SattlerCVPR18Benchmarking6DoFOutdoorLoc}. 
%Further, their advantage is  computational efficiency  and low memory requirement as they only need storing a global descriptor and camera pose per database image. 
%The task of estimating the pose of the query image via the known poses of the top $k$ retrieved database images is called pose approximation. 
%Methods falling into this  category are inspired by visual geo-localization~\cite{ZamirECCV10AccurateImageLocalization,Torii2019TPAMI,ToriiICCVWS11VisualLocalizationByLinearCombination,SattlerCVPR19UnderstandingLimitationsPoseRegression} and aim to efficiently approximate the pose of the query image 
%via the known poses of the top $k$ retrieved database images.

We represent a camera pose as a tuple $\mathbf{P} = (\mathbf{c},\mathbf{q})$. 
Here, $\mathbf{c}\in \mathbb{R}^3$ is the position of the camera in the global coordinate system of the scene and $\mathbf{q}\in\mathbb{R}^4$ is the rotation of the camera encoded as a unit quaternion. 
% We follow the barycentric interpolation scheme from~\cite{ToriiICCVWS11VisualLocalizationByLinearCombination,SattlerCVPR19UnderstandingLimitationsPoseRegression} and 
We compute the pose of the query image as a weighted linear combination $\mathbf{P}_q = \sum_{i = 1}^k w_i \mathbf{P}_i$, 
%\begin{equation}
%    \mathbf{P}_q = \sum_{i = 1}^k w_i \mathbf{P}_i \enspace ,
%    \label{eq:PI}
%\end{equation}
where $\mathbf{P}_i$ is the pose of the top $i$ retrieved image and $w_i$ is a corresponding weight\footnote{Note that, $\mathbf{q}_q = \sum_i w_i \mathbf{q}_i$ 
%in general is not a unit quaternion. Thus, we 
is re-normalized to be a quaternion.}.
%$\mathbf{q}_q$ before further processing.}. 
As a consequence, for $k=1$ we directly use the pose of the top retrieved image.

We consider three variants: 
\textbf{equal weighted barycenter} (\textbf{EWB}) assigns the same weight to all of the top $k$ retrieved images with $w_i=1/k$. % the  $\mathbf{P}_q$  corresponds to the  \textbf{equal weighted barycenter} (EWB)
\textbf{Barycentric descriptor interpolation} (\textbf{BDI})~\cite{ToriiICCVWS11VisualLocalizationByLinearCombination,SattlerCVPR19UnderstandingLimitationsPoseRegression}  
%
%
%
%Alternatively, we follow %the barycentric interpolation scheme from
% ~\cite{ToriiICCVWS11VisualLocalizationByLinearCombination,SattlerCVPR19UnderstandingLimitationsPoseRegression} to estimate the weights $w_i$ by  finding the weights as best interpolation in the descriptor space
estimates $w_i$ as the best barycentric approximation of the query descriptor via the database descriptors with   
\begin{equation}
    \left\| \mathbf{d}_q - \sum_{i = 1}^k w_i \mathbf{d}_i \right\|_2 \enspace \text{subject to} \enspace \sum_{i=1}^k w_i = 1 \enspace.
    \label{eq:BDI}
\end{equation}
Here, $\mathbf{d}_q$ and $\mathbf{d}_i$ are the global image-level descriptors of the query image and the top $i$ retrieved database image, respectively. 

%When these weights are then used to interpolate the poses, we refer to the method by  \textbf{barycentric descriptor interpolation} (BDI).  
% Finally, we experiment with a
In the third approach, $w_i$ is based on the %powered 
\textbf{cosine similarity} (\textbf{CSI}) between L2 normalized descriptors:  % where
%, which in case of L2 normalized descriptors is equivalent with the dot product. 
%Hence, assuming that the descriptors $\mathbf{d}_i$ have unit norm\footnote{If not we  them.} 
% The weight 
%$w_i$ defined as
\begin{equation}
    w_i = \frac{1}{z_i} \left(\mathbf{d}_q^T \mathbf{d}_i\right)^\alpha  \enspace \text{, with} \enspace  z_i= \sum_{j = 1}^k \left(\mathbf{d}_q^T \mathbf{d}_j\right)^\alpha \enspace. 
    \label{eq:CalphaI}
\end{equation}
% The introduction of the $alpha$ parameter enables a smooth filtering of descriptors giving low similarities. 
Setting $\alpha = 0$ reduces this method to \textbf{EWB}. At the opposite, as $\alpha \to \infty$, the pose obtained is the one of the image giving the highest similarity.
We fix $\alpha=8$ based on preliminary results on the 
Cambridge Landmarks~\cite{KendallICCV15PoseNetCameraRelocalization} dataset.
%which is a smaller and   datasets such as 

%\PAR{Requirements on retrieval representations} Both BDI and CSI assume %approximation schemes assume %are based on the assumption that descriptor similarity reflects camera pose similarity. 
% In the limit case where all top $k$ retrieved images are relevant for the query and the descriptors are completely invariant to viewpoint changes, \ie, $\mathbf{d}_1 = \mathbf{d}_2 = \cdots = \mathbf{d}_k$, all three schemes will resort to averaging the database poses. Similarly, EWB benefits from ranking images with a pose similar to the query higher than images with less similar poses. 
%In contrast to the place / object detection tasks typically used to 
% Thus, \textbf{Task 1} benefits from image representations that are not invariant to viewpoint changes. 
% In contrast,
% commonly used 
%train global image descriptors~\cite{Arandjelovic2016CVPR,Radenovic2019PAMI,LiuICCV19StochasticAttractionRepulsionEmbedding}, where the objective is robustness to viewpoint variations, all three pose approximation schemes thus benefit from a limited robustness to viewpoint changes. 
% are trained such that images depicting the same place / object, even when taken from widely differing viewpoints, result in descriptors that are more similar to each other than to descriptors corresponding to unrelated images. 

%%%%%%%%
%%
%%%%%%%%
\PAR{Task 2a: Pose estimation without a global map} 
In theory, using the top 1 retrieved image would be sufficient for this task if the relative pose between the query and this image could be estimated accurately, including the scale of the translation~\cite{BalntasECCV18RelocNetMetricLearningRelocalisation,DingICCV19CamNetRetrievalForReLocalization}.
%While work in this direction exists~\cite{BalntasECCV18RelocNetMetricLearningRelocalisation,DingICCV19CamNetRetrievalForReLocalization}, it has only been evaluated on small scenes with very limited changes in viewpoint between query and database images. 
In practice, retrieving $k>1$ images improves the accuracy because \emph{k} relative poses can be considered~\cite{ZhangS3DPVT06ImageBasedLocUrbanEnvironments,ZhouICRA20ToLearnLocalizationFromEssentialMatrices,LaskarICCVWS17CameraRelocalizationRelativePosesCNN,Torii2019TPAMI}. 
%Retrieving $k>1$ images allows to improve accuracy 
%by estimating the query pose relative to the known poses of the database images~\cite{ZhangS3DPVT06ImageBasedLocUrbanEnvironments,ZhouICRA20ToLearnLocalizationFromEssentialMatrices,LaskarICCVWS17CameraRelocalizationRelativePosesCNN,Torii2019TPAMI}. 
Once the relative poses between query and database images are estimated, triangulation can be used to compute the absolute pose~\cite{ZhangS3DPVT06ImageBasedLocUrbanEnvironments,ZhouICRA20ToLearnLocalizationFromEssentialMatrices,LaskarICCVWS17CameraRelocalizationRelativePosesCNN}. 
However, pose triangulation fails if the query pose is colinear with the poses of the database images, which is often the case in autonomous driving scenarios.

Therefore, we follow~\cite{Torii2019TPAMI} where the retrieved database images with known poses are used to build the 3D map of the scene on-the-fly\footnote{Note that compared to the query pose, the 3D points are very seldomly colinear with the reference poses and can thus be accurately triangulated.} and then register the query image within this map using PNP. 
%~\cite{KneipCVPR11ANovelParametrizationAbsoluteCamPose,Kukelova13ICCV}. 
% This approach is based on \cite{Are Large-Scale 3D Models Really Necessary for Accurate Visual Localization?}. Instead of building and storing a big 3D Model offline, local models are built on line.
%For the top $k$ retrieved images, 
%where  performing local feature matching between all retrieved image pairs, we build a 3D structure is triangulation using the known global camera poses and intrinsic calibrations of the database images. 
%Local feature matching between  the query image and the  $k$ database photos yields a set of 2D-3D matches from which we can deduce the pose with 
 % between query features and 3D points in the local 3D map. 
%In turn, these 2D-3D correspondences are used for camera pose estimation. 
% For a query image, we retrieve various indexed images. Then, we match the local features of all possible pairs in the indexed images, and build a 3D model from the position of those images. The query image is then localized in this small scale model.
% As the 3D model needs to be reconstructed at testing time, this method can not localize an image from a single retrieved image.
Similar to pose triangulation, this \textbf{local SFM} approach fails %\footnote{But the query pose can still be approximated as in \textbf{Task 1} \cite{Torii2019TPAMI}.} 
if (i) less than two images among the top $k$ database images depict the same place as the query image, 
(ii) the viewpoint change among the retrieved images and/or between the query and the retrieved images is too large (to be handled by local feature matching) 
or (iii) the baseline between the retrieved database images is not large enough to allow stable triangulation of enough 3D points.
%\footnote{ This is also a failure case for methods based on pose triangulation~\cite{ZhouICRA20ToLearnLocalizationFromEssentialMatrices,LaskarICCVWS17CameraRelocalizationRelativePosesCNN}.}.
Thus, this approach requires retrieving a diverse set of images depicting the same scene as the query image from a variety of viewpoints. 
As such, methods for Task 2a benefit from image representations that are robust but not invariant to viewpoint changes.

% For our experiments, we use two types of local features, D2-Net~\cite{DusmanuCVPR19D2NetDeepLocalFeatures} and R2D2~\cite{RevaudNIPS19R2D2ReliableRepeatableDetectorsDescriptors}, to avoid overfitting to a single local image descriptors. 
% D2-Net and R2D2 were chosen as both robustly handle viewpoint and viewing conditions changes, \eg, between day and night or between seasons. 

% \TODO{Mention which features we are using, similar to task (3) below.}

%%%%%%%%
%%
%%%%%%%%
\PAR{Task 2b: Pose estimation with a global map} 
%Methods falling into the third category (\textbf{Task 2b})
 %relying on a global 3D map of the scene~\cite{IrscharaCVPR09FromSFMLocationRecognition,SattlerBMVC12ImRetLocalizationRevisited}
 %. Rather than using retrieved images to (approximately) determine where an image was taken, 
% use image retrieval to determine which part of a scene is visible in the query image in order to restrict the search range of 2D-3D matching.
%They take their inspiration from landmark retrieval. 
%between features in the query image and 3D points in the map  visible in the top ranked images. 
%In contrast to methods from the other categories, such approaches typically have higher memory and computational requirements due to the need of computing, storing and potentially maintaining the 3D map. 
% While on the one hand following a similar idea like \textbf{Task 2a}, t
In contrast to Task 2a, this task uses a pre-built global 3D model of the scene rather than reconstructing it locally on-the-fly.
We follow a standard local feature-based approach from the literature ~\cite{IrscharaCVPR09FromSFMLocationRecognition,SattlerBMVC12ImRetLocalizationRevisited,SarlinCVPR19FromCoarsetoFineHierarchicalLocalization,Germain20193DV}:  
%\ie, we match features between the query and the retrieved database images~
an SFM model of the scene provides the correspondences between local features in the database images and 3D points in the map. Establishing 2D-2D matches between the query image and top ranked database images yields a set of 2D-3D matches which are then used for pose estimation via PNP and RANSAC. 

% As for task (2), we exeriment with both D2-Net and R2D2 features. 
%To build our 3D model, we are Starting from the poses of the database images provided by each dataset, and  first we identify the top 20 other database image for each database photos.  For this task, the GeM-AP global descriptor~\cite{RevaudICCV19LearningwithAPTrainingImgRetrievalListwiseLoss} is used. 
%Afterwards, features are matched between the resulting image pairs and the matches are used to triangulate the 3D structure of the scene. 

In theory, retrieving a single relevant image among the top $k$  
is sufficient as long as the viewpoint change between the query and this image can be handled by the local features. 
%Naturally, r
Retrieving more relevant images  
%result in more correct 2D-3D matches and thus a higher i
increases the chance for accurate pose estimation. %to correctly and more accurately estimate the query pose. 
For efficiency, $k$ should be as small as possible\footnote{In scenes with ambiguous structures, \eg, the InLoc dataset~\cite{Taira2018CVPR}, retrieving more images can lead to a decrease in accuracy due to wrong, but geom. consistent, matches. We did not observe this in our experiments.} as local feature matching is often the bottleneck in terms of processing time.
% However, having more images might also introduce ambiguities, thus, the ideal\footnote{The optimal $k$ is query dependent.} $k$ is not too small but not too large either. 
%\NP{Interestingly, it should not always be the case that more retrieved images lead to better performances. Indeed, it is possible to incorporate local inconsistencies with more images (such as 2 images showing similar furniture), which will deteriorate the quality of the local matches, and thus of the localization.}
% \NP{Retrieving using the global context of an image, instead of a local pattern, is one of the advantages identified for hierarchical approaches in localization.}

Overall, we expect this task to benefit from retrieval representations that are moderately robust to viewpoint changes while still allowing reliable local feature matching. % estimation.

%%%%%%%%
%%
%%%%%%%%
\PAR{Visual localization metrics} We follow common practices~\cite{ShottonCVPR13SceneCoordinateRegression,SattlerCVPR18Benchmarking6DoFOutdoorLoc} to measure localization performance 
by computing the position and rotation errors between an estimated and a reference pose. 
For evaluation, we use the percentage of query images localized within a given pair of (position, rotation) error thresholds $(X\text{m}, Y^\circ)$~\cite{ShottonCVPR13SceneCoordinateRegression,SattlerCVPR18Benchmarking6DoFOutdoorLoc}.

\subsection{Landmark retrieval and place recognition tasks}
\label{sec:framework:retrieval} 
In order to correlate visual localization with landmark retrieval/place recognition performance, we evaluate the latter two tasks on the same datasets used for localization. 

\PAR{Landmark retrieval} This is an instance retrieval task where all images containing the main object of interest shown in the query image are to be retrieved from a large database of images.
%\footnote{A retrieved image is relevant even if  the landmark is seen from an opposite side or even interior.}
%Given a short-list of the top $k$ retrieved images, landmark retrieval systems are evaluated based on how many of these images are relevant to the query. 
Thus, image representations should be as robust as possible to viewpoint and viewing condition changes in order to identify all relevant images. 
%\NP{Images of two opposite sides of the same landmark would be considered as a relevant pair for this task, while not being useful for localization.}

In order to determine whether a retrieved image is relevant for a query, we follow the 3D model-based definition from~\cite{RadenovicPAMI19FineTuningCNNImRet}:  
the similarity of two images is computed as the intersection over union (IoU) for the sets of 3D scene points observed by both images in an SFM model. %that can be easily computed 
%{. \NP{this can be dleeted with what I added}}.
% Since the 3D models used in this work are based on SfM reconstructions, c
%Computing the IoU score is simple as this data is readily provided by the SfM process used to compute the 3D models~
%\cite{RadenovicPAMI19FineTuningCNNImRet}. 
We consider a database image relevant for a given query image if this IoU score is strictly positive, \ie they have shared 3D points.
%\footnote{Alternatively,  we can consider a database image as relevant if it was taken within a neighborhood of the query image (see Supplementary).}.
This is in contrast to classical landmark retrieval where relevant images might depict unrelated parts of the same landmark, \eg, opposite sides of the same building. % not necessarily needs to see the same part of the landmark.

% \TODO{to be integrated with the rest}
In order to compute the visible 3D points of the query images, we compute an SFM model containing the database and the query images using R2D2 features~\cite{RevaudNIPS19R2D2ReliableRepeatableDetectorsDescriptors} and COLMAP~\cite{SchonbergerCVPR16StructureFromMotionRevisited}. % Finally, to compute the image relevance with the IoU score
% We build an R2D2 SFM map using the query and the database images. 
To accelerate the image matching, we only match image pairs if their viewing frusta (limited by a far plane) overlap~\cite{BalntasECCV18RelocNetMetricLearningRelocalisation}. 
% This depends on the distance to the far plane of the frustum. 
The strategy can fail in two ways: the query images can either be too far away for the frusta to overlap or there are not enough good local feature correspondences in the resulting image pair. 
Hence some query images do not have reconstructed 3D points and are ignored during evaluation. 
%Note that this affected only a few images, except for Baidu Mall where the amount of affected images was about 7\%.

% Note that to avoid matching all possible pairs of images between the database and the query images we restrict to pairs that have overlapping viewing frusta, which depends on the distance to the considered far plane. We construct this 3D SFM reconstruction of the query and indexes by matching index by matching pairs of images only if they share camera frustum This thus has two potential points of failure. The images can be too far for the frustum to match, or if the local features correspondences are not good enough. In practice, on Aachen v1 less than 0.5\% query do not have associated ground-truth, while on Baidu it can be up to 5\% of the images due to strong blur on many queries. On RobotCar, this phenomenon is negligible.
 
%An alternative solution would be to define the similarity directly based on overlapping viewing frusta, but to take the actual scene content into account~\cite{BalntasECCV18RelocNetMetricLearningRelocalisation},   would require depth estimates. We did not experimented with this alternative as  depth maps were not available for the considered datasets.

%which were not available for the considered datasets.
%which we were not able to obtain.

\begin{figure*}[t!]
\begin{center}
 \includegraphics[width=0.2\textwidth]{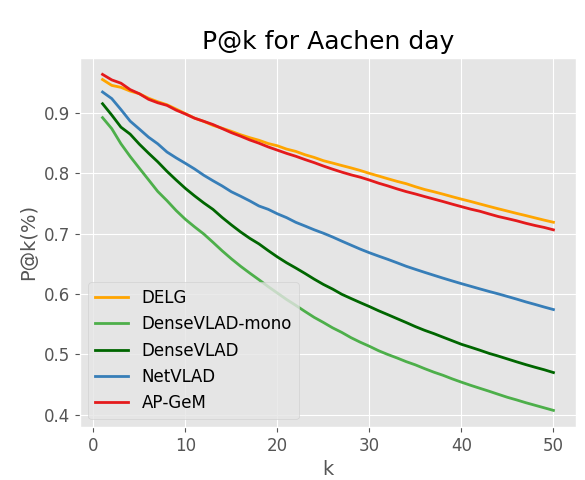}%
\includegraphics[width=0.2\textwidth]{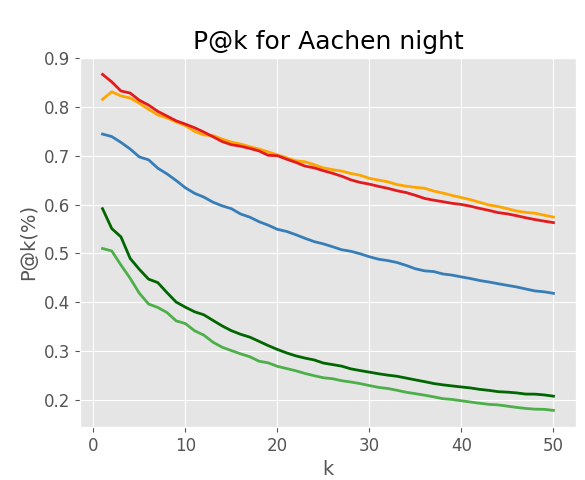}%
\includegraphics[width=0.2\textwidth]{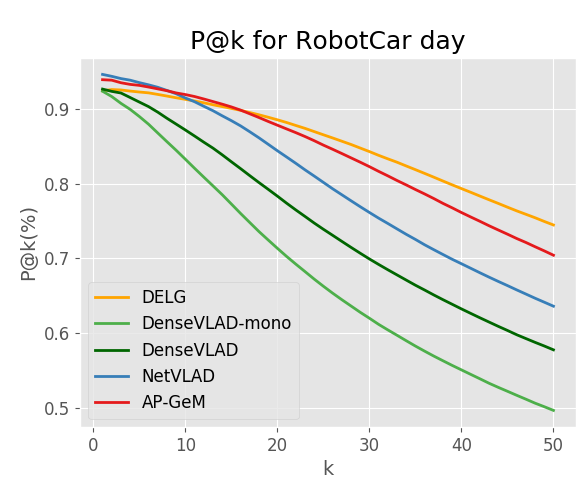}%
\includegraphics[width=0.2\textwidth]{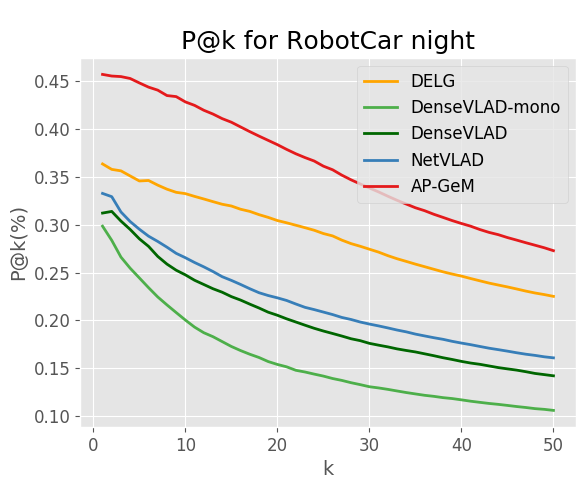}%
\includegraphics[width=0.2\textwidth]{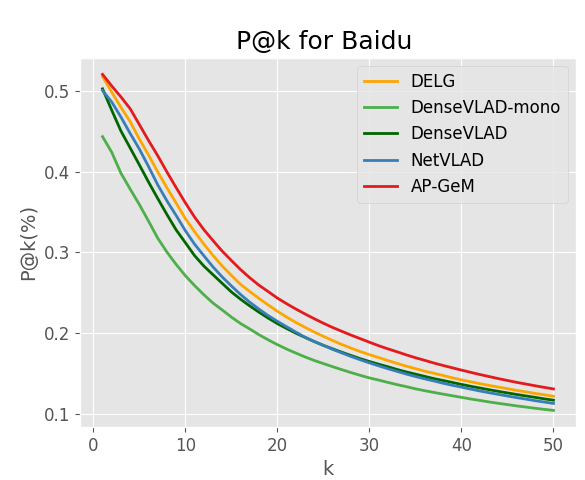}\\%
 \includegraphics[width=0.2\textwidth]{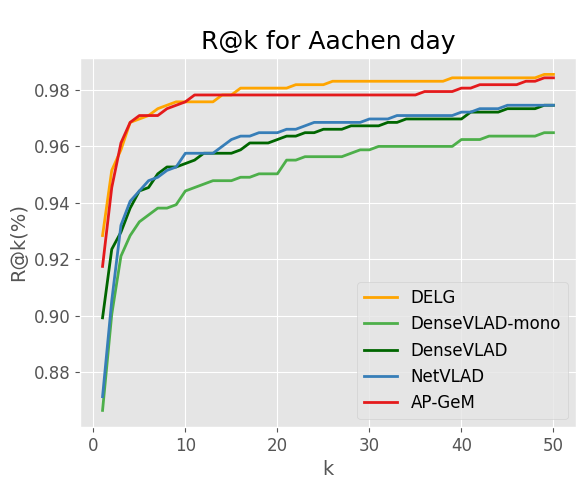}%
\includegraphics[width=0.2\textwidth]{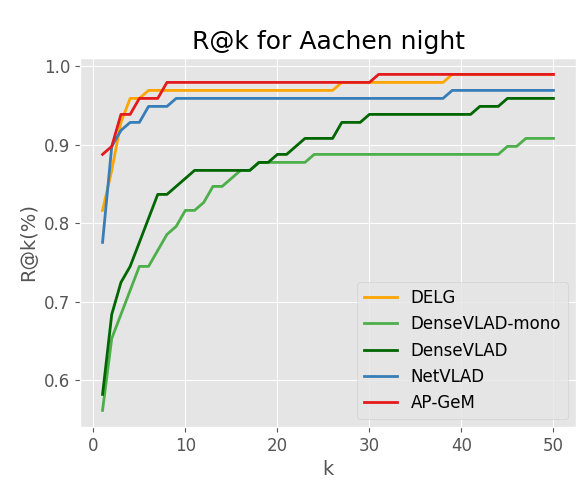}%
\includegraphics[width=0.2\textwidth]{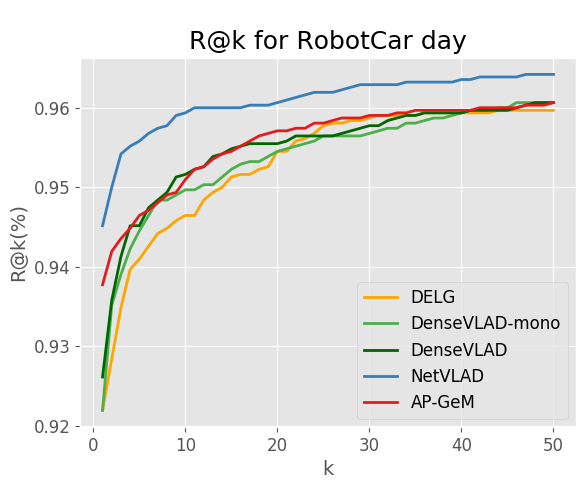}%
\includegraphics[width=0.2\textwidth]{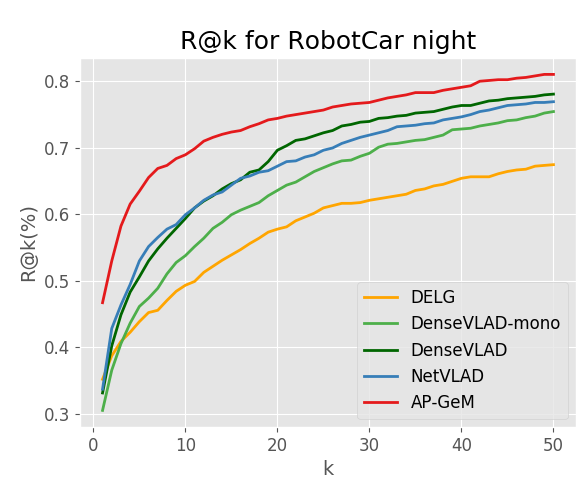}%
\includegraphics[width=0.2\textwidth]{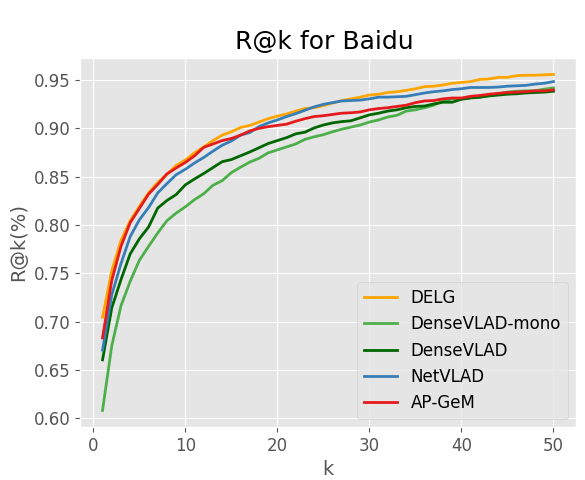}%
\end{center}
\vspace{-0.3cm}
   \caption{\textbf{Landmark retrieval/place recognition}. Image retrieval (top row) and place recognition (bottom row) performance as measured with Precision@$k$ respectively Recall@$k$. Results per dataset are shown per column. %night queries on Aachen, day and night queries on RobotCar, and Baidu. 
   There are clear differences between the representations with learned descriptors typically outperforming the handcrafted DenseVLAD descriptors. Best performance on both measures are obtained with AP-GeM and DELG, except on RobotCar night where DELG performs significantly worse than AP-GeM.}
   %\TODO{Noe, please add a y-label to the left-most figure and remove the legends in all but the top-left plot.}}%
\label{fig:retrieval}
\end{figure*}

 %The viewing frustum of a camera is defined as a truncated pyramid, where a near and far plane define the camera's visibility range~\cite{RTR4}. 
%Computing the viewing frustum requires an accurate estimate of the scene depth from the viewpoint of a camera. 
%Since SfM point clouds can have rather inaccurate point positions, we do not use this similarity criterion in this paper. \NP{We kind of do use this as we only match based on the fruta}
%Note that inaccurate point positions do not cause problems for the co-visibility-based similarity criterion. 

\PAR{Landmark retrieval metric} The classical mean Average Precision (mAP) metric, most commonly used in the literature~\cite{PhilbinCVPR07ObjectRetrievalFastSpatialMatching,PhilbinCVPR08LostInQuantization,ToliasPR14VisualQueryExpansionFeatureAggregation,RadenovicCVPR18RevisitingOxfordParisImRetBenchmarking} to measure landmark retrieval performance, reports a single number integrating over different numbers of retrieved images. 
We use the related \emph{mean Precision@$k$} ($\textrm{P@k}$) measure %and \emph{Recall} ($\textrm{R@k}$) 
to determine the link between number of retrieved images and localization performance.

\PAR{Place recognition} This task aims to approximately determine the place from which a given query image was taken. 
Since the place is defined by the location of the retrieved images, this requires at least one relevant reference image amongst the top $k$ retrieved ones. 
A database image is typically considered relevant if it was taken within a neighborhood of the query image~\cite{ToriiTPAMI2018,ToriiPAMI15VisualPlaceRecognRepetitiveStructures,ArandjelovicACCV14DislocationDistinctivenessForLocation,SattlerCVPR16LargeScaleLocationRecognitionGeometricBurstiness}. % (especially in geo-localization when only GPS coordinates of the image are known). 
When the camera orientation is available, the angle between the cameras' orientations can also be taken into account.
Alternatively, the above mentioned IoU similarity can be considered as well to determine whether or not a database image represents the same place~\cite{ChenCVPR11CityScaleLandmarkIdentification}. 
We consider both measures, discussing the latter in the main paper and showing results/correlations with the pose distance in the supplementary material.

%To this end, retrieval is used to identify database images taken from a similar pose. f
%\cite{ToriiCVPR15PlaceRecognitionByViewSynthesis} considers a query and database image to originate from the same place if the two images were taken within 25 meters of each other.
%Alternatively, we can follow  \cite{ToriiCVPR15PlaceRecognitionByViewSynthesis} where we consider a database image to be relevant to a given query if their difference in pose is within $X$ meters and optionally $Y$ degrees of each other 
%(cf. Eq.~\ref{eq:position_error} and Eq.~\ref{eq:orientation_error}),  thus taking camera orientation into account. 
%\NP{As, contrary to \cite{ToriiCVPR15PlaceRecognitionByViewSynthesis}, we have access to the full camera position and rotation, we also use the method introduced in the previous part.}
% we consider two

\PAR{Place recognition metric} We follow the standard protocol %from the literature 
and measure place recognition performance via 
%\footnote{While we use $\textrm{R@k}$ in both scenario, note that the relevance of a retrieved image is computed differently.}
\emph{Recall@$k$} (R@$k$)~\cite{ToriiTPAMI2018,ToriiTPAMI2018,ArandjelovicACCV14DislocationDistinctivenessForLocation,ArandjelovicCVPR16NetVLADPlaceRecognition}. 
%Recall@$k$ ($\textrm{R@k}$) is defined 
R@$k$ measures the percentage of query images with \textit{at least} one relevant database image amongst the top $k$ retrieved ones. 
%At first glance, it is not clear this metric correspond to any usecase. Given a query image of the Eiffel Tower, its is not the same retrieving 10 images of the Eiffel Tower, and confirming you are looking at it, or 1 of the Eiffel Tower and 9 of other monuments. This metric does not make a difference between those 2 scenarios. However, we will show that it correlates well with \textbf{Task 2} and \textbf{Task 3} performances. 

\section{Experimental evaluation}
\label{sec:experiments}

%In the followings first explains the experimental setup. %, \ie, the datasets and retrieval representations used. 
After first describing our experimental setup, Sec.~\ref{sec:experiments:retrieval} evaluates the representations on retrieval and place recognition tasks. 
Sec.~\ref{sec:experiments:pose_approximation} and \ref{sec:experiments:local_sfm} present results for pose approximation (Task~1) and accurate visual localization (Task~2). 
%Finally, we correlate the results and present our global findings. 

% In this section we first give some brief information about the benchmark setup, more details about datasets, global and local features, local and global SfM as well as the different evaluation metrics  can be found in the Supplementary Material. 

% Then in section \ref{sec:vislocexp} we first focus on visual localization experiments comparing the performances obtained with various global features in each of the tasks
%  detailed in Sec.~\ref{sec:framework}. 
%Next, we show %retrieval and place recognition results %some 
%comparative results between different tasks and analyse the correlation between them. %se tasks and visual localization. 
% Finally, in section \ref{placerecexp} we provide landmark retrieval and place recognition results on the same datasets and discuss the relation between these tasks and visual localization. 
%try to evaluate these datasets with some image retrieval metrics using different localization related ground truth relevance scores. 

\PAR{Experimental setup}
% In section~\ref{sec:related} we  mentioned the most popular tendencies to have global image representations for generic, and in particular, landmark retrieval. \NP{we explained that most global image representations are trained from landmark retrieval, which needs to be invariant to a wide variety of transformations}. Evaluating all of them in a single benchmark would be beyond the scope of this paper, therefore we selected some of the most representative approaches, which %also have shown 
We use 
% show (close to) state-of-the art performance on several datasets in the literature. 
% We consider one BoW-like representation,
\textbf{DenseVLAD}~\cite{ToriiTPAMI2018} and three popular deep image representations, \textbf{NetVLAD}~\cite{ArandjelovicCVPR16NetVLADPlaceRecognition},
\textbf{AP-GeM}~\cite{RevaudICCV19LearningwithAPTrainingImgRetrievalListwiseLoss}, and
\textbf{DELG}~\cite{CaoX20UnifyingDeepLocalGlobalFeatures}, for image retrieval. 
DenseVLAD pools densely extracted SIFT~\cite{Lowe04IJCV} descriptors through the VLAD representation~\cite{JegouCVPR10AggregatingLocalDescriptors}, resulting in a compact image-level descriptor. 
We use two variants: DenseVLAD extracts descriptors at multiple scales, while DenseVLAD-mono uses only a single scale. 
NetVLAD uses CNN features instead of SIFT features and was trained on the Pitts30k~\cite{ArandjelovicCVPR16NetVLADPlaceRecognition,ToriiTPAMI2018} dataset.
Both DenseVLAD and NetVLAD have been used for visual localization~\cite{SarlinCVPR19FromCoarsetoFineHierarchicalLocalization,Germain20193DV,Torii2019TPAMI,SattlerCVPR18Benchmarking6DoFOutdoorLoc,SattlerCVPR19UnderstandingLimitationsPoseRegression} and place recognition~\cite{ArandjelovicCVPR16NetVLADPlaceRecognition} before. 
AP-GeM and DELG represent state-of-the-art representations for landmark retrieval~\cite{RadenovicCVPR18RevisitingOxfordParisImRetBenchmarking}, while AP-GeM was recently used for visual localization as well~\cite{humenberger2020robust}. %and place recognition. 
Both models were trained on the Google Landmarks dataset (GLD)~\cite{NohICCV17LargeScaleAttentiveDeepLocalFeatures}, where each training image has a class label based on the landmark visible in the image. 
Relevance between images is established based on these labels. Hence, two images can be relevant to each other without showing the same part of a landmark. %\NP{Should we add qualitative examples ?}%  not seeing  shared part of the landmark. %On the contrary to train NetVLAD such pairs are considered negatives. 
% \TODO{Some more details on AP-GeM and DELG. G: Do we need more? I can add technical differences if needed especially if they can explain differences on the results.}. 
We use the best pre-trained models\footnote{It is common practice that localization methods rely on pre-trained models, \eg \cite{DusmanuCVPR19D2NetDeepLocalFeatures,Germain20193DV,Sattler2017PAMI} use off-the-shelf DenseVLAD or NetVLAD for their results and \cite{humenberger2020robust} uses AP-GeM.} released by the authors for all experiments (\cf suppl.~mat.). %. \NP{See details in supplementary.}
% Our choice for these features was motivated, on the one hand, by the fact that they have been successfully used for visual localization~\cite{SattlerBMVC12ImRetLocalizationRevisited,SarlinCVPR19FromCoarsetoFineHierarchicalLocalization}, geo-localization and landmark retrieval~\cite{ArandjelovicCVPR16NetVLADPlaceRecognition,RadenovicCVPR18RevisitingOxfordParisImRetBenchmarking} and, on the other hand, by the fact that the authors made their code and models publicly available. 

\begin{figure*}[t]
\begin{center}
 \includegraphics[width=0.2\textwidth]{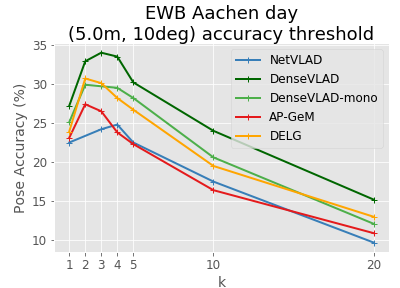}%
 \includegraphics[width=0.2\textwidth]{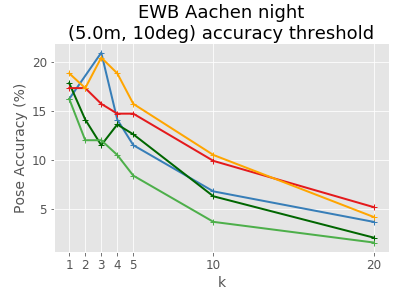}%
 \includegraphics[width=0.2\textwidth]{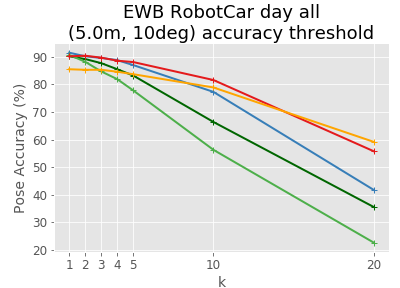}%
 \includegraphics[width=0.2\textwidth]{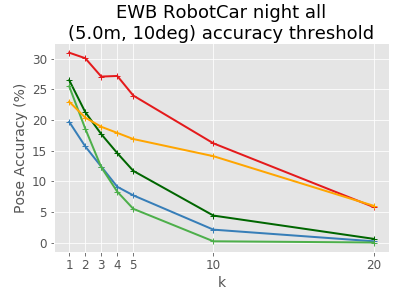}%
 \includegraphics[width=0.2\textwidth]{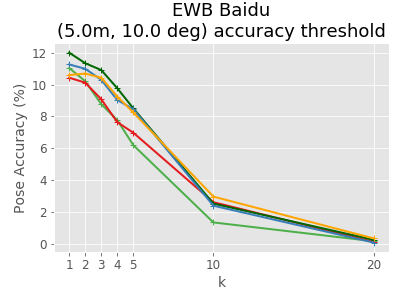}%
\\%
\includegraphics[width=0.2\textwidth]{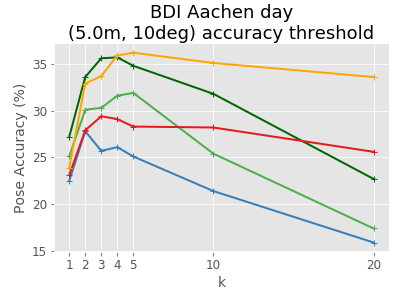}%
\includegraphics[width=0.2\textwidth]{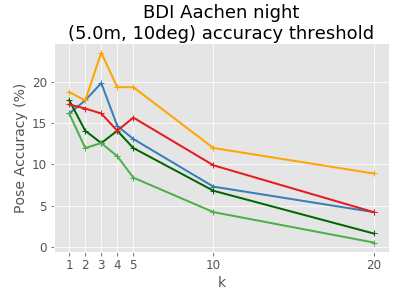}%
\includegraphics[width=0.2\textwidth]{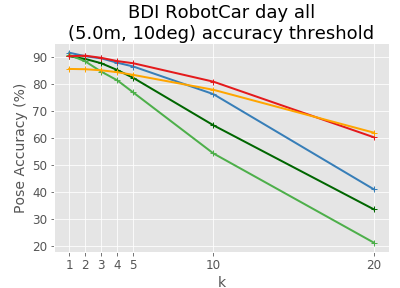}%
\includegraphics[width=0.2\textwidth]{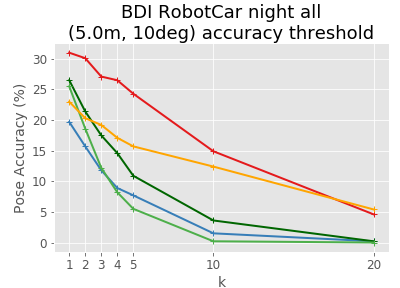}%
\includegraphics[width=0.2\textwidth]{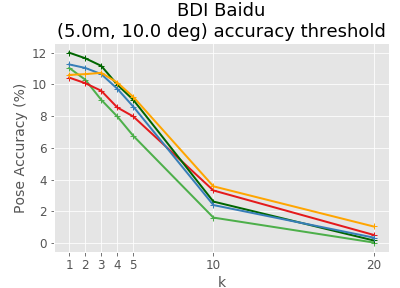}%
\\%
\includegraphics[width=0.2\textwidth]{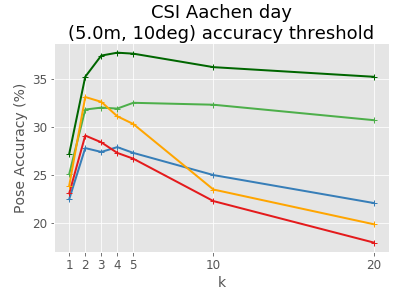}%
\includegraphics[width=0.2\textwidth]{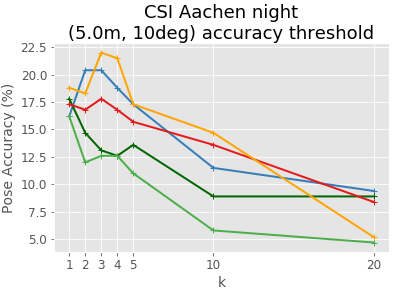}%
\includegraphics[width=0.2\textwidth]{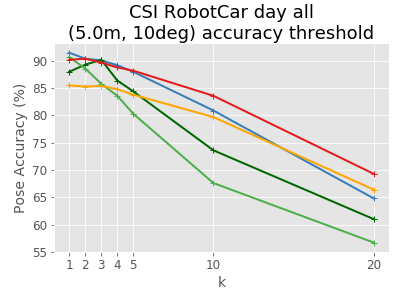}%
\includegraphics[width=0.2\textwidth]{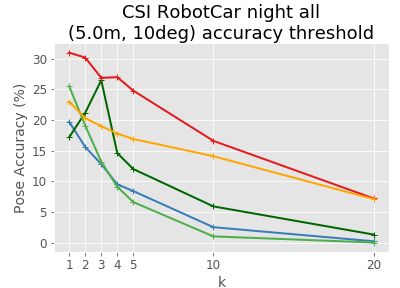}%
\includegraphics[width=0.2\textwidth]{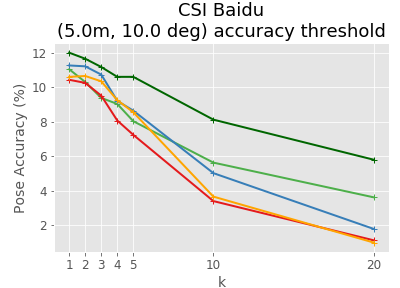}%
\end{center}
\vspace{-0.3cm}
   \caption{\textbf{Task 1 (pose approximation)}. The rows show results obtained via equal weighted barycenter (EWB), barycentric descriptor interpolation (BDI), and cosine similarity (CSI), respectively. %Results per dataset are shown per column. 
   %day and night queries on Aachen, day and night queries on RobotCar, and Baidu.  
   The best results are obtained with CSI, however simply using the top-retrieved pose works best for RobotCar and Baidu due to their limited variation between query and reference poses. 
   Interestingly, the results on Baidu and the daytime queries for Aachen and RobotCar show that the rather low-level DenseVLAD descriptors perform as good or better than the learned descriptors. }
   %DenseVLAD is less robust to viewpoint changes, and thus more likely to retrieve images taken from a similar pose. If illumination changes are present, learned descriptors perform better due to their higher robustness to such changes.}
   %\TODO{Noe, please add a y-label to the left-most figure and remove the legends in all but the top-left plot. Try to use larger fonts and thicker lines. Also, please make sure that all plots have the same width. For Aachen night, please use the same y-range for all three plots. Remove from the title the low accuracy threshold we can add in the text and we will gain space. }}%
\label{fig:pose_approximation}
\end{figure*}

For Tasks 2a and 2b, \ie, pose estimation without and with a global map, we use R2D2~\cite{RevaudNIPS19R2D2ReliableRepeatableDetectorsDescriptors} to extract local image features and COLMAP~\cite{SchonbergerCVPR16StructureFromMotionRevisited} for SFM. We observed similar behavior for D2-Net~\cite{DusmanuCVPR19D2NetDeepLocalFeatures} %\NP{,R2D2~\cite{RevaudNIPS19R2D2ReliableRepeatableDetectorsDescriptors}} 
and SIFT~\cite{Lowe04IJCV} (\cf suppl.~mat.). %\NP{The only diverging behavior is obtain for SIFT on RobotCar where performances decease at $k=50$}  %on which we observe similar behaviour can be found in the supp. material. 

We use three public datasets for our experiments:  
\textbf{Aachen Day-Night-v1.1}~\cite{SattlerCVPR18Benchmarking6DoFOutdoorLoc,SattlerBMVC12ImRetLocalizationRevisited,Zhang2020ARXIV}, \textbf{RobotCar Seasons} ~\cite{Maddern2017IJRR,SattlerCVPR18Benchmarking6DoFOutdoorLoc}, and \textbf{Baidu Mall}~\cite{SunCVPR17DatasetBenchmarkingLocalization}. 
RobotCar represents an autonomous driving scenario with little viewpoint change between query and reference images. 
In contrast, the outdoor dataset Aachen exhibits stronger viewpoint changes as the camera can freely move through the scene. 
Both datasets come with \textbf{day-} and \textbf{nighttime} queries. 
Nighttime queries introduce the challenge of handling strong illumination changes as all reference images were taken during the day.
The Baidu dataset contains medium viewpoint and limited illumination changes between the query and database images but exhibits strong differences in image quality, occlusion from people, and other distractions such as reflections on storefronts.

Following \cite{SattlerCVPR18Benchmarking6DoFOutdoorLoc}, for all three datasets we use three threshold pairs for evaluating localization for \textbf{low} (5m, 10$^\circ$), \textbf{medium} (0.5m, 5$^\circ$), and \textbf{high} (0.25m, 2$^\circ$) accuracy.
We only show here results with thresholds \emph{low} and \emph{high}  as these are the most relevant for pose approximation and accurate pose estimation (for \emph{medium}, see suppl.~mat.). % but 

\subsection{Landmark retrieval and place recognition}
\label{sec:experiments:retrieval}
Figure~\ref{fig:retrieval} shows the results for landmark retrieval (top) and place recognition (bottom) tasks from Sec.~\ref{sec:framework:retrieval}. 
As expected, the learned descriptors (NetVLAD, AP-GeM, and DELG) typically outperform the SIFT-based DenseVLAD. 
There are two interesting observations: 
(1) NetVLAD outperforms both AP-GeM and DELG under the R@$k$ measure for small $k$ 
%of top ranked images 
on the daytime RobotCar queries. 
This can be attributed to the fact that NetVLAD was trained on street-view images captured at daytime from a vehicle while AP-GeM and DELG were trained with a large variety of landmark images taken from very different viewpoints.
%\TODO{Verify that we are really using the Pittsburgh dataset. Yes I used the  Pittsburg model. )}
(2) On R@$k$ for the RobotCar nighttime queries, DELG performs significantly worse than the others. %dataset under the Recall@$k$. 
We attribute this to the low-quality nighttime images, which often exhibit strong motion blur and color artifacts which are not reflected in the training set of DELG. 
AP-GeM, trained on the same data, avoids this problem through adequate data augmentation. 
%\TODO{Does this explanation make sense?}

\PAR{Discussion} Our experiments confirm that the state-of-the-art 
DELG and AP-GeM descriptors remain the best choices for place retrieval/recognition tasks. 
To complement this, Table~\ref{tab:RoxfordRparis} shows the performance obtained on $\mathcal{R}$Oxford ($\mathcal{R}$O) and $\mathcal{R}$Paris ($\mathcal{R}$P) landmark retrieval benchmarks using the Medium (m) and Hard (h) protocols \cite{RadenovicCVPR18RevisitingOxfordParisImRetBenchmarking}. 
Here, DELG and AP-GeM descriptors also outperform DenseVLAD and NetVLAD by a large margin. 
However, if the aim is place recognition, \ie, if the requirement is to find at least one relevant image, 
the gap is significantly lower and in some cases the ranking might even change (\eg.,~the Robotcar dataset). 

%There is no large difference between visual localization and standard place retrieval/recognition datasets for these tasks. 

% \begin{table}[hhh]
\begin{table}[t]
 \begin{center}
{\footnotesize{
%  \vspace{-0.3cm}
\resizebox{0.9\linewidth}{!}{
\begin{tabular}{||l|c|c|c|c||}
\hline
 & $\mathcal{R}$O(m) &  $\mathcal{R}$O (h) & $\mathcal{R}$P(m) & $\mathcal{R}$P(h)  \\
\hline
%VLAD (rSIFT) & 33.9 & 13.2 & 43.6 & 17.5 \\
%DenseVLAD & 36.8 & 13.0 & 42.53 & 13.68  \\
DenseVLAD & 36.8 & 13.0 & 42.5 & 13.7  \\
NetVLAD & 37.1 & 13.8  & 59.8 & 35.0  \\
AP-GeM & 67.4 & 42.8 & 80.4 & 61.0 \\
DELG & 69.7 & 45.1 & 81.6 & 63.4 \\
\hline
\end{tabular}
}
}}
\end{center}
 \caption{Performance evaluation (mAP) on $\mathcal{R}$Oxford ($\mathcal{R}$O) and $\mathcal{R}$Paris ($\mathcal{R}$P) using the Medium (m) and Hard (h) protocols. }
 \label{tab:RoxfordRparis}
%\vspace{-0.3cm}
\end{table}

\begin{figure*}[t]
\begin{center}
% ##
\includegraphics[width=0.2\textwidth]{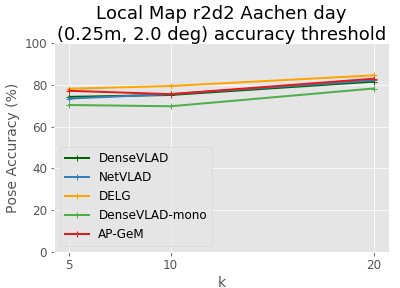}%
\includegraphics[width=0.2\textwidth]{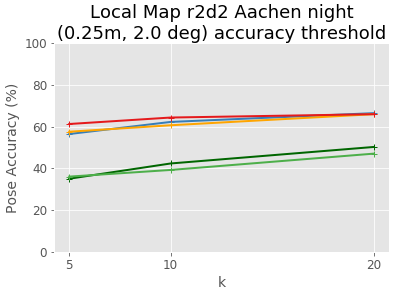}%
\includegraphics[width=0.2\textwidth]{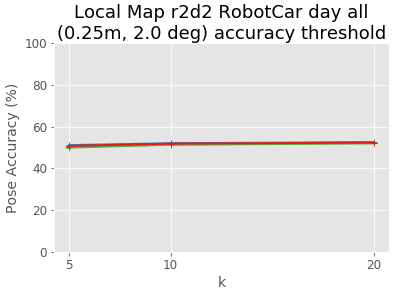}%
\includegraphics[width=0.2\textwidth]{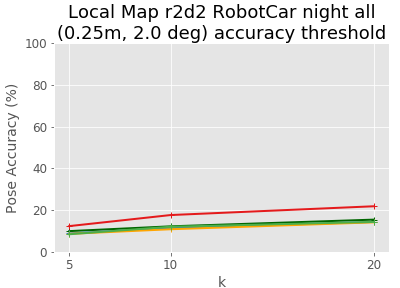}%
\includegraphics[width=0.2\textwidth]{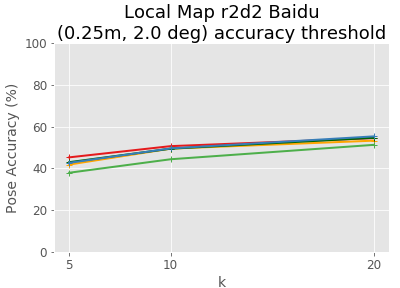}%
\\
\includegraphics[width=0.2\textwidth]{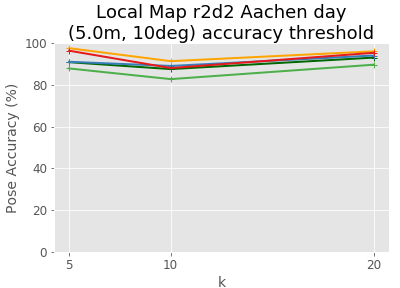}%
\includegraphics[width=0.2\textwidth]{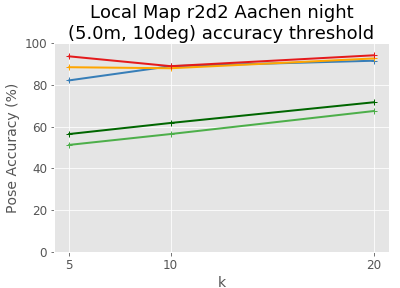}%
\includegraphics[width=0.2\textwidth]{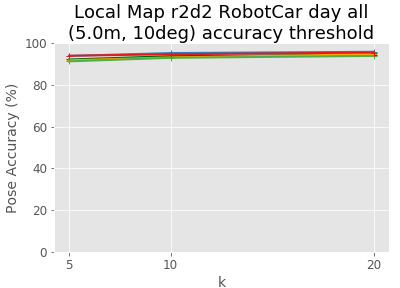}%
\includegraphics[width=0.2\textwidth]{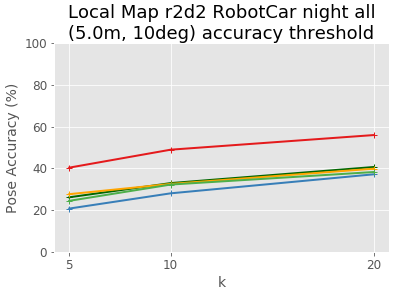}%
\includegraphics[width=0.2\textwidth]{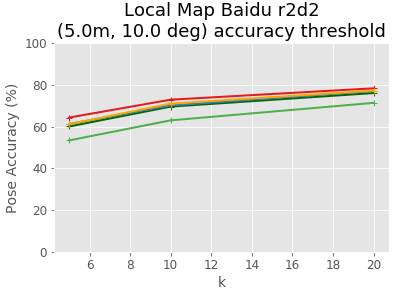}%
%\\
%\includegraphics[width=0.2\textwidth]{plots/retrieval vs pose/local/aachen/P@k_vs_pose_local_map_Aachen Day-Night v1.1_day_r2d2.png}%
%\includegraphics[width=0.2\textwidth]{plots/retrieval vs pose/local/aachen/P@k_vs_pose_local_map_Aachen Day-Night v1.1_night_r2d2.png}%
%\includegraphics[width=0.2\textwidth]{plots/retrieval vs pose/local/robotcar/P@k_vs_pose_local_map_RobotCar-Seasons_day_r2d2.png}%
%\includegraphics[width=0.2\textwidth]{plots/retrieval vs pose/local/robotcar/P@k_vs_pose_local_map_RobotCar-Seasons_night_r2d2.png}%
%\includegraphics[width=0.2\textwidth]{plots/retrieval vs pose/local/baidu/P@k_vs_pose_local_map_Baidu__r2d2.png}%
\end{center}
\vspace{-0.3cm}
   \caption{\textbf{Task 2a (pose estimation without a global map)}.
   %/with a local map) results}.  
   Top/bottom row: percentage of images localized within the high/low accuracy
   %strictest/loosest pose 
   threshold as a function of the number $k$ of retrieved images. 
%   Results per dataset are shown per column. 
   All representations perform similarly on outdoor day images. AP-GeM slightly outperforms the other descriptors on night images as well as on Baidu, while DenseVLAD performs worst in these cases.} 
   %Comparing the results to Fig.~\ref{fig:retrieval}, we do not observe a clear correlation between landmark retrieval/recognition and Task 2a.}% 
%   Bottom row: correlation between localization results for the loosest threshold and P$@k$ (for $k=5,10,20). We do not observe a clear correlation between Task 2a and the landmark retrieval task.}% these sets.}
   %day and night queries on Aachen, day and night queries on RobotCar, and Baidu. 
\label{fig:local_SFM}
\end{figure*} 

\subsection{Task 1: Pose approximation}
\label{sec:experiments:pose_approximation}
Figure~\ref{fig:pose_approximation} shows pose approximation results for the three approaches discussed in Sec.~\ref{sec:framework:localization}. % on the three datasets.  
We show the percentage of query images localized within a given error threshold \wrt the ground truth poses as a function of the number $k$ of retrieved images used for pose approximation. 
We only report results for the low (5m, 10$^\circ$) thresholds as even fewer images are localized for stricter thresholds (\cf suppl.~mat.). 
Equal weighted barycenter (EWB) uses the same weight for each of the top $k$ retrieved images. 
In contrast, both barycentric descriptor interpolation (BDI) and cosine similarity (CSI) give a higher weight to higher ranked images. 
They thus assume correlation between 
% 
% \TODO{The text below I moved here from section 4, and suggest that we rather integrate it with the discussions in this section.}
%  Both BDI and CSI assume %approximation schemes assume %are based on the assumption 
descriptor and pose similarity. % and pose similarity are correlated. 
As can be seen in Fig.~\ref{fig:pose_approximation}, retrieving $k>1$ images only improves performance on the Aachen dataset because 
there is a larger pose difference between query and reference images than for the other datasets. 
This allows better poses to be estimated by interpolating between the poses of the retrieved database images. 
Here, CSI performs best as the exponent $\alpha$ in Eq.~\ref{eq:CalphaI} effectively downweigths unrelated images, whereas unrelated images among the top $k$ receive a larger weight for BDI and EWB.

Comparing Fig.~\ref{fig:retrieval}~and~\ref{fig:pose_approximation}, we observe a correlation between these curves and P@$k$ but none with R@$k$, except for Aachen day\footnote{This is confirmed by the scatter plots in the supplementary material.}. In the latter case, we observe that while DenseVLAD performs poorly on the landmark retrieval/place recognition tasks it offers good performance for pose approximation. The reason might be that 
%In contrast to AP-GeM and DELG, the top performing methods on the retrieval/recognition tasks, 
DenseVLAD descriptors are less robust to viewpoint changes.
As discussed in Sec.~\ref{sec:framework:localization}, this is desirable for pose approximation as more similar poses will lead to more similar descriptors. 
However, DenseVLAD is less robust to illumination changes than AP-GeM and DELG. 
This explains why the latter perform better on the nighttime queries of RobotCar and Aachen.

Finally, while AP-GeM performs similar for retrieval/recognition on Aachen (see Fig.~\ref{fig:retrieval}), interpolating top retrieved poses with DELG outperforms AP-GeM. 
% for which considering more than the top image degraded the performance. 
This suggests that DELG retrieves reference images that are better spread through the scene, which is beneficial for pose interpolation. %while the number of relevant retrieved images is similar the ones retrieved by DELG  are better spreader around the query pose. 

\PAR{Discussion} Overall, our results show that while pose approximation metrics is somewhat 
 correlated to the top precision of
image retrieval,  there is no correlation with the place recognition task (R@$k$).
These suggest that learning scene representations tailored to pose approximation, instead of using off-the-shelf methods trained for landmark retrieval, is an interesting direction for future work.  

The best results for RobotCar and Baidu are obtained for $k=1$, in which case all three methods perform the same. 
For RobotCar, this result is surprising as interpolating between two consecutive images should give finer approximation. 
This indicates that there is potential for improvement by %using 
% The fact that this is not the case suggests that the descriptors are not discriminative enough to either properly interpolate or to identify the two relevant reference images. This further suggests that 
designing image retrieval representations specifically suited for pose interpolation. %suitable for pose approximation is an interesting research direction.

\begin{figure*}[t]
\begin{center}
% ##
\includegraphics[width=0.2\textwidth]{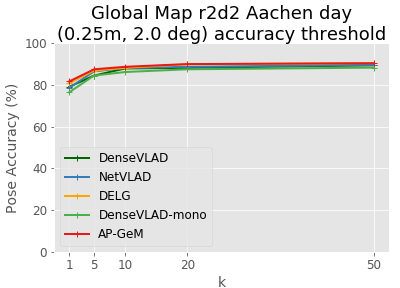}%
\includegraphics[width=0.2\textwidth]{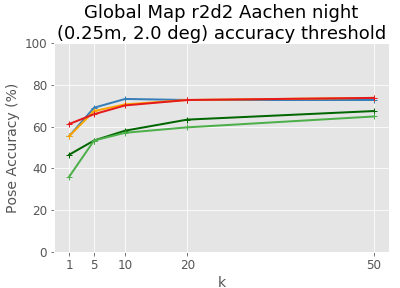}%
\includegraphics[width=0.2\textwidth]{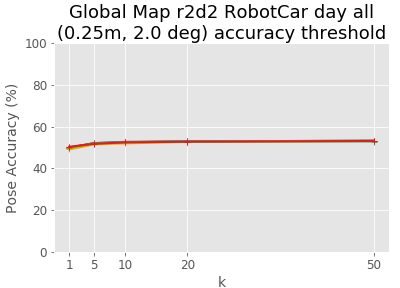}%
\includegraphics[width=0.2\textwidth]{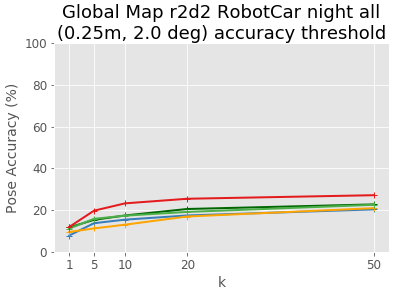}%
\includegraphics[width=0.2\textwidth]{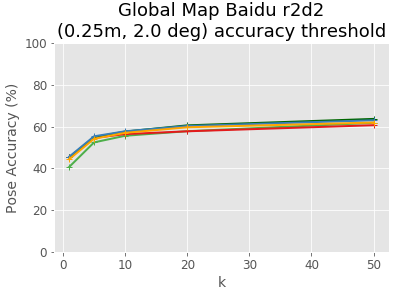}%
\\
\includegraphics[width=0.2\textwidth]{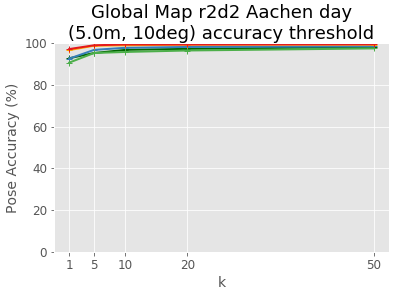}%
\includegraphics[width=0.2\textwidth]{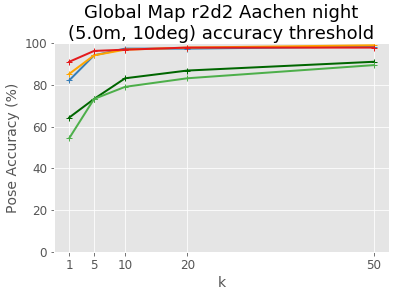}%
\includegraphics[width=0.2\textwidth]{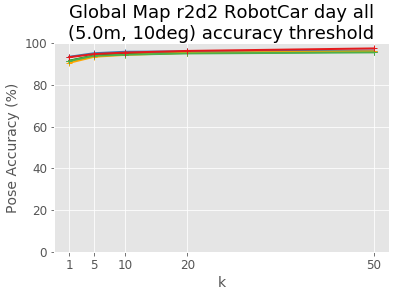}%
\includegraphics[width=0.2\textwidth]{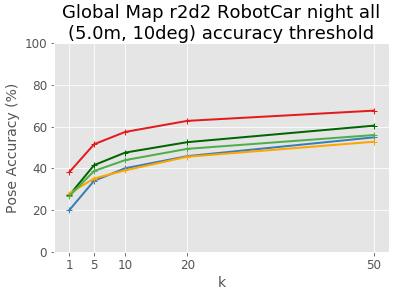}%
\includegraphics[width=0.2\textwidth]{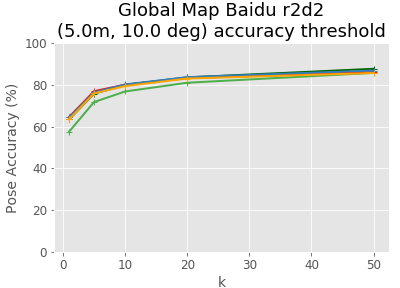}%
\\
\includegraphics[width=0.2\textwidth]{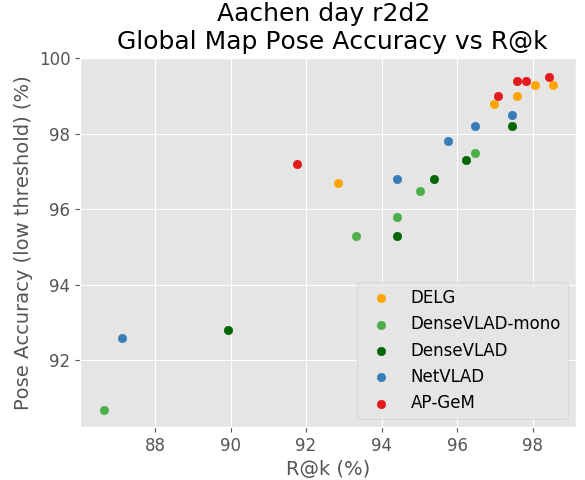}%
\includegraphics[width=0.2\textwidth]{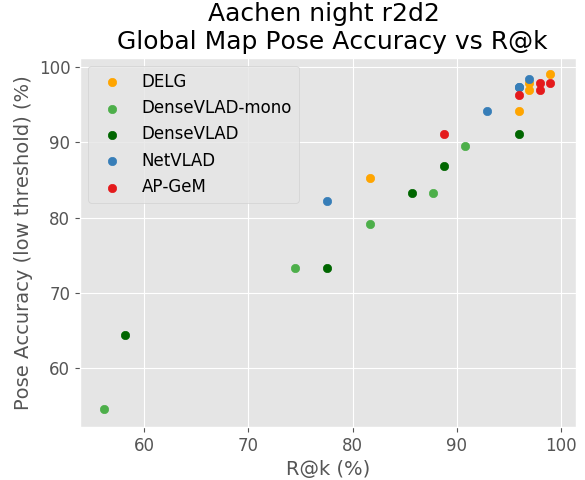}%
\includegraphics[width=0.2\textwidth]{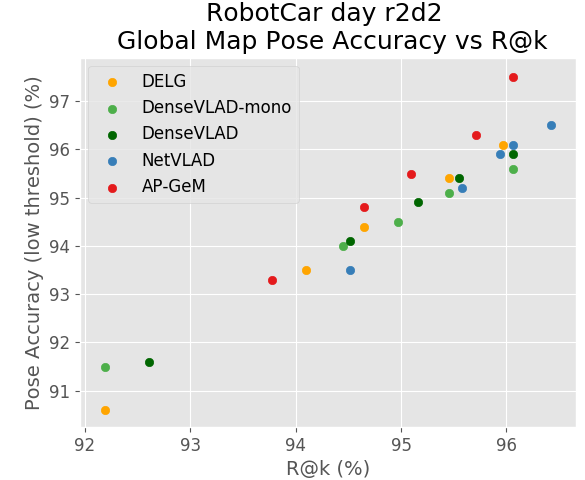}%
\includegraphics[width=0.2\textwidth]{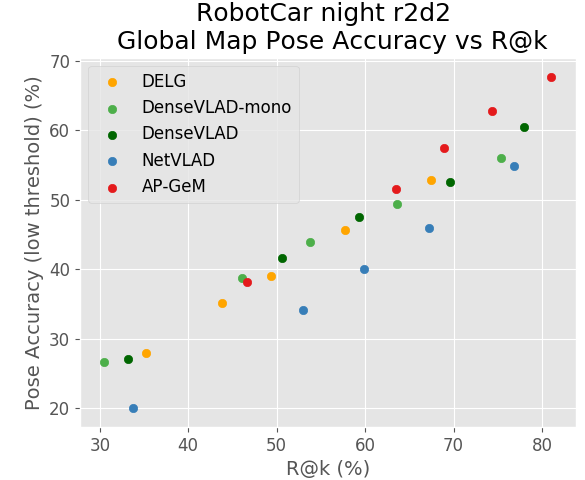}%
\includegraphics[width=0.2\textwidth]{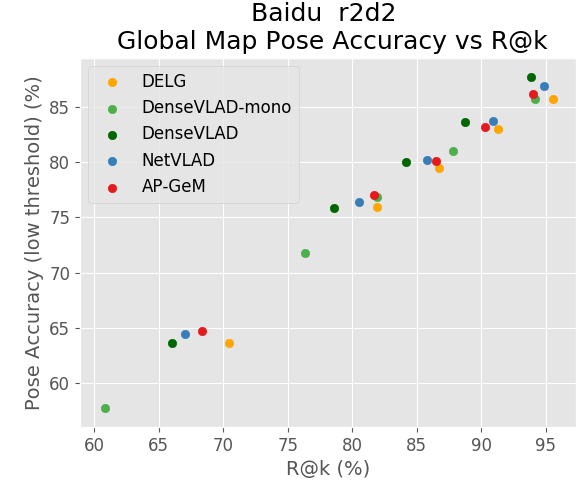}%
\end{center}
\vspace{-0.3cm}
   \caption{\textbf{Task 2b (pose estimation with a global map)}. 
   Top/middle row: percentage of images localized within the high/low
   %strictest/loosest 
   accuracy threshold as a function of the number $k$ of retrieved images. 
   %Results per dataset are shown per column. 
  The representations perform similarly for daytime scenarios and the indoor case, but AP-GeM best handles nighttime illumination changes.  
   Bottom row: scatter plots showing low accuracy localization results versus the retrieval metric R$@k$  for $k=1,5,10,20,50$. We observe a clear correlation between Task 2b and the landmark recognition task. %for the loose threshold (top right part) but not for the strict threshold (bottom left).
   }% 
   
   %In the top row we show local SfM results obtained with  the lowest accuracy thresholds, while on the bottom row we show how the results are correlated with R$@k$. Results per dataset are shown per column. 
   %day and night queries on Aachen, day and night queries on RobotCar, and Baidu. 
    % \TODO{We might combine the local and global into a single figure keeping only the low or high accuracy thresholds}}%
\label{fig:global_SFM}
\end{figure*}

\subsection{Task 2: Accurate pose estimation}
\label{sec:experiments:local_sfm}
We show results obtained with high and low accuracy thresholds for Task 2a (local SFM) in Fig.~\ref{fig:local_SFM} and for Task 2b (global map) in Fig.~\ref{fig:global_SFM}. 
%  In both cases 
%, but in general similar behavior can be observed for high and  low accuracy threshold (see Supplementary material).  
% we also show correlation plots between visual localization results and  the R$@k$ retrieval recall curves in Fig. \ref{fig:retrieval}. 

\PAR{Task 2a} We observe that all representations, even the handcrafted DenseVLAD descriptors, perform similarly well on the outdoor daytime and the indoor images, with DenseVLAD-mono being the exception in the latter case. 
We again observe that learned descriptors can perform better than DenseVLAD under day-night changes, with AP-GeM yielding the best results overall. 

Comparing Fig.~\ref{fig:retrieval} and Fig.~\ref{fig:local_SFM}, we do not see a clear correlation between retrieval/recognition and Task 2a: despite significant differences in P@$k$ and R@$k$, all methods perform similarly well for Task 2a on Aachen and RobotCar daytime images. 
In contrast, the better retrieval/recognition performance of AP-GeM on RobotCar night translates to better performance on Task 2a. 
However, the better P@$k$ performance of DELG and AP-GeM on Aachen night does not translate to a better Task 2a performance. 
In fact, there is no correlation between P@$k$, which decreases with increasing $k$ (\cf Fig.~\ref{fig:retrieval}), and Task 2a performance, which remains the same or increases with increasing $k$.

\PAR{Discussion} To achieve good pose accuracy using a local SFM model for a given set of retrieved images, a high R@$k$ score is not sufficient. 
This is due to the fact that more than one relevant image is needed to build the local map. 
At the same time, not all of the top $k$ retrieved images need to be relevant, \ie, a high P$@k$ score is not needed. 
% These 
% , they should contain enough relevant images to build the local Map (hence not correlated either with R$@k$)

Overall, retrieval/recognition performance is not a good indicator for Task 2a performance\footnote{Correlation plots between Task 2 performance and P@$k$ respectively R@$k$ are shown in the suppl.~mat.}.
This indicates that better retrieval representations can be learned that are tailored to the task of pose estimation without a global map, \eg, by designing a loss that optimizes pose accuracy. 

% %Hence local SfM is better correlated with precision than recall or mean precision. Indeed  there is not a clear correlation as shown in bottom row of 
% %\ref{fig:local_SFM}. 
% Indeed to get a high mean precision all images should be relevant in top k, while to get a good local SfM map we need to have sufficient images to build the map. Which explains that while P$@k$ is decreasing the locals SFM results are  rather similar between $k=5, 10, 20$. The only exception is Baidu where the percentage of relevant images is too low and considering more retrieved images increases the chances to get enough images to build the local map.
% \TODO{Does this make sense? Martin what you observed?}

%By the way

\PAR{Task 2b} 
A similar behavior of the different representations as in Task 2a can be seen in Fig.~\ref{fig:global_SFM}, \ie, all methods perform well on daytime and indoor images while learning-based methods perform better on the Aachen nighttime queries. 

The lower performance on the RobotCar nighttime and Baidu images can be explained by the comparably low R@$k$ for both datasets (\cf Fig.~\ref{fig:retrieval}), especially for $k\leq 10$. %, especially for 
% First we can observe that indeed when we have a low recall
Without retrieving at least one relevant image 
% (RobotCar night and Baidu for $k\leq10$),  which means the system did not even found at least one relevant image 
amongst the top $k$, pose estimation is bound to fail. % due to a lack of correct 2D-3D matches. 
% obviously it cannot register the query image. 

\PAR{Discussion} As shown in the bottom row of Fig.~\ref{fig:global_SFM}, there is a clear correlation between a high R@$k$ and good performance for the coarse pose threshold. 
 A higher R@$k$ increases the chance that an image can be localized at all. 
This validates the common practice to use state-of-the-art representations trained for retrieval/recognition for localization. 
However, the row also shows that a high R@$k$ does not necessarily imply a high pose accuracy. 
One explanation is that the retrieved images are relevant but share little visual overlap with the query image. 
In this case, all matches will be found in a small area of the query image resulting in an unstable (and thus likely inaccurate) pose estimate.

% On the contrary when the recall is high there are better chances that the image is localized, but not necessarily with high accuracy (\eg Robotcar day)

%\subsection{Task 2b: Global SfM}
%\label{sec:experiments:global_sfm}

\section{Conclusion}
Image retrieval plays an important role in modern visual localization systems.
Retrieval techniques are often used to efficiently approximate the pose of the query image or as an intermediate step towards obtaining a more accurate pose estimate. %, both with and without a global 3D map. 
Most localization systems simply use state-of-the-art image representations trained for landmark retrieval or place recognition. 
In this paper, we analyzed the correlation between the tasks of visual localization and retrieval/recognition through detailed experiments. 

Our results show that state-of-the-art image-level descriptors for place recognition are a good choice when localizing an image against a pre-built map as performance on both tasks is correlated. 
We can see that on the night images as well as on Baidu, AP-GeM often outperforms the other features. One of the reason might be that AP-GeM is the only feature that was trained not only with geometric data augmentation but also with color jittering. This might explain why it better handles day-night variations.
% % As can be expected, the learned descriptors (NetVLAD, AP-GeM, and DELG) typically outperform the SIFT-based DenseVLAD on this tasks.
% Features crafted for specific invariances from  data augmentation outperform when those invariances are present in the test set : AP-GeM is the only method trained with  % between day and night. % and between the high quality high resolution DSLR cameras and low resolution and low quality images taken with older generation cell phones (Baidu dataset). 

We also show that the tasks of pose approximation and localization without a pre-built map (local SFM) are not directly correlated with landmark retrieval/place recognition.
In the case of pose approximation, representations that reflect pose similarity in their descriptor similarities, \ie, exhibit robustness only to illumination changes, are preferable as they tend to retrieve closer images. 
For local SFM, there is a complex relationship between the retrieved images that is not captured by the classical Precision@$k$ and Recall@$k$ measures used for retrieval and recognition. 
Our results suggest that developing suitable representations tailored to these tasks are interesting directions for future work. 
% As can be expected, the learned descriptors (NetVLAD, AP-GeM, and DELG) typically outperform the SIFT-based DenseVLAD on most tasks. 
Our code and evaluation protocols are publicly available to support such research.

\small{
\PAR{Acknowledgments} This work received funding through the EU Horizon 2020 project RICAIP (grant agreeement No 857306) and the European Regional Development Fund under IMPACT No.~CZ.02.1.01/0.0/0.0/15 003/0000468.
}

% Amongst the trained descriptors, %NetVLAD, AP-GeM and DELG, 
% only NetVLAD  was trained for the place recognition task with the constraint that positive image pairs needs to have scene overlap. 
% \TODO{need to double check this }
% AP-GeM and DELG were trained in the classical Landmark retrieval setup, where  
% a relevant image pair  not necessarily needs to see the same part of the query landmark. 
% Thus, we would expect that NetVLAD outperforms the two other features on our setting, where positive pairs have visual overlap. 
% Yet, Fig.~\ref{fig:retrieval} shows that AP-GeM and DELG usually outperform NetVLAD. 
% The main reason might be that while NetVLAD was trained on a single city dataset (Pittsburgh~\cite{ToriiCVPR15PlaceRecognitionByViewSynthesis}), AP-GeM and DELG were trained on a large variety of landmarks from all over the world, with images taken under various conditions and by a large variety of cameras. 
% The larger variety in the training data likely leads to a better generalization to new scenes. %and hence generalises better to new scenes. 

\newpage
%\onecolumn
\appendix

\section{Introduction}
This supplementary material complements the main paper with further details on the experimental setup (Section~\ref{sec:setup}) and additional results (Section~\ref{sec:results}) to strengthen the findings.

Section~\ref{sec:setup} is structured as follows. First, in Section~\ref{sec:framework:datasets} we give more information on the localization datasets used in our experiments. In Section~\ref{sec:global_features}
we describe the global image
representations that were compared in the benchmark
%and in Section~\ref{sec:local_features} the local features used to build local or global SFM. For both types of features we also
and we provide links to the codes we used to extract them.
In Section~\ref{sec:SFM}, we briefly recall the SFM pipeline and point to the codes on which we relied upon for running the experiments. In Section~\ref{sec:metrics} we recall the evaluation metrics to make the plots in the supplementary easily understandable.

In Section~\ref{sec:results}, we provide additional figures to Sections {\bf 4.1}, {\bf 4.2} and {\bf 4.3} of the main paper, corresponding here to Sections~\ref{sec:placeRec},~\ref{sec:approximate}, and \ref{sec:accurate}. In Section~\ref{sec:posevsRetrieval}, we show additional correlation analyses between localization and retrieval measures.

\section{Experimental setup}
\label{sec:setup}

\subsection{Datasets}
\label{sec:framework:datasets}

To evaluate the role of image retrieval in visual localization, we selected three public datasets aimed at benchmarking visual localization: \textbf{Aachen Day-Night-v1.1}~\cite{SattlerCVPR18Benchmarking6DoFOutdoorLoc,SattlerBMVC12ImRetLocalizationRevisited,Zhang2020ARXIV}, \textbf{RobotCar Seasons}~\cite{Maddern2017IJRR,SattlerCVPR18Benchmarking6DoFOutdoorLoc} and \textbf{Baidu Mall}~\cite{SunCVPR17DatasetBenchmarkingLocalization}.
%The Aachen and RobotCar datasets are widely used and cited for visual localization~\cite{SattlerCVPR18Benchmarking6DoFOutdoorLoc} but do not provide public ground truth poses for the test images. The Baidu dataset, on the other hand, provides ground truth poses what makes it very interesting for our benchmark %(\cite{WeinzaepfelCVPR19VisualLocObjectsOfInterestDenseMatchRegression} also used it to evaluate their method and to compare it with the state of the art).
Altogether, the selected datasets cover a variety of application scenarios: large-scale outdoor handheld localization under varying conditions (Aachen Day-Night), small-scale indoor handheld localization with occlusions (Baidu), and large-scale autonomous driving (RobotCar Seasons).
%We hope that by providing multiple competitive baselines on this dataset, it receives more interest.
%The selected datasets represent different application scenarios and cover a variety of challenges image retrieval has to face.
%Thus, they provide a meaningful database for the presented benchmark.

%\subsubsection{Aachen Day-Night}
%\label{sec:framework:datasets:aachen}

\PAR{The Aachen Day-Night-v1.1~\cite{SattlerCVPR18Benchmarking6DoFOutdoorLoc,SattlerBMVC12ImRetLocalizationRevisited,Zhang2020ARXIV} dataset}
% The Aachen Day-Night dataset
%contains 4,447 high-quality training/database and 1015 test/query images from the old inner city of Aachen, Germany. The database images are taken under day-time conditions using multiple handheld cameras. The query images are captured with 3 mobile phones during day and night-time. This dataset represents a handheld scenario similar to augmented or mixed reality applications in city-scale environments. Among the training (resp.~test) images, 119 day (resp. 93 night) images come from the newly released Aachen v1.1~\cite{Zhang2020ARXIV}. We also use the corrected test images through submission on \textit{visuallocalization.net}. As we did not have access to the new poses, but only the ones of Aachen v1, we used this images of the first edition of the dataset only to compute the retrieval metrics.
contains 6,697 high-quality training/database and 1015 test/query images from the old inner city of Aachen, Germany. The database images are taken under daytime conditions using handheld cameras. The query images are captured with three mobile phones at day and at night. This dataset represents a handheld scenario similar to augmented or mixed reality applications in city-scale environments.
%We had access to the ground truth poses of a subset of the test images which we used to compute the retrieval metrics described in Section~3.2 of the main paper.

\PAR{The RobotCar Seasons~\cite{Maddern2017IJRR,SattlerCVPR18Benchmarking6DoFOutdoorLoc} dataset}
% The RobotCar Seasons dataset
is based on a subset of the RobotCar dataset~\cite{Maddern2017IJRR}, captured in the city of Oxford, UK. The training sequences (26,121 images) are captured during daytime, the query images (11,934) are captured during different traversals and under changing weather, time of the day, and seasonal conditions. In contrast to the other two datasets used, the RobotCar dataset contains multiple synchronized cameras and the images are provided in sequences. However, in our benchmark we did not use this additional information. Note that integrating this information in the benchmark can be an interesting follow up of this paper.
%spatial (multiple cameras) fusion of images can be evaluated with this dataset.

%\subsubsection{Baidu Image-Based Localization}
%\label{sec:framework:datasets:robotcar}
\PAR{The Baidu Mall~\cite{SunCVPR17DatasetBenchmarkingLocalization} dataset}
% The third dataset is the Baidu localization dataset which
was captured in a modern indoor shopping mall in China. It contains 689 training images captured with high resolution cameras in the empty mall and about 2,300 mobile phone query images taken a few months later while the mall was open. The images were semi-manually registered into a LIDAR scan in order to obtain the ground truth poses. The query images are of much poorer quality compared to the database images. In contrary to the latter, where all images were taken in parallel or perpendicular with respect to the main corridor of the mall, query images were taken from more varying viewpoints.
Furthermore, the images contain reflective and transparent surfaces, moving people, and repetitive structures which are all important challenges for visual localization and image retrieval.
%This dataset covers an indoor scenario which, because of large viewpoint changes close to structures and walls, is particularly challenging for image retrieval but also very relevant for visual localization (\eg~because of the lack of GNSS indoors).

\subsection{Global image representations for retrieval}
\label{sec:global_features}

Our benchmark compares the following 4 popular image representations using the models provided by the authors.
%a BOW-like representation,DenseVlad~\cite{ToriiCVPR15PlaceRecognitionByViewSynthesis}, and three popular deep image representations, NetVlad~\cite{ArandjelovicCVPR16NetVLADPlaceRecognition}, and AP-GeM~\cite{RevaudICCV19LearningwithAPTrainingImgRetrievalListwiseLoss}.

\PAR{DenseVLAD\footnote{Code available at~\url{http://www.ok.ctrl.titech.ac.jp/~torii/project/247/}.} \cite{ToriiCVPR15PlaceRecognitionByViewSynthesis}}
To obtain the DenseVLAD image representation for an image, first RootSIFT~\cite{ArandjelovicCVPR12ThreeThingsEveryoneObjRet,LoweIJCV04DistinctiveImageFeaturesScaleInvariantKeypoints} descriptors are extracted on a multi-scale (we used 4 different scales corresponding to region widths of 16, 24, 32 and 40 pixels), regular, densely sampled grid, and then aggregated into an intra-normalized~VLAD~\cite{JegouCVPR10AggregatingLocalDescriptors} descriptor followed by PCA (principle component analysis) compression, whitening, and L2 normalization~\cite{JegouECCV12NegativeEvidencesCoOccurrences}.
We also used DenseVLAD-mono, which is a variant where the local features are extracted only on a single scale (with the region width equal to 24).
%DenseVLAD was often used in structure-based visual localization methods to scale them to large scenes~\cite{ToriiCVPR15PlaceRecognitionByViewSynthesis,SattlerCVPR17AreLargeScale3DModelsNecessaryForLocalization} and even compared to recent deep regression models~\cite{SattlerCVPR19UnderstandingLimitationsPoseRegression}.
%We selected this representation as there exists a direct comparison of pose approximation with DenseVLAD compared to recent deep regression models~\cite{SattlerCVPR19UnderstandingLimitationsPoseRegression}.

\PAR{NetVLAD\footnote{Matlab code and pretrained models are available at~\url{https://github.com/Relja/netvlad}. We used the VGG-16-based NetVLAD model trained on Pitts30k~\cite{ArandjelovicCVPR16NetVLADPlaceRecognition}.} \cite{ArandjelovicCVPR16NetVLADPlaceRecognition}}
The main component of the NetVLAD architecture is a generalized VLAD layer
that aggregates mid-level convolutional features extracted from the entire image into a compact single vector representation for efficient indexing similarly to  VLAD~\cite{JegouCVPR10AggregatingLocalDescriptors}.
The resulting aggregated representation is then compressed using PCA to obtain a final compact descriptor of the image.
NetVLAD is trained with geo-tagged image sets consisting of groups of images taken from the same locations at different times and seasons, allowing the network to discover which features are useful or distracting and what changes should the image representation be robust to.
This makes NetVLAD very interesting for the visual localization pipeline and motivated our choice to integrate it in our benchmark.
Furthermore, NetVLAD has been used in state-of-the-art localization pipelines~\cite{SarlinCVPR19FromCoarsetoFineHierarchicalLocalization,Germain20193DV} and in combination with D2-Net~\cite{DusmanuCVPR19D2NetDeepLocalFeatures}.
%and R2D2~\cite{RevaudNIPS19R2D2ReliableRepeatableDetectorsDescriptors} features.

\PAR{AP-GeM\footnote{Pytorch implementation and models are available at \url{https://europe.naverlabs.com/Research/Computer-Vision/Learning-Visual-Representations/Deep-Image-Retrieval/}. We used the Resnet101-AP-GeM model trained on Google Landmarks~\cite{NohICCV17LargeScaleAttentiveDeepLocalFeatures} to extract image representations.}
\cite{RevaudICCV19LearningwithAPTrainingImgRetrievalListwiseLoss}}
This model, similarly to \cite{RadenovicPAMI19FineTuningCNNImRet}, uses a generalized-mean pooling layer (GeM) to aggregate CNN-based descriptors of several image regions at different scales
%an extension of
%\cite{GordoIJCV17EndToEndDeepVisualReprImRet}, which considers sum-aggregates of CNN-based descriptors of several image regions at different scales, where the RMAC pooling~\cite{ToliasICLR16ParticularObjRetIntegralMaxpoolingCNN} weights are learned in an end-to-end manner.
%However, AP-GeM~\cite{RevaudICCV19LearningwithAPTrainingImgRetrievalListwiseLoss}, as well as \cite{RadenovicPAMI19FineTuningCNNImRet}, uses a generalized-mean pooling layer (GeM) instead of RMAC pooling, and
but instead of a contrastive loss, it directly optimizes the Average Precision (AP) approximated by histogram binning to make it differentiable. It is one of the state-of-the art image representation on  popular landmark retrieval benchmarks (\eg, $\mathcal{R}$Oxford and $\mathcal{R}$Paris \cite{RadenovicCVPR18RevisitingOxfordParisImRetBenchmarking}).
The model we used was trained on the Google Landmarks v1 dataset (GLD)~\cite{NohICCV17LargeScaleAttentiveDeepLocalFeatures}, where each training image has a class label based on the landmark contained in the image.  AP-GeM has been used successfully used for visual localization in \cite{humenberger2020robust}.

\PAR{DELG\footnote{We used the TensorFlow code publicly available at \url{https://github.com/tensorflow/models/tree/master/research/delf/delf/python/delg} and the model with a ResNet50 backbone architecture trained on GLD \cite{NohICCV17LargeScaleAttentiveDeepLocalFeatures}.}\cite{CaoX20UnifyingDeepLocalGlobalFeatures}} DELG is designed to extract local and global features using one CNN.
After a common backbone, the model is split into two parts (heads) one to detect relevant local features and one which describes the global content of the image as a compact descriptor. The two networks  are jointly trained in an end-to-end manner
%head, according to an idea first explored in \cite{Sarlin2019CVPR}. Here, both the local and global heads are trained in an end-to-end manner, while \cite{Sarlin2019CVPR} uses distillation of 2 networks. The global head is trained
using the ArcFace\cite{ArcFaceCVPR2019Deng} loss for the compact descriptor %, which is a classification loss.
and leveraging the Google Landmark v1\cite{NohICCV17LargeScaleAttentiveDeepLocalFeatures} dataset, which provides image-level labels. The method was originally designed for image search, where the local features enable geometric verification and re-ranking. \\

Our choice of VLAD with densely extracted features (DenseVLAD, NetVLAD) is based on \cite{TairaCVPR18InLocndoorVisualLocalization,ToriiTPAMI2018}. It is shown that DenseVLAD and DenseFV (Fisher Vectors) significantly outperform SparseVLAD and SparseFV (based on local features) under strong illumination changes as using densely extracted features eliminates potential repeatability problems of feature detectors. Furthermore, \cite{ToriiTPAMI2018} reports that DenseVLAD performs on par with advanced sparse bag of visual words representations. At the same time, both DenseVLAD and NetVLAD are used in state-of-the-art localization pipelines~\cite{SarlinCVPR19FromCoarsetoFineHierarchicalLocalization,TairaCVPR18InLocndoorVisualLocalization,Taira2019ICCV}.
Investigating aggregation of modern local features (e.g. D2-Net, R2D2) via ASMK~\cite{ToliasPR14VisualQueryExpansionFeatureAggregation}, VLAD~\cite{JegouCVPR10AggregatingLocalDescriptors} or FV~\cite{PerronninCVPR07FisherKernels} is an interesting research direction with practical benefits. %In this paper, we  focused on image representations that are currently commonly used in state-of-the-art localization pipelines

\subsection{SFM pipeline with and without a global model}
\label{sec:SFM}
%For our global SFM we follow the standard approach from the literature, \ie, we match global features between the query and the retrieved database images~\cite{IrscharaCVPR09FromSFMLocationRecognition,SattlerBMVC12ImRetLocalizationRevisited,SarlinCVPR19FromCoarsetoFineHierarchicalLocalization,Germain20193DV,humenberger2020robust}.
%We create an SFM 3D model for each dataset and each local feature type we considered (RootSIFT, D2Net and R2D2),  relying on the provided
%camera poses in the dataset and triangulating local keypoints matched between database images.

% Martin ----------------------------------
For our global SFM experiments, we created an SFM model for each local feature type considered (SIFT, D2Net and R2D2). To not introduce a bias towards one specific global feature type and because matching all possible training image pairs would %, on the one hand,
potentially introduce noise %,  and on the other hand,
and would require large computational resources, we selected the image pairs to match by their frustum overlaps. In detail, we fitted a sphere into the overlapping space of two frusta (we used 50m as maximum distance) and used the radius of this sphere as our measure for image overlap. For Aachen Day-Night and Baidu, we used all image pairs with an overlapping-sphere-radius of 10m or more. For RobotCar, this threshold resulted in too many image pairs to process, thus we only used the 50 most overlapping pairs. To generate the 3D model we triangulated the 3D points from the local feature matches (obtained from the image pairs) using the provided camera poses of the training images and applied bundle adjustment for global optimization.
In order to localize an image within this map, we match the query images with the top $k$ retrieved database images and use PNP~\cite{KneipCVPR11ANovelParametrizationAbsoluteCamPose,Kukelova13ICCV} to register them within the map.

Our local SFM experiments are inspired by the SFM-on-the-fly approach from~\cite{Torii2019TPAMI} where the retrieved database images are used to create a small SFM map on-the-fly and the query images are registered within this map using PNP. The SFM pipeline is the same as described above with the difference that the database image pairs to match are generated using all possible pairwise combinations of the retrieved images.
%----------------------------------

%Our local SFM model is inspired by the SFM-on-the-fly approach from~\cite{Torii2019TPAMI} where the retrieved database images  with known camera poses are used to triangulate the 3D structure of the scene on-the-fly and then register the query in this map with PnP~\cite{KneipCVPR11ANovelParametrizationAbsoluteCamPose,Kukelova13ICCV}.

In both cases, for SFM and query image registration, we used COLMAP\footnote{Code available at \url{https://colmap.github.io/}.}~\cite{SchonbergerCVPR16StructureFromMotionRevisited}.

\subsection{Evaluation metrics}
\label{sec:metrics}

% In order to improve
To increase the readability of the following experimental results, we first recall the evaluation metrics used in the main paper.

\PAR{Visual localization metrics.} To measure localization performance, we follow common practice from the literature~\cite{KendallICCV15PoseNetCameraRelocalization,ShottonCVPR13SceneCoordinateRegression,SattlerCVPR18Benchmarking6DoFOutdoorLoc}.
Let $\mathtt{R}\in \mathbb{R}^{3\times 3}$ be the camera rotation and $\mathbf{c}\in\mathbb{R}^3$ be the camera position, \ie, a 3D point $\mathbf{X}_g$ in world coordinates is mapped to local camera coordinates as $\mathbf{X}_l = \mathtt{R}(\mathbf{X}_w - \mathbf{c})$. Following~\cite{SattlerCVPR18Benchmarking6DoFOutdoorLoc}, the position and rotation errors between an estimated and the reference pose are defined as
\begin{eqnarray}
c_\text{error} & = & \lVert \mathbf{c}_\text{estimated} - \mathbf{c}_\text{reference} \rVert_2 \enspace , \label{eq:position_error} \\
R_\text{error} & = & \arccos\left(\frac{\text{trace}\left(\mathtt{R}_\text{estimated}^{-1} \cdot \mathtt{R}_\text{reference}\right) - 1}{2} \right) \enspace , \label{eq:orientation_error}
\end{eqnarray}
where $R_\text{error}$ is the angle of the smallest rotation aligning $\mathtt{R}_\text{estimated}$ and $\mathtt{R}_\text{reference}$.

For evaluation, we use the %median position and orientation errors, as well as the
percentage of images localized within a given error threshold $(X\text{m}, Y^\circ)$~\cite{ShottonCVPR13SceneCoordinateRegression,SattlerCVPR18Benchmarking6DoFOutdoorLoc}, \ie, the percentage of query images for which $c_\text{error} < X$ and $R_\text{error}<Y$. Following \cite{SattlerCVPR18Benchmarking6DoFOutdoorLoc}, for all three datasets we use three different threshold pairs for evaluating \textbf{low} (5m, 10$^\circ$), \textbf{medium} (0.5m, 5$^\circ$), and \textbf{high} (0.25m, 2$^\circ$) accuracy localization.
%

%For \textbf{Aachen Day-Night v1.1} and \textbf{RobotCar Seasons}, this metric is computed by submitting obtained camera positions to visuallocalization.net. For \textbf{Baidu-mall}, we directly use the provided ground truth poses.

\PAR{Retrieval metrics.} As indicated in the main paper, (index, retrieved image) pairs are considered positive based on either the IoU similarity or the
distance between the camera poses. We use them to compute both \emph{Precision@$k$} (P@$k$) and \emph{Recall@$k$} (R@$k$). In the case of IoU similarity, we use the presence of overlapping triangulated 3D points between query and database images to classify a pair as positive.
In the case of pose similarity,
%(referred as IoU based relevance mesure).
%\todo{ We might want to give statistics of how many query images were ignored during evaluation because we had not enough   reconstructed 3D points.}
%However, as mentioned in the  paper, alternatively
we consider a database image as relevant if it was taken within a neighborhood of the query image ~\cite{ToriiCVPR15PlaceRecognitionByViewSynthesis,ToriiPAMI15VisualPlaceRecognRepetitiveStructures,ArandjelovicACCV14DislocationDistinctivenessForLocation,SattlerCVPR16LargeScaleLocationRecognitionGeometricBurstiness}. % (especially in geo-localization when only GPS coordinates of the image are known).
When the camera orientation is available (which is true in our case), the angle between the cameras' orientations can also be taken into account. Following \cite{ToriiPAMI15VisualPlaceRecognRepetitiveStructures}, we consider two images relevant if they were taken within 25 meters from each other and if the camera orientations are less than 45 degrees apart. %We refer to it as pose distance based relevance measure.
While in the main paper we mainly focused on IoU measure-based relevance, here we provide P@$k$ / R@$k$ results obtained with both relevance measures.

\begin{figure*}[t]
\begin{center}
\includegraphics[width=0.20\textwidth]{plots/retrievalmetrics/aachen/Pk_Aachenday_.png}%
\includegraphics[width=0.20\textwidth]{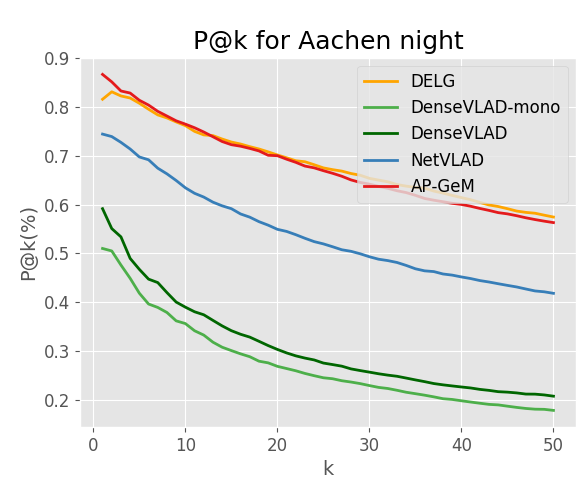}%
\includegraphics[width=0.20\textwidth]{plots/retrievalmetrics/robotcar/Pk_RobotCarday_.png}%
\includegraphics[width=0.20\textwidth]{plots/retrievalmetrics/robotcar/Pk_RobotCarnight_.png}%
\includegraphics[width=0.20\textwidth]{plots/retrievalmetrics/baidu/Pk_Baidu_.png}%
\\
\includegraphics[width=0.20\textwidth]{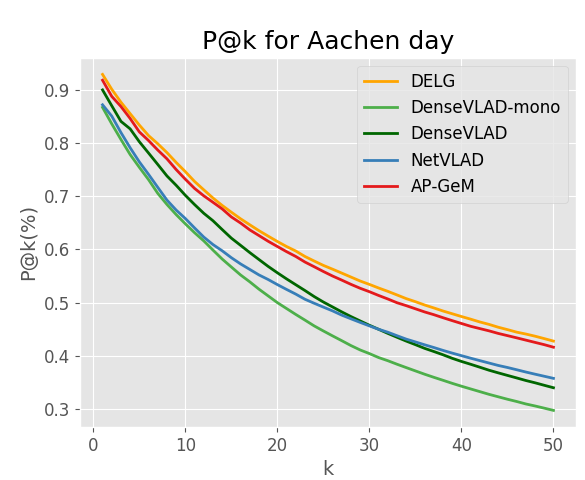}%
\includegraphics[width=0.20\textwidth]{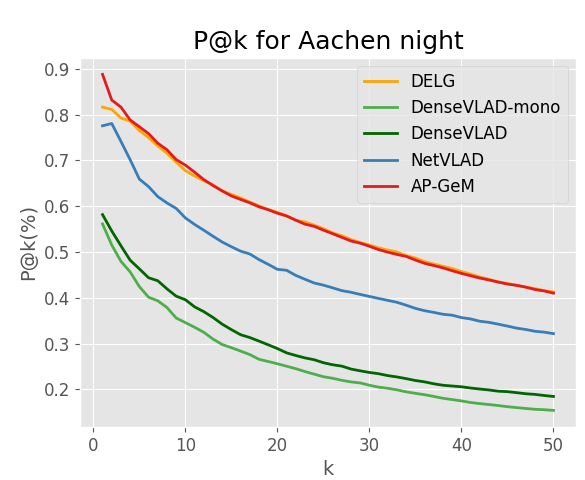}%
\includegraphics[width=0.20\textwidth]{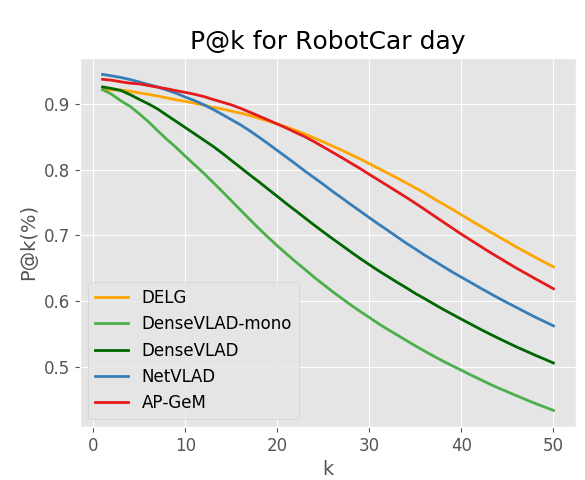}%
\includegraphics[width=0.20\textwidth]{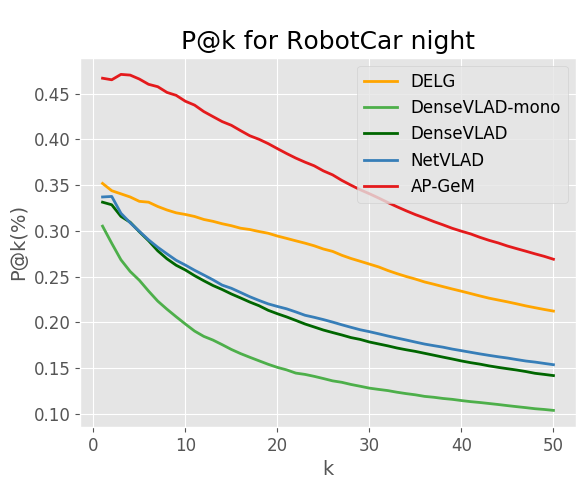}%
\includegraphics[width=0.20\textwidth]{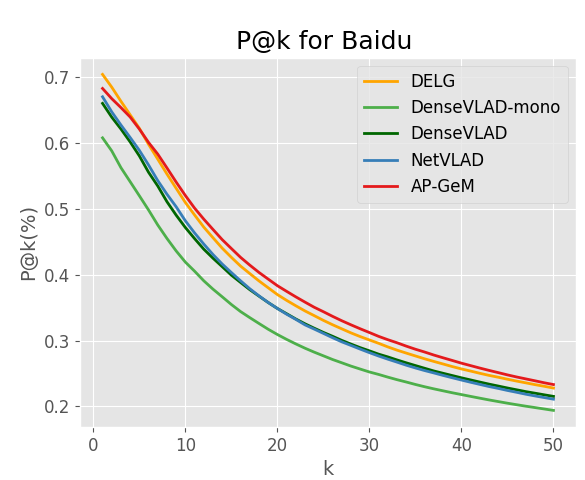}%
\end{center}
   \caption{\textbf{P@$k$ curves}. Landmark retrieval result where the ground truth relevance between two images was defined by visual overlap (top row) respectively pose similarity (bottom row).}
   %defining whether two images are related based on jointly observed 3D points, \ie, visual overlap. (bottom row) defining whether two images are related based on pose similarity, \ie, both images are taken from roughly the same position and orientation.}
\label{fig:retrieval_precision}
\end{figure*}

\begin{figure*}[t]
\begin{center}
\includegraphics[width=0.20\textwidth]{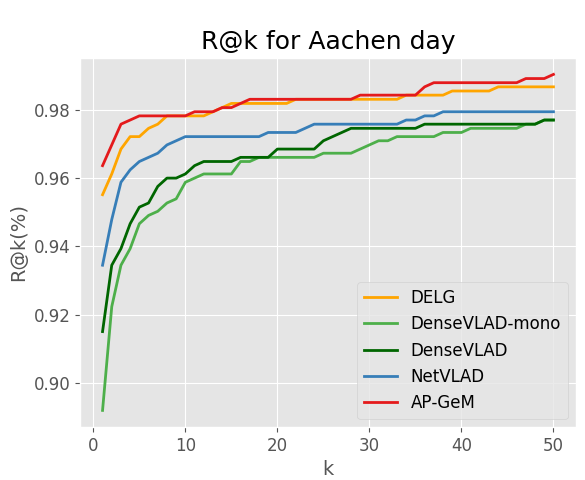}%
\includegraphics[width=0.20\textwidth]{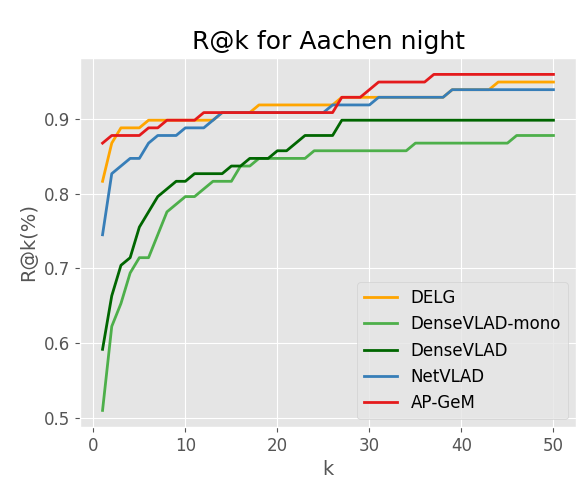}%
\includegraphics[width=0.20\textwidth]{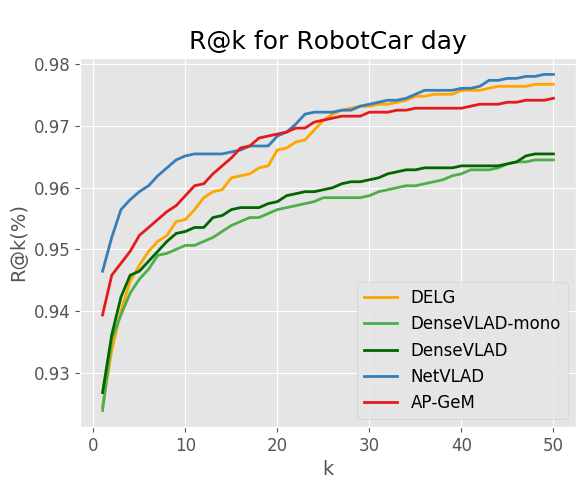}%
\includegraphics[width=0.20\textwidth]{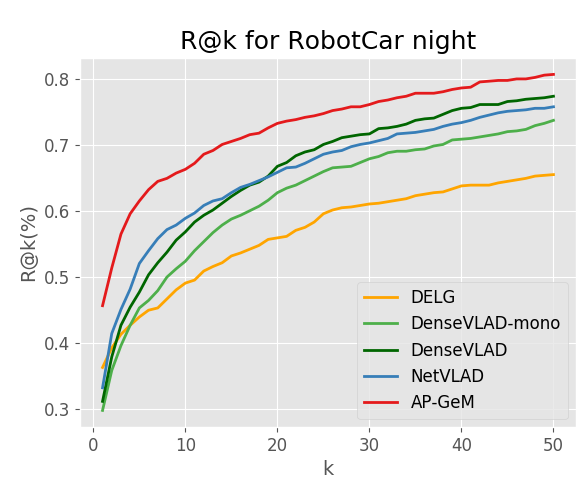}%
\includegraphics[width=0.20\textwidth]{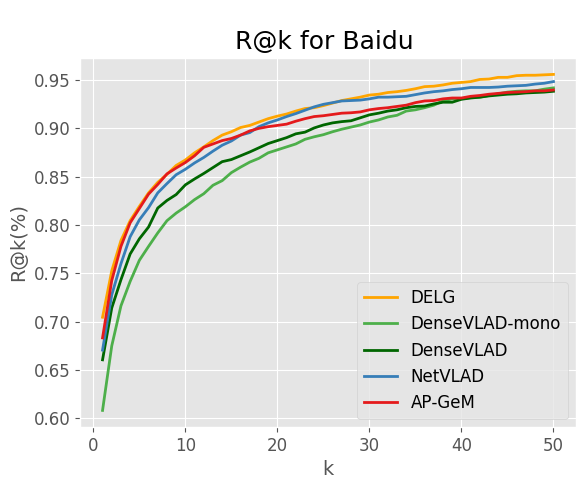}%
\\
\includegraphics[width=0.20\textwidth]{plots_pose/retrievalmetrics/aachen/Rk_Aachenday_.png}%
\includegraphics[width=0.20\textwidth]{plots_pose/retrievalmetrics/aachen/Rk_Aachennight_.png}%
\includegraphics[width=0.20\textwidth]{plots_pose/retrievalmetrics/robotcar/Rk_RobotCarday_.png}%
\includegraphics[width=0.20\textwidth]{plots_pose/retrievalmetrics/robotcar/Rk_RobotCarnight_.png}%
\includegraphics[width=0.20\textwidth]{plots_pose/retrievalmetrics/baidu/Rk_Baidu_.png}%
\end{center}
   \caption{
   \textbf{R@$k$ curves}. Place recognition results where the ground truth relevance between two images is defined by visual overlap (top row) respectively pose proximity (bottom row).}
    %\textbf{R@$k$ curves}: (top row) defining whether two images are related based on jointly observed 3D points, \ie, visual overlap. (bottom row) defining whether two images are related based on pose similarity, \ie, both images are taken from roughly the same position and orientation.}
\label{fig:retrieval_recall}
\end{figure*}

\section{Additional experimental results}
\label{sec:results}

This section presents results that complement the main paper.
In Section~\ref{sec:placeRec}, we present the retrieval and place recognition results
(\cf Sec.~3.2, Place recognition, in the main paper).
We compare the results obtained with ground truth image relevance defined
based on visual overlap with the ones defined based on pose proximity.
%two different metrics described in  Section~\ref{sec:metrics} to define if two images are related or not, namely the  IoU based and the pose distance based relevance measure.
%alternative pose-based
%In Section~\ref{sec:posevsRetrieval}, we show the correlation between the interpolation-based localization results and \emph{Recall@$k$}.
 %based on the pose similarity-based relevance definition. We also
% provide pose approximation results with a more stricter localization threshold
%(\textbf{medium}: (0.5m, 5$^\circ$)).
In Section~\ref{sec:approximate}, we present further details on our experiments on pose approximation.
In Section~\ref{sec:accurate}, we provide results for various localization accuracy settings with SFM maps built with different local feature types.
Finally, in Section~\ref{sec:posevsRetrieval}, we study the correlation between visual localization and place recognition/landmark retrieval metrics as well as the correlation between the interpolation-based localization results and \emph{Recall@$k$}.

\subsection{Landmark retrieval and place recognition}
\label{sec:placeRec}

%\subsection{Retrieval Metrics according to Pose and 3D overlap definitions}

As discussed in the main paper, we use two retrieval metrics from the literature:  \emph{Precision@$k$} (P@$k$) and  \emph{Recall@$k$} (R@$k$) %(\cf above for their definition).
where we considered two different ways to define whether two images are related / relevant to each other.
One is based on visual overlap (measured using jointly visible 3D points) %which we use for the P@$k$ in the main paper
and the other is based on pose proximity (whether the two images were roughly taken from the same position and, if available, orientation). %, which is used for the R@$k$.
%Note that in the former case

In Fig.~\ref{fig:retrieval_precision}, we show results for the landmark retrieval
scenario evaluated with the P@$k$ metric considering ground truth relevance scores
obtained either with visual overlap or with pose proximity.
%In the following, we show results for both metrics for all the datasets and for both definitions of relevance.
%Fig.~\ref{fig:retrieval_precision} shows results for the P@$k$ metric for both definitions.
Similarly, Fig.~\ref{fig:retrieval_recall} shows results for the place recognition scenario measured by
R@$k$ using the two relevance scores.
%both definitions. %( using the Place Recognition positives and the 3D Points positives.
Overall, we can see that the results with both relevance scores are
very similar yielding to the same rankings between image representations.
%are consistent.
This shows that the exact definition of the relevance score between two images does not play a major role when evaluating which retrieval representation works well for landmark retrieval and place recognition tasks.
% As it is harder to have 3D-matching points (for example, Baidu-mall has extremely blurry images), while every image has a corresponding positive according to the distances differences definition, the scales of the y axis are overall different.

\begin{figure*}[t]
\begin{center}
%\includegraphics[width=0.3\textwidth]{plots/interpolation/aachen/Aachen Day-Night v1.1_day_mid_equi.png}%
%\includegraphics[width=0.3\textwidth]{plots/interpolation/aachen/Aachen Day-Night v1.1_day_mid_interp_no-legend.png}%
%\includegraphics[width=0.3\textwidth]{plots/interpolation/aachen/Aachen Day-Night v1.1_day_mid_weighting_no-legend.png}%
%\\
\includegraphics[width=0.3\textwidth]{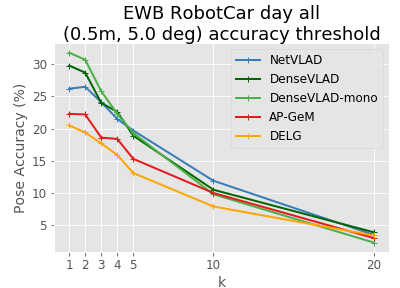}%
\includegraphics[width=0.3\textwidth]{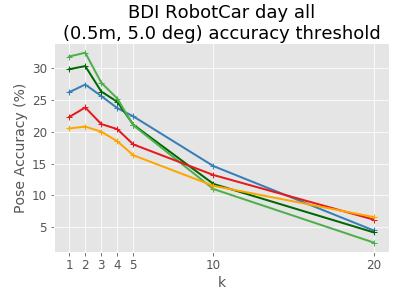}%
\includegraphics[width=0.3\textwidth]{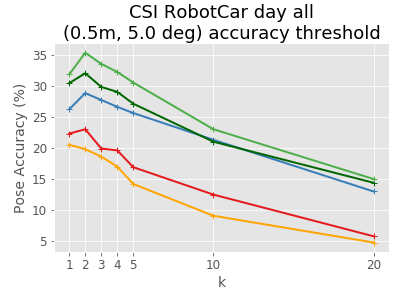}%
%\\
%\includegraphics[width=0.3\textwidth]{plots/interpolation/baidu/Baidu_(0.5m, 5.0 deg)_equi_no-legend.png}%
%\includegraphics[width=0.3\textwidth]{plots/interpolation/baidu/Baidu_(0.5m, 5.0 deg)_interp_no-legend.png}%
%\includegraphics[width=0.3\textwidth]{plots/interpolation/baidu/Baidu_(0.5m, 5.0 deg)_weighting_no-legend.png}%
\end{center}
  \caption{\textbf{Task 1: Pose approximation with medium accuracy threshold}.
  We only show results for RobotCar day as less than 1-2\% of the queries were localized for the others. On these plots, we observe that  NetVLAD and DenseVLAD outperform DELG and AP-GeM.}
  % Due to the small variation in camera pose between query and reference images, a relative large percentage of the query images can be localized with medium precision for the RobotCar dataset.   For RobotCar day, NetVLAD and DenseVLAD are performing better than DELG and AP-GeM at this threshold, while the opposite is the case for the low threshold (\cf Fig. 3 in the main paper). This again demonstrates the advantage of using descriptors with limited robustness to viewpoint changes.  Less than 1\% of the queries are localized for the other two datasets. As a result, the curves are very noisy.}
\label{fig:interp_supp}
\end{figure*}

\subsection{Task 1: Pose approximation}
\label{sec:approximate}

 %For the interpolation methods, we report every threshold, along with the theoretical top-1 best. As shown, for lower interpolation thresholds, the performances of interpolation methods are hardly satisfying.

In the main paper, we show pose approximation results for the three approaches (EWB, BDI, and CSI) considering the low localization accuracy threshold (5m, 10$^\circ$). In Fig.~\ref{fig:interp_supp}, here we show pose approximation results for RobotCar day for the medium accuracy threshold (0.5m, 5$^\circ$). We only show results for this dataset as for the other datasets less than 1-2\% of the queries were successfully localized with medium accuracy. The better results obtained for this dataset are due to the small variation in camera pose between query and reference images.
Note that the lower performance for the
RobotCar nighttime queries is due to the lower quality of the nighttime images (which exhibit strong motion blur) and the challenge of bridging the appearance gap between day- and nighttime images. % induced by variations between day- and nighttime images.
%While due to the small variation in camera pose between query and reference images, about one third of  the query images can be localized with medium precision for the RobotCar day dataset,  for the other datasets
%
Interestingly, our results show that for the medium accuracy threshold, NetVLAD and DenseVLAD perform better than DELG and AP-GeM.
This is in contrast to the low accuracy threshold where DELG and AP-GeM perform better (see main paper) and means that more images were successfully localized with DELG and AP-GeM but not very accurately.

\subsection{Task 2: Accurate pose estimation}
\label{sec:accurate}
As discussed in Section~4 of the main paper, in the following we present the full set of results for \textbf{Task 2a} and \textbf{Task 2b} considering all three accuracy thresholds and using three different local feature types to build the SFM maps, namely R2D2~\cite{RevaudNIPS19R2D2ReliableRepeatableDetectorsDescriptors}, D2-Net~\cite{DusmanuCVPR19D2NetDeepLocalFeatures}, and SIFT~\cite{LoweIJCV04DistinctiveImageFeaturesScaleInvariantKeypoints}.

\begin{figure*}[t]
\begin{center}
\includegraphics[width=0.3\textwidth]{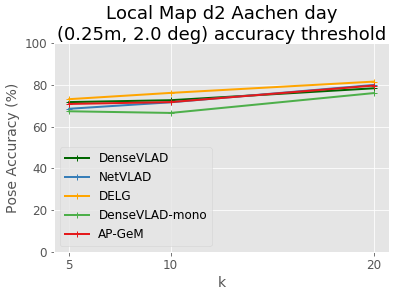}%
\includegraphics[width=0.3\textwidth]{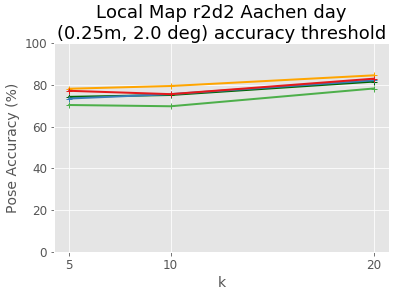}%
\includegraphics[width=0.3\textwidth]{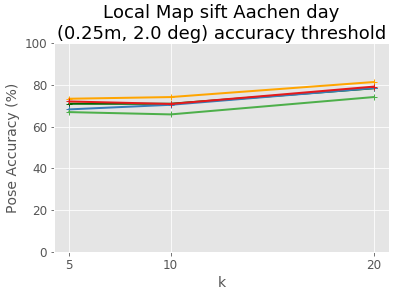}%

\includegraphics[width=0.3\textwidth]{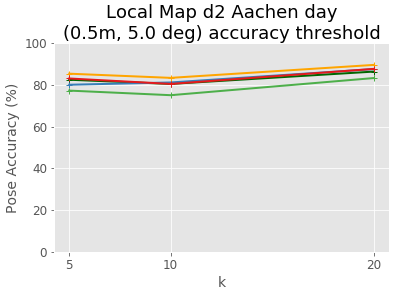}%
\includegraphics[width=0.3\textwidth]{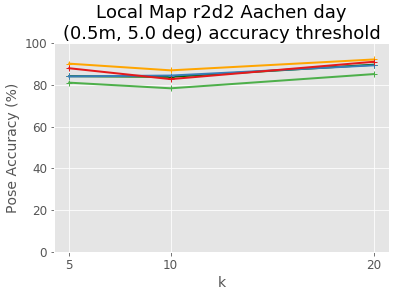}%
\includegraphics[width=0.3\textwidth]{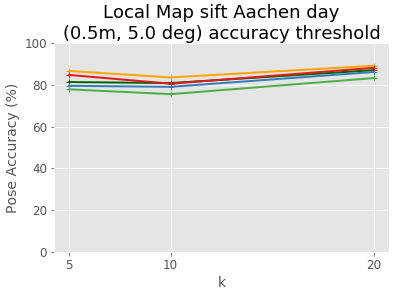}%

\includegraphics[width=0.3\textwidth]{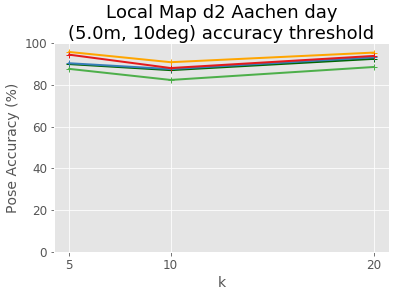}%
\includegraphics[width=0.3\textwidth]{plots/localsfm/aachen/AachenDayNightv11_day_low_local_r2d2_nolegend.png}%
\includegraphics[width=0.3\textwidth]{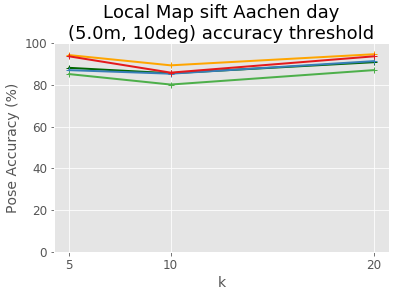}%

\includegraphics[width=0.3\textwidth]{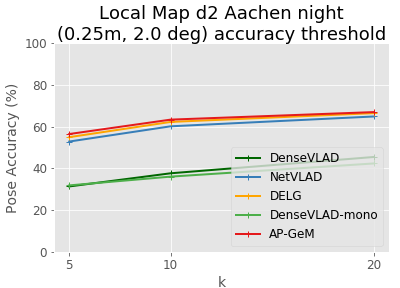}%
\includegraphics[width=0.3\textwidth]{plots/localsfm/aachen/AachenDayNightv11_night_high_local_r2d2_nolegend.png}%
\includegraphics[width=0.3\textwidth]{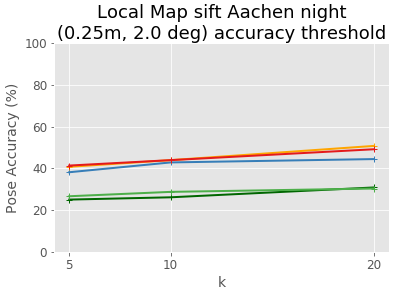}%

\includegraphics[width=0.3\textwidth]{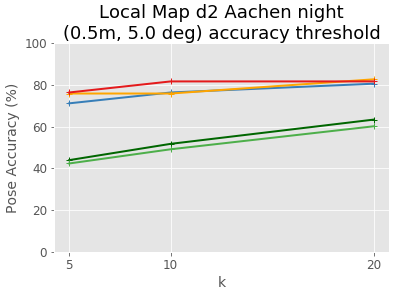}%
\includegraphics[width=0.3\textwidth]{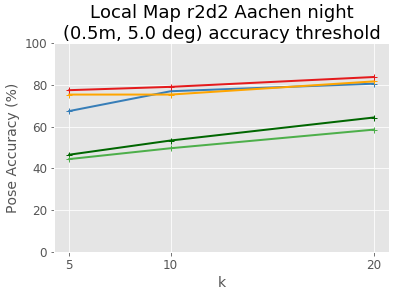}%
\includegraphics[width=0.3\textwidth]{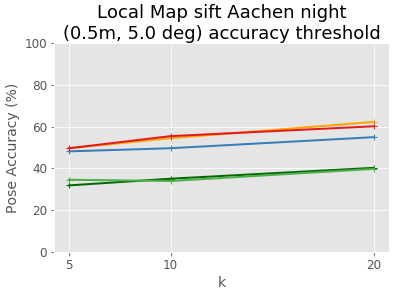}%

\includegraphics[width=0.3\textwidth]{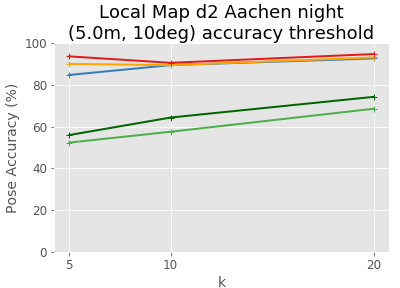}%
\includegraphics[width=0.3\textwidth]{plots/localsfm/aachen/AachenDayNightv11_night_low_local_r2d2_nolegend.png}%
\includegraphics[width=0.3\textwidth]{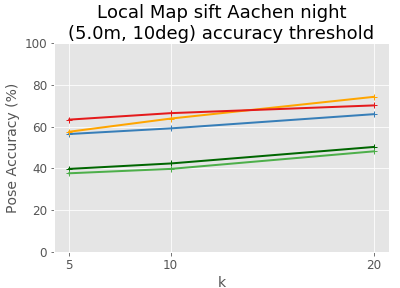}%

\end{center}
   \caption{\textbf{Task 2a (pose estimation without a global map) results for Aachen Day-Night}. We show results for D2-Net (left column), R2D2 (middle column), and SIFT (right column).}% Overall, we observe the same trend as for the Global-SFM case : the learned features overperform at night, while at day DenseVLAD-mono seems to be lagging behind in performance, and DELG slightly leading. The difference in performances between global descriptors is bigger for this task than for 2b.}
\label{fig:localsfm_supp_aachen}
\end{figure*}

\begin{figure*}[t]
\begin{center}
\includegraphics[width=0.3\textwidth]{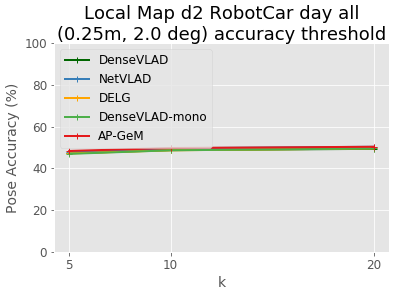}%
\includegraphics[width=0.3\textwidth]{plots/localsfm/robotcar/RobotcarSeasons_dayall_high_local_r2d2_nolegend.png}%
\includegraphics[width=0.3\textwidth]{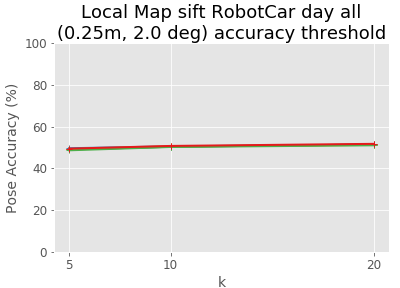}%

\includegraphics[width=0.3\textwidth]{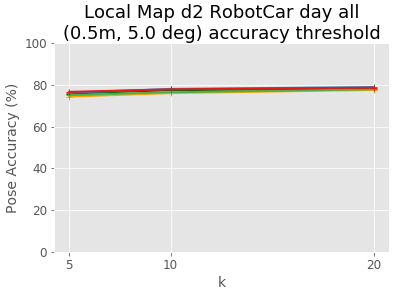}%
\includegraphics[width=0.3\textwidth]{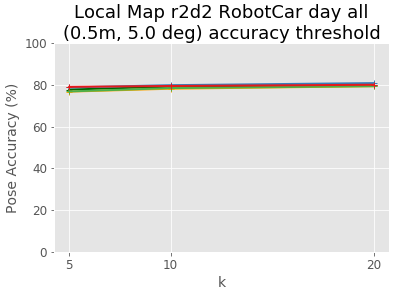}%
\includegraphics[width=0.3\textwidth]{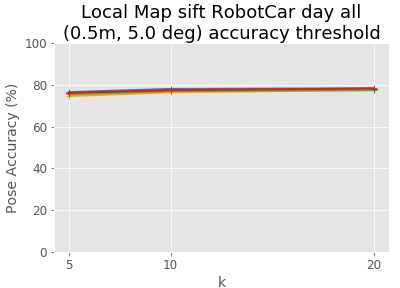}%

\includegraphics[width=0.3\textwidth]{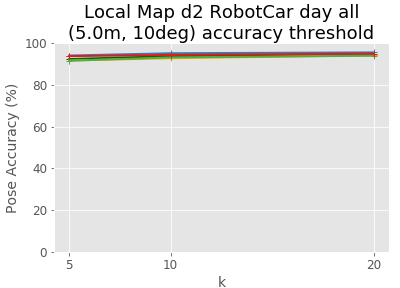}%
\includegraphics[width=0.3\textwidth]{plots/localsfm/robotcar/RobotcarSeasons_dayall_low_local_r2d2_nolegend.png}%
\includegraphics[width=0.3\textwidth]{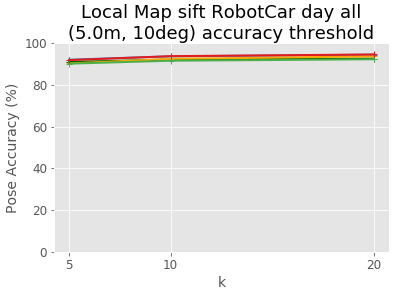}%

\includegraphics[width=0.3\textwidth]{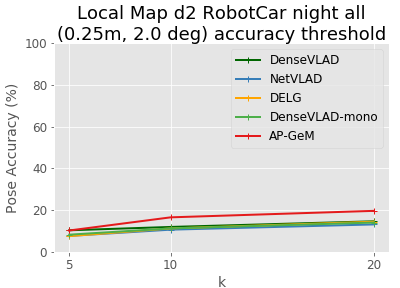}%
\includegraphics[width=0.3\textwidth]{plots/localsfm/robotcar/RobotcarSeasons_nightall_high_local_r2d2_nolegend.png}%
\includegraphics[width=0.3\textwidth]{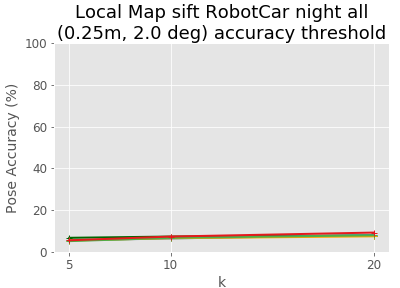}%

\includegraphics[width=0.3\textwidth]{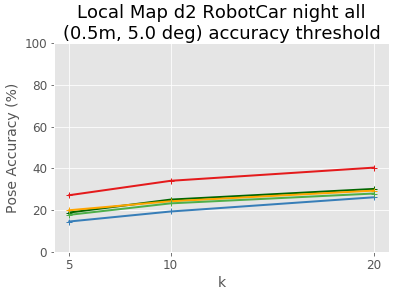}%
\includegraphics[width=0.3\textwidth]{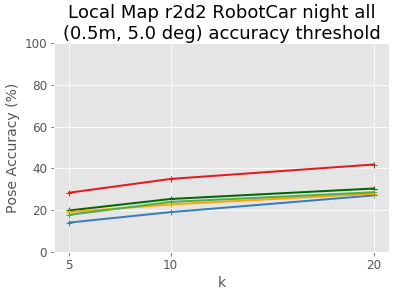}%
\includegraphics[width=0.3\textwidth]{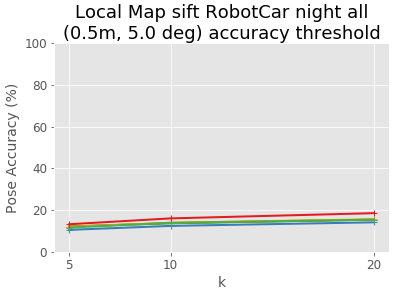}%

\includegraphics[width=0.3\textwidth]{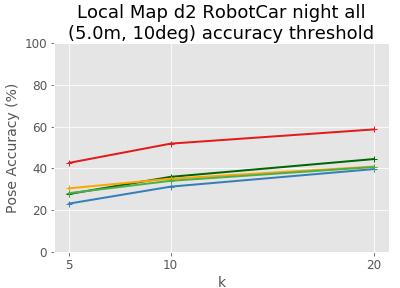}%
\includegraphics[width=0.3\textwidth]{plots/localsfm/robotcar/RobotcarSeasons_nightall_low_local_r2d2_nolegend.png}%
\includegraphics[width=0.3\textwidth]{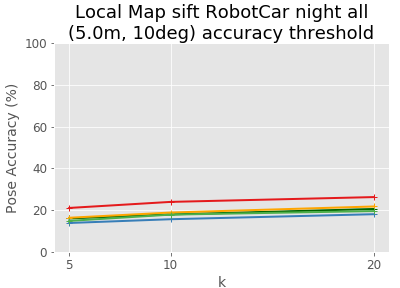}%

\end{center}
   \caption{\textbf{Task 2a (pose estimation without a global map) results for RobotCar}. We show results for D2-Net (left column), R2D2 (middle column), and SIFT (right column).}%\caption{\textbf{Task 2a (pose estimation with a local map) results for RobotCar-Seasons}. Compared to the results for Aachen Local and Global, the gap between the local and global results of RobotCar-Seasons did not increase, and the results are overall the same : AP-GeM performs better at night while at day, all the global features offer overall similar performances.}
\label{fig:localsfm_supp_robotcar}
\end{figure*}

\begin{figure*}[t]
\begin{center}
\includegraphics[width=0.3\textwidth]{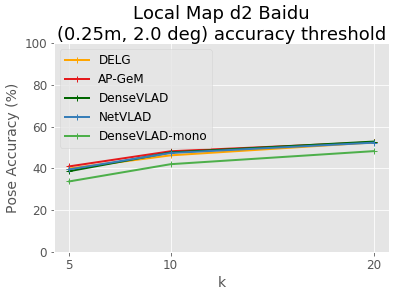}%
\includegraphics[width=0.3\textwidth]{plots/localsfm/baidu/Baidumall_025m20deg_local_r2d2_nolegend.png}%
\includegraphics[width=0.3\textwidth]{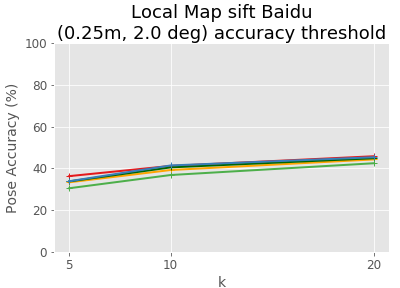}%

\includegraphics[width=0.3\textwidth]{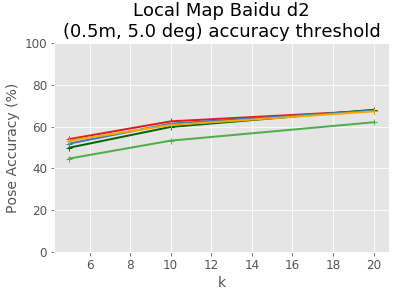}%
\includegraphics[width=0.3\textwidth]{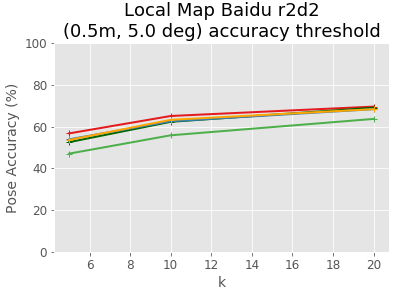}%
\includegraphics[width=0.3\textwidth]{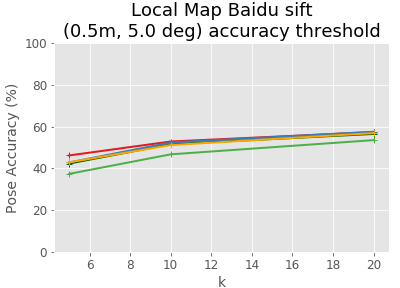}%

\includegraphics[width=0.3\textwidth]{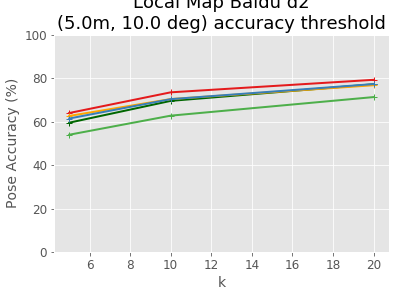}%
\includegraphics[width=0.3\textwidth]{plots/localsfm/baidu/Baidu_50m100deg_local_r2d2_nolegend.png}%
\includegraphics[width=0.3\textwidth]{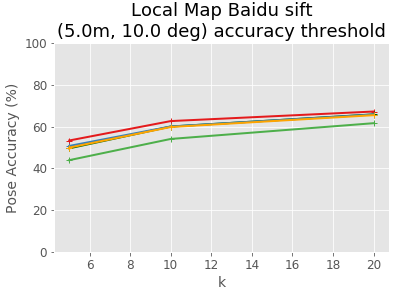}%
\end{center}
   \caption{\textbf{Task 2a (pose estimation without a global map) results for Baidu}. We show results for D2-Net (left column), R2D2 (middle column), and SIFT (right column).}%\caption{\textbf{Task 2a (pose estimation with a local map) results for Baidu-mall}. Again, the results are similar to the global SFM. AP-GeM is slightly ahead while DenseVLAD-monoscale is lagging behind.}
\label{fig:localsfm_supp_baidu}
\end{figure*}

\begin{figure*}[t]
\begin{center}
\includegraphics[width=0.3\textwidth]{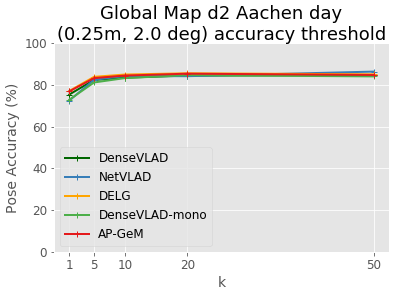}%
\includegraphics[width=0.3\textwidth]{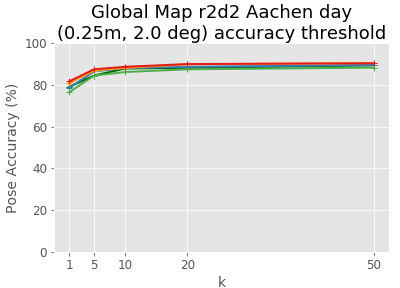}%
\includegraphics[width=0.3\textwidth]{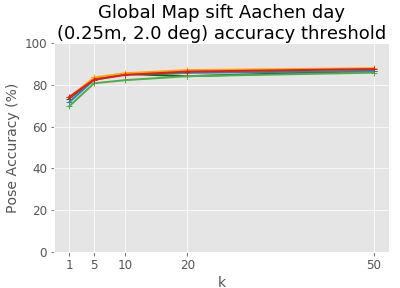}%

\includegraphics[width=0.3\textwidth]{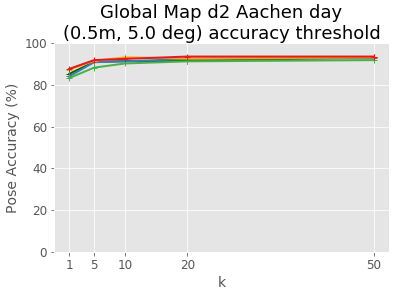}%
\includegraphics[width=0.3\textwidth]{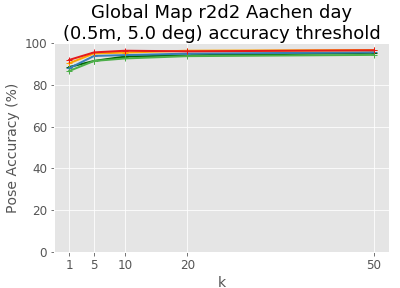}%
\includegraphics[width=0.3\textwidth]{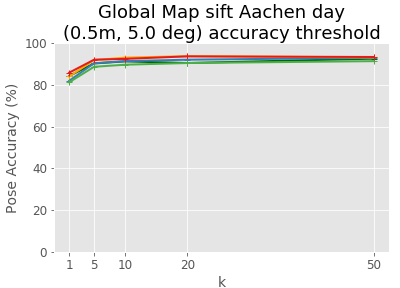}%

\includegraphics[width=0.3\textwidth]{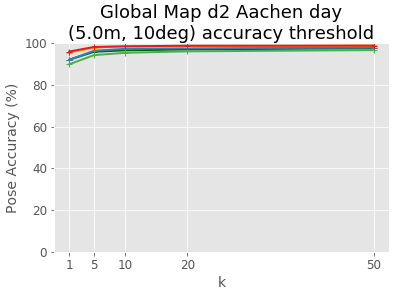}%
\includegraphics[width=0.3\textwidth]{plots/globalsfm/aachen/AachenDayNightv11_day_low_global_r2d2_nolegend.png}%
\includegraphics[width=0.3\textwidth]{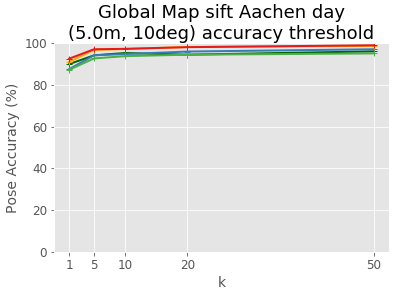}%

\includegraphics[width=0.3\textwidth]{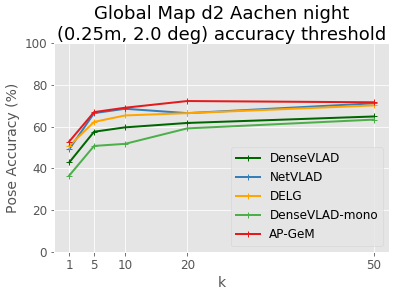}%
\includegraphics[width=0.3\textwidth]{plots/globalsfm/aachen/AachenDayNightv11_night_high_global_r2d2_nolegend.png}%
\includegraphics[width=0.3\textwidth]{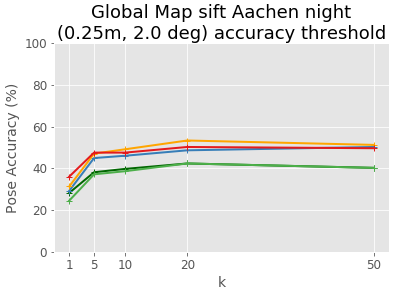}%

\includegraphics[width=0.3\textwidth]{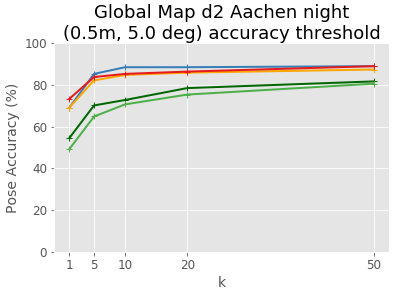}%
\includegraphics[width=0.3\textwidth]{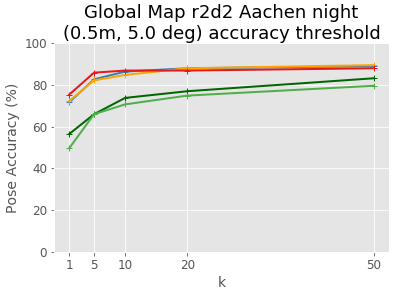}%
\includegraphics[width=0.3\textwidth]{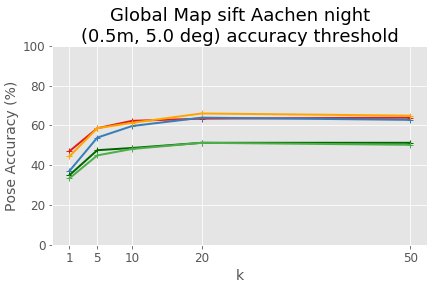}%

\includegraphics[width=0.3\textwidth]{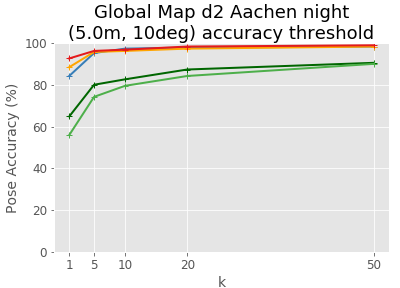}%
\includegraphics[width=0.3\textwidth]{plots/globalsfm/aachen/AachenDayNightv11_night_low_global_r2d2_nolegend.png}%
\includegraphics[width=0.3\textwidth]{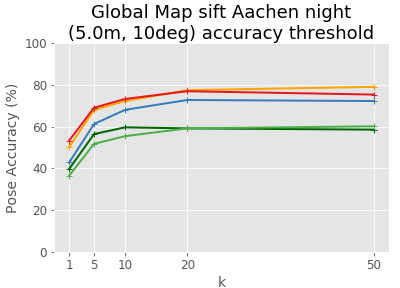}%

\end{center}
   \caption{\textbf{Task 2b (pose estimation with a global map) results for Aachen Day-Night}. We show results for D2-Net (left column), R2D2 (middle column), and SIFT (right column).}
\label{fig:globalsfm_supp_aachen}
\end{figure*}

\begin{figure*}[t]
\begin{center}
\includegraphics[width=0.3\textwidth]{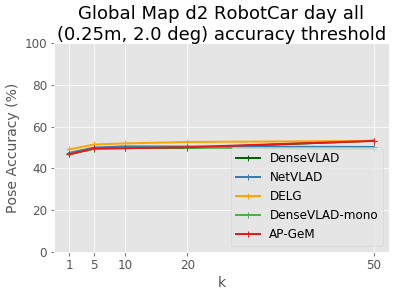}%
\includegraphics[width=0.3\textwidth]{plots/globalsfm/robotcar/RobotcarSeasons_dayall_high_global_r2d2_nolegend.png}%
\includegraphics[width=0.3\textwidth]{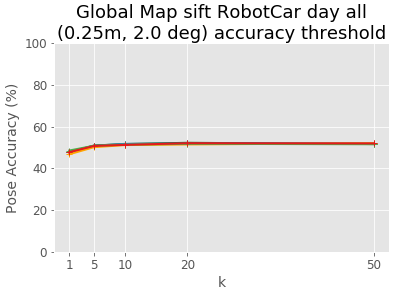}%

\includegraphics[width=0.3\textwidth]{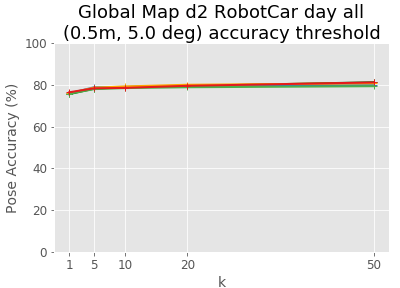}%
\includegraphics[width=0.3\textwidth]{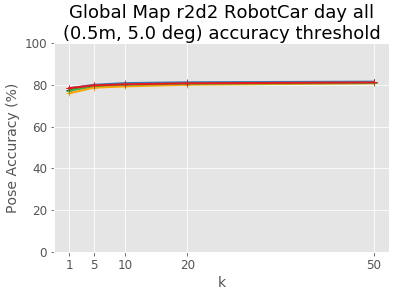}%
\includegraphics[width=0.3\textwidth]{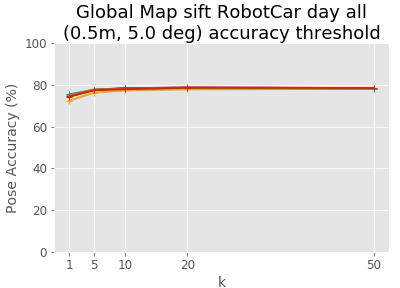}%

\includegraphics[width=0.3\textwidth]{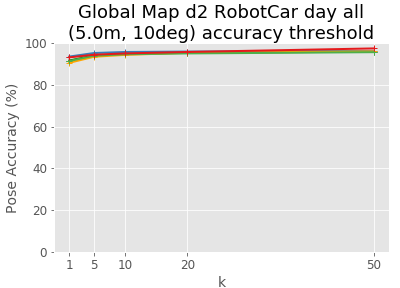}%
\includegraphics[width=0.3\textwidth]{plots/globalsfm/robotcar/RobotcarSeasons_dayall_low_global_r2d2_nolegend.png}%
\includegraphics[width=0.3\textwidth]{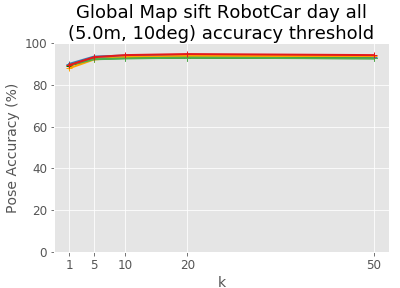}%

\includegraphics[width=0.3\textwidth]{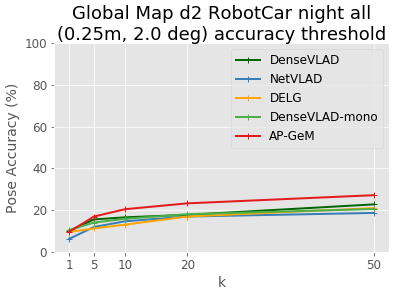}%
\includegraphics[width=0.3\textwidth]{plots/globalsfm/robotcar/RobotcarSeasons_nightall_high_global_r2d2_nolegend.png}%
\includegraphics[width=0.3\textwidth]{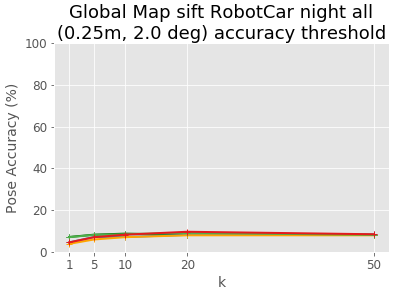}%

\includegraphics[width=0.3\textwidth]{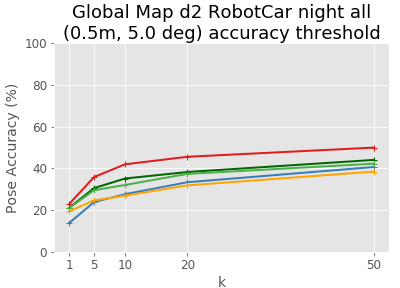}%
\includegraphics[width=0.3\textwidth]{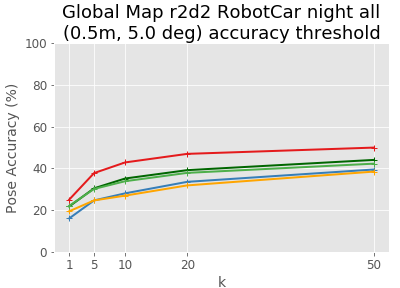}%
\includegraphics[width=0.3\textwidth]{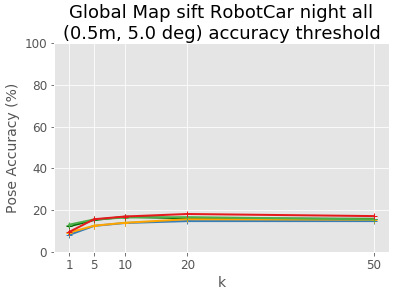}%

\includegraphics[width=0.3\textwidth]{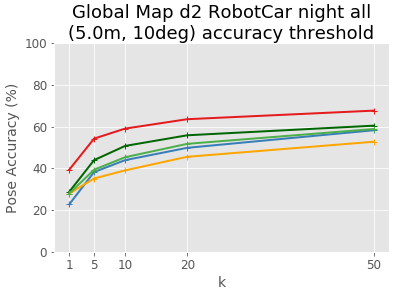}%
\includegraphics[width=0.3\textwidth]{plots/globalsfm/robotcar/RobotcarSeasons_nightall_low_global_r2d2_nolegend.png}%
\includegraphics[width=0.3\textwidth]{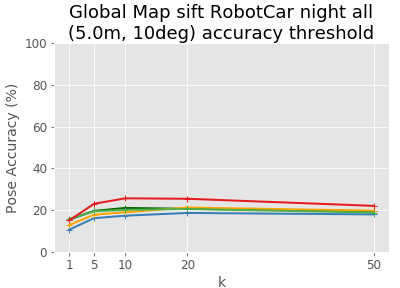}%

\end{center}
   \caption{\textbf{Task 2b (pose estimation with a global map) results for RobotCar}. We show results for D2-Net (left column), R2D2 (middle column), and SIFT (right column).}%On the day condition (first 3 rows), the performances of the global descriptors are about similar. At night, GeM-AP again performs better than other approaches. Interestingly, DELG offers poorer performances than the other learned descriptors and DenseVLAD. It also appears that SIFT is a bottleneck to comparing the performances of the retrieval methods at night. While there is a clear difference between different retrieval methods for the other local features, with SIFT those differences tend to vanish. The localization performances also decrease with more retrieved images used, which hints at the fact we may be adding local inconsistencies for SIFT, which is less discriminative, while the 2 learned approaches, R2D2 and D2, do not face such issue.}
\label{fig:globalsfm_supp_robotcar}
\end{figure*}

\begin{figure*}[t]
\begin{center}
\includegraphics[width=0.3\textwidth]{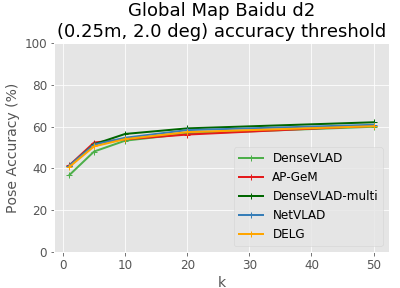}%
\includegraphics[width=0.3\textwidth]{plots/globalsfm/baidu/Baidu_025m20deg_global_r2d2_nolegend.png}%
\includegraphics[width=0.3\textwidth]{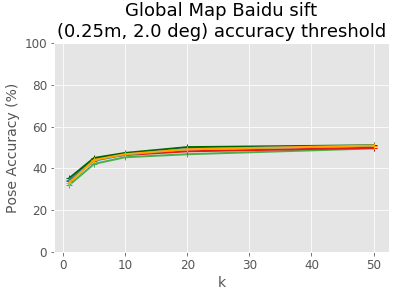}%
\\
\includegraphics[width=0.3\textwidth]{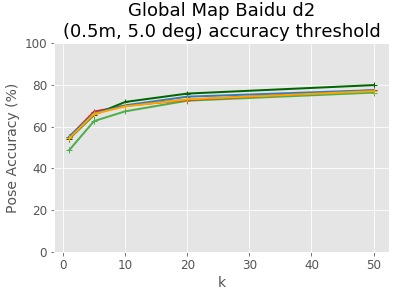}%
\includegraphics[width=0.3\textwidth]{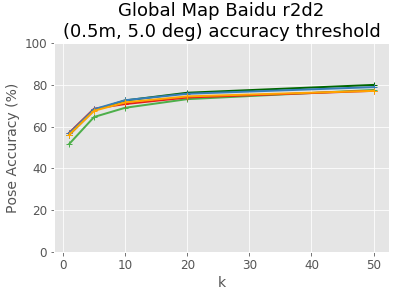}%
\includegraphics[width=0.3\textwidth]{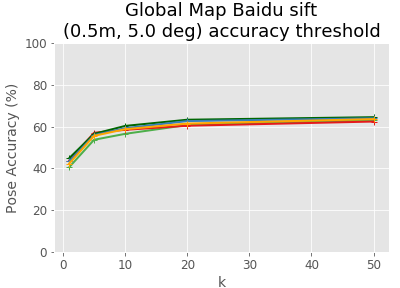}%
\\
\includegraphics[width=0.3\textwidth]{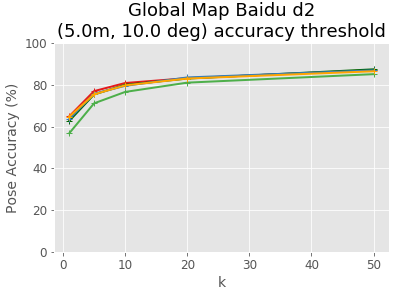}%
\includegraphics[width=0.3\textwidth]{plots/globalsfm/baidu/Baidu_50m100deg_global_r2d2_nolegend.png}%
\includegraphics[width=0.3\textwidth]{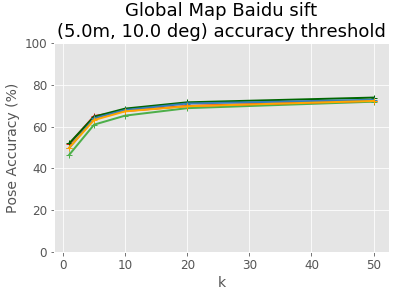}%
\end{center}
   \caption{\textbf{Task 2b (pose estimation with a global map) results for Baidu}. We show results for D2-Net (left column), R2D2 (middle column), and SIFT (right column).}%In Baidu, there does not appear to be a big gap. Overall DenseVLAD mono-scale lags behind. GeM-AP, DenseVLAD, NetVLAD and DELG are within 2\% pose accuracy to each others, with a gap with DenseVLAD-mono closing as we increase the amount of retrieved images.}
\label{fig:globalsfm_supp_baidu}
\end{figure*}

Figures~\ref{fig:localsfm_supp_aachen} to \ref{fig:localsfm_supp_baidu} show results for Task 2a (pose estimation without global map) and Figures~\ref{fig:globalsfm_supp_aachen} to \ref{fig:globalsfm_supp_baidu} show results for Task 2b (pose estimation with global map). %of the experiments and we discuss the results for the two tasks in more detail below.
%We can observe that while the performance using
%We observe no novel behavior for the medium threshold, which justifies our decision to only show results for the low and high accuracy thresholds in the main paper.

We first observe that the choice of local features does not change the ranking of image representations for any of the three datasets considered.
%The ranking of the retrieval representations is consistent between using R2D2 and the other two local representations (D2-Net and SIFT).
%As can be expected, increasing the number of retrieved images typically improves performance, but comes at the price of longer runtimes (not shown here).
Furthermore, as already observed in~\cite{RevaudNIPS19R2D2ReliableRepeatableDetectorsDescriptors,DusmanuCVPR19D2NetDeepLocalFeatures}, SIFT is outperformed by R2D2 and D2-Net.
As a consequence, we observe that the performance gap between different image representations shrinks when using SIFT: retrieving a relevant image does not help if the local feature type used is unable to establish good correspondences.
%Still, using SIFT features only reduces the gaps and does not lead to completely different behavior compared to using R2D2 or D2-Net features.
We can therefore confirm the statement of the main paper that our conclusions are not tied to a specific type of local features.

Similarly, while increasing the accuracy threshold decreases the number of localized images, the rank according to the global representations does not change.
Nevertheless, we observe that the gap between deep representations and DenseVLAD is smaller for higher accuracy, which suggest that the images retrieved with DenseVLAD, when relevant, are well suited for localization.

\subsection{Visual localization versus retrieval metrics}
\label{sec:posevsRetrieval}

This section analyses the correlation
between typical visual localization (measured via the percentage of the images retrieved at a given accuracy threshold) and retrieval metrics (measured by precision and recall).
We only consider the lowest accuracy threshold for the following results. On the one hand we observed that the ranking is similar for various localization accuracy thresholds. On the other hand, high pose accuracy is not necessary for both landmark retrieval and place recognition. To compute the retrieval metrics, we only use the visual overlap based ground truth relevance because, as we see in Section~\ref{sec:placeRec}, the two strategies yield similar results.
Finally, concerning local features we only show correlations with the results obtained with the R2D2 SFM maps for the accurate localization scenarios.

\begin{figure*}[t]
\begin{center}
\includegraphics[width=0.2\textwidth]{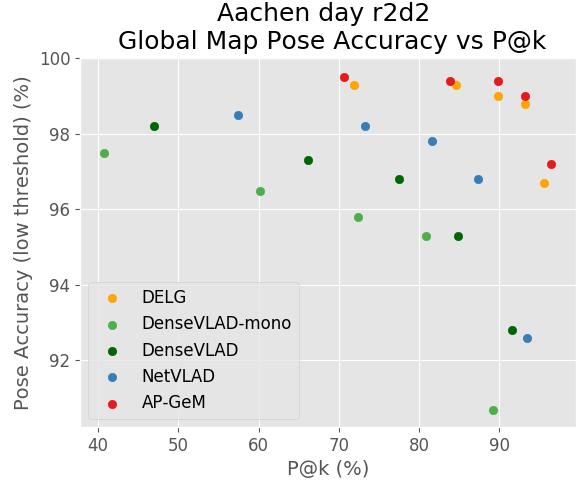}%
\includegraphics[width=0.2\textwidth]{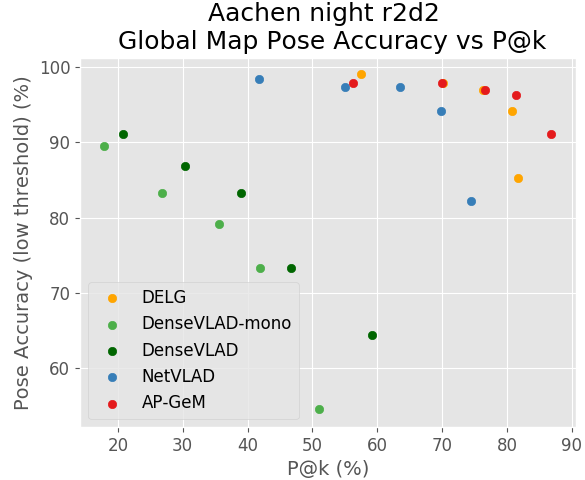}%
\includegraphics[width=0.2\textwidth]{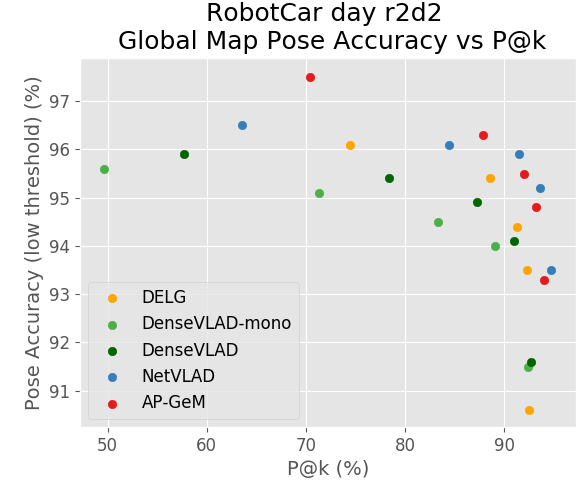}%
\includegraphics[width=0.2\textwidth]{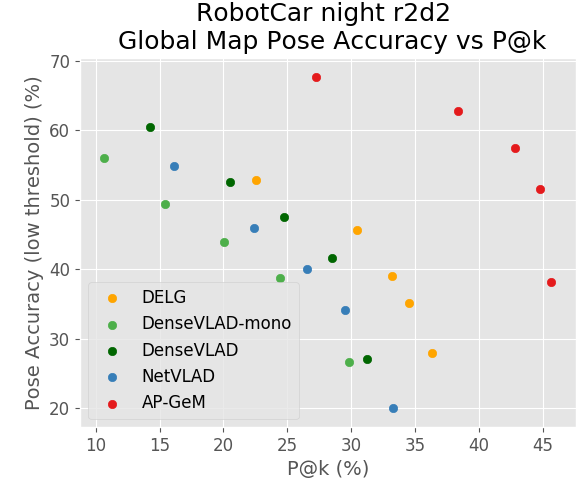}%
\includegraphics[width=0.2\textwidth]{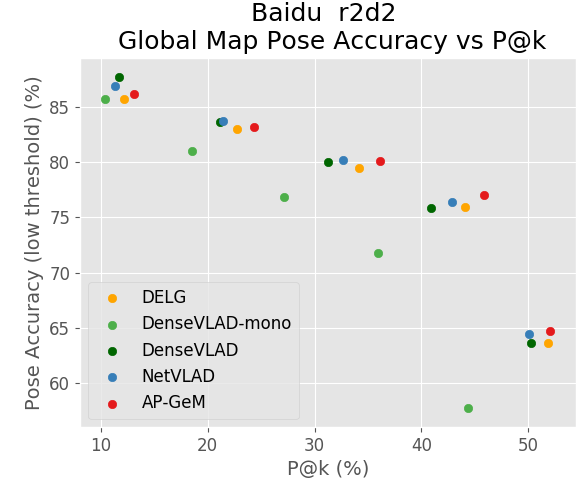}%
\\
\includegraphics[width=0.2\textwidth]{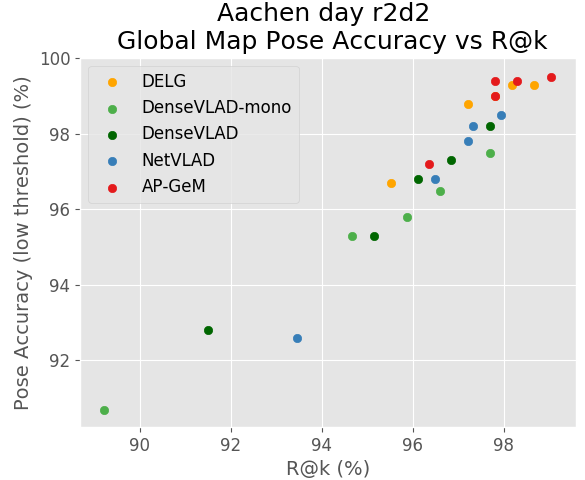}%
\includegraphics[width=0.2\textwidth]{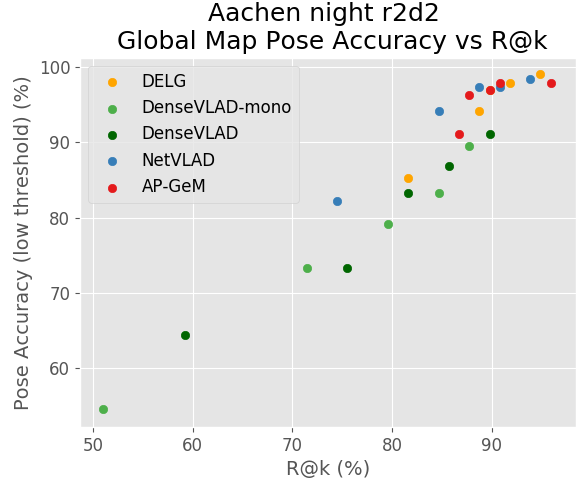}%
\includegraphics[width=0.2\textwidth]{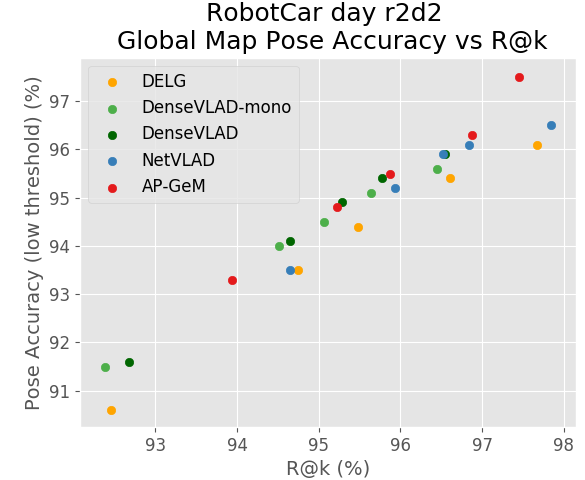}%
\includegraphics[width=0.2\textwidth]{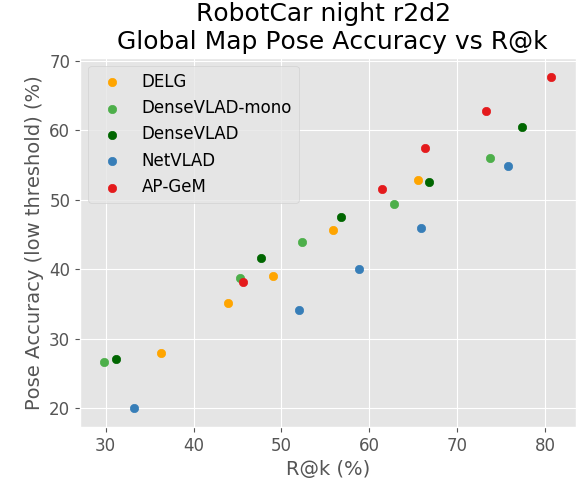}%
\includegraphics[width=0.2\textwidth]{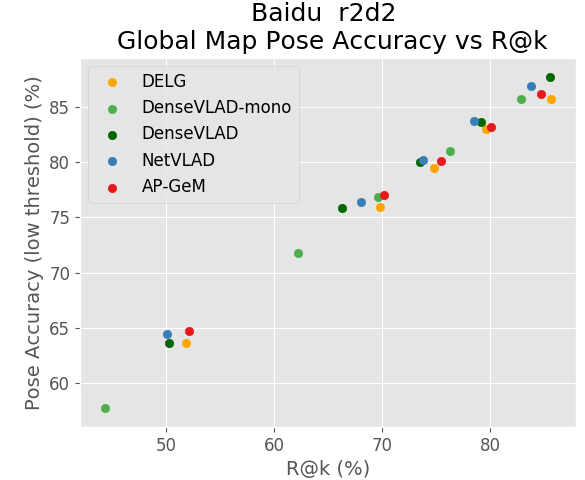}

\end{center}
   \caption{\textbf{Task 2b (pose estimation with a global map) vs retrieval metrics}. %For visibility, we add back the R@$k$ plots.
   We see that while there appears to be a clear correlation between R@$k$ and pose accuracy for this task, the Precision at top $k$ (P@$k$) does not seem to correlate with localization performance.}
   %We see that while there appears to be a clear correlation between the R@$k$ and pose accuracy for this task, the link is less clear for P@$k$..}
\label{fig:retrieval_vs_pose_global}
\end{figure*}

To analyse the correlation, we generate scatter plots where we select pairs of (Pose Accuracy, Retrieval Metric) for corresponding top $k$ retrieved images.
%all of the couples of (Pose Accuracy, Retrieval Metric) for all the amounts of retrieved images used in the pose accuracy experiments.
%
%
Figure~\ref{fig:retrieval_vs_pose_global} shows scatter plots for localization based on a global SFM map. As we see in the main paper, there appears to be a clear correlation between the R@$k$ and pose accuracy for this task. On the other hand, the Precision at top $k$ (P@$k$) does not seem to correlate with localization performance confirming that it is not necessary that all top retrieved images are relevant for a query as long as a few relevant images are found. % to have all top retrieved images relevant but it is sufficient to have a few good ones.

\begin{figure*}[t]
\begin{center}

\includegraphics[width=0.2\textwidth]{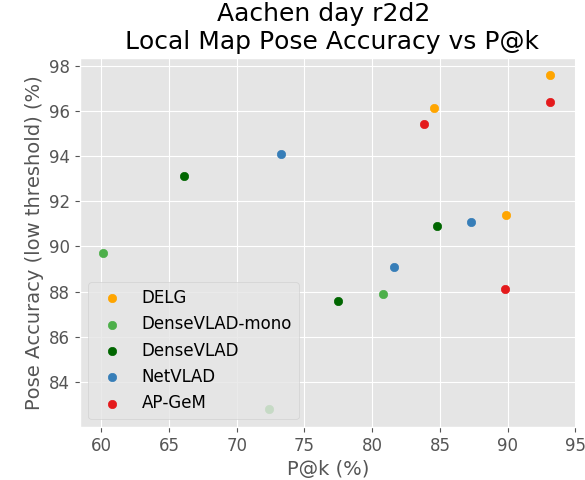}%
\includegraphics[width=0.2\textwidth]{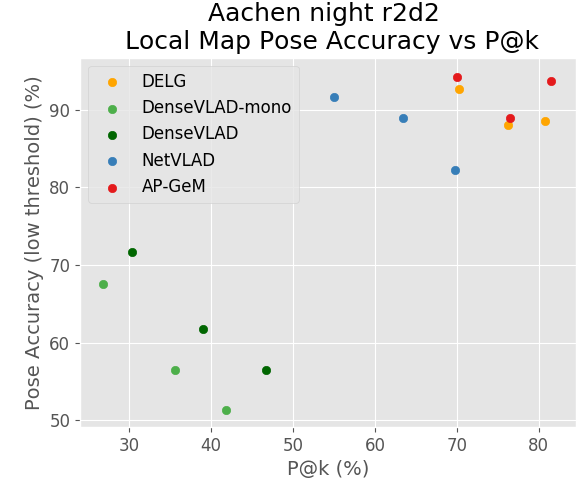}%
\includegraphics[width=0.2\textwidth]{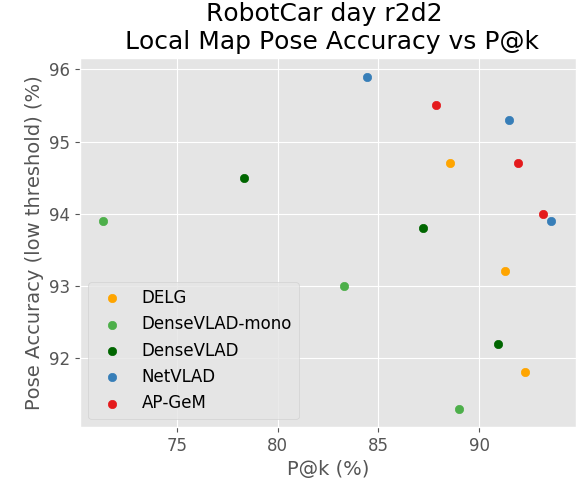}%
\includegraphics[width=0.2\textwidth]{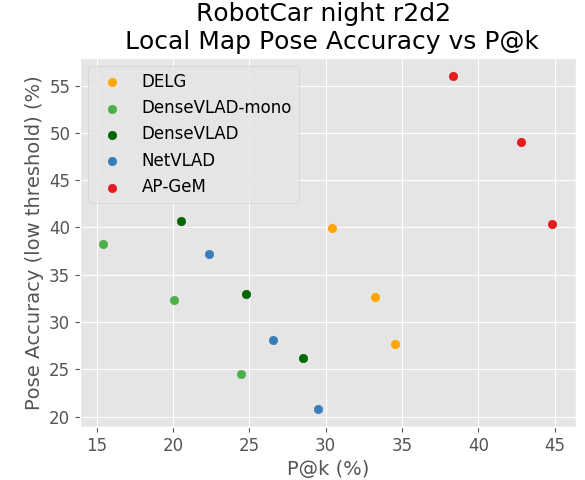}%
\includegraphics[width=0.2\textwidth]{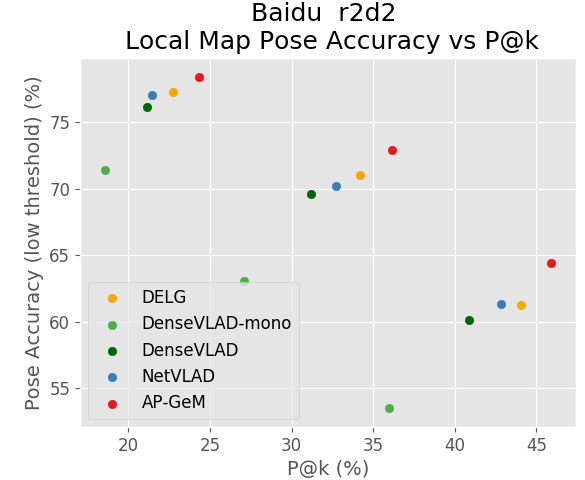}%
\\
\includegraphics[width=0.2\textwidth]{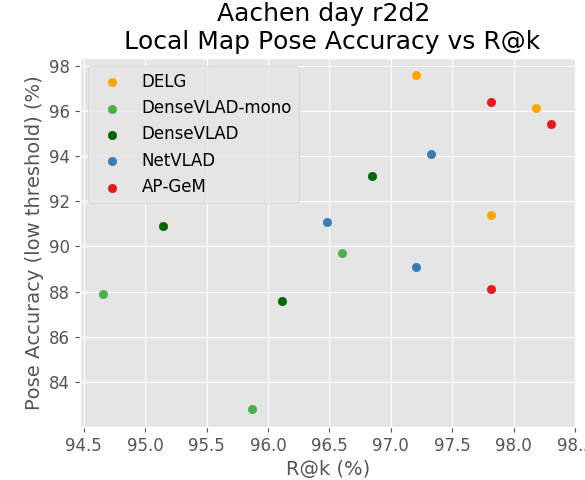}%
\includegraphics[width=0.2\textwidth]{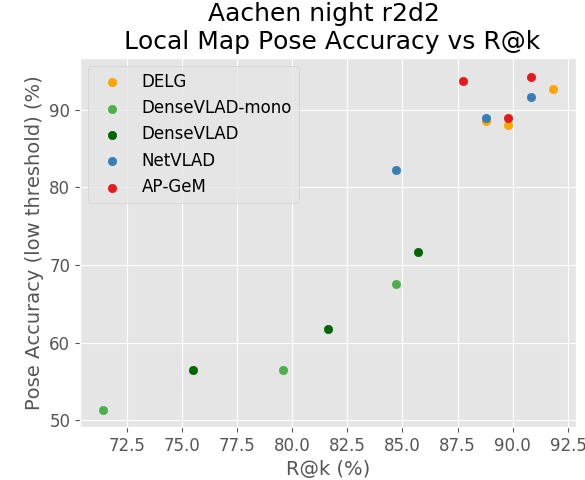}%
\includegraphics[width=0.2\textwidth]{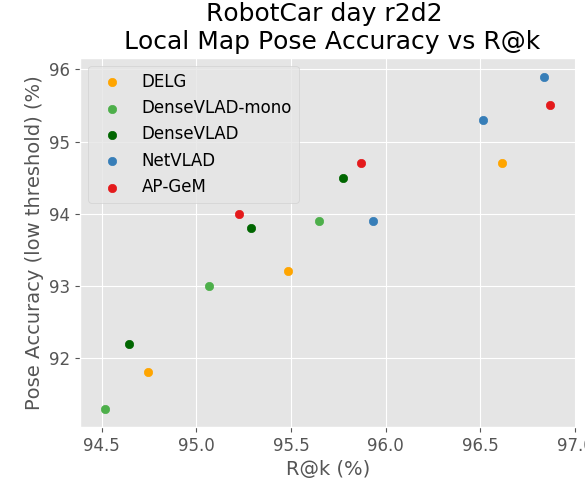}%
\includegraphics[width=0.2\textwidth]{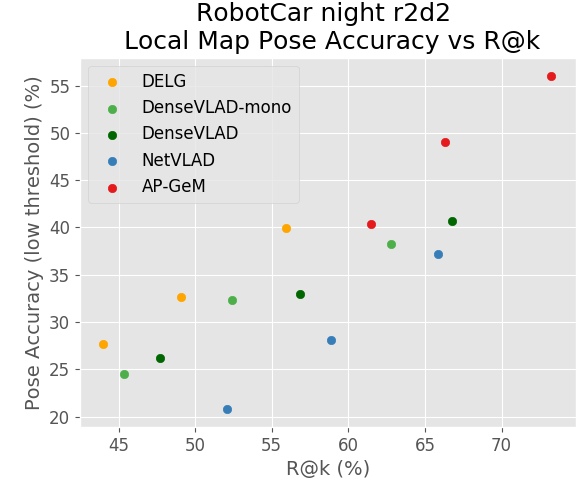}%
\includegraphics[width=0.2\textwidth]{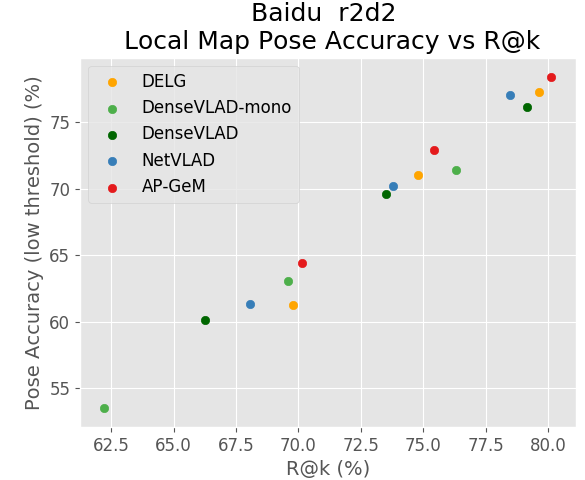}%
\end{center}
   \caption{\textbf{Task 2a (pose estimation with local map) vs retrieval metrics}. We can observe some weak correlation between pose estimation with local SFM and the Recall at $k$ (R@$k$), but there is no obvious correlation with the Precision at $k$ (P@$k$).}% which is somewhat surprising.}
  % While there appears to be a big spread in the point cloud on R@$k$ versus Pose Accuracy, R@$k$ seems somewhat indicative of the final pose accuracy results. Again, P@$k$ seems to correlate well with final pose accuracy results.}
\label{fig:retrieval_vs_pose_local}
\end{figure*}

Figure~\ref{fig:retrieval_vs_pose_local} shows scatter plots for localization based on the local SFM approach. We can observe a weaker correlation between this method and the Recall at $k$ (R@$k$) than when using a global map.
Again, we do not observe an obvious correlation with the Precision at $k$ (P@$k$). % which is somewhat unexpected as we could think that to build the local map it is important that all retrieved images to be relevant.
  % While there appears to be a big spread in the point cloud on R@$k$ versus Pose Accuracy, R@$k$ seems somewhat indicative of the final pose accuracy results. Again, P@$k$ seems to correlate well with final pose accuracy results.}

\begin{figure*}[t]
\begin{center}
\includegraphics[width=0.2\textwidth]{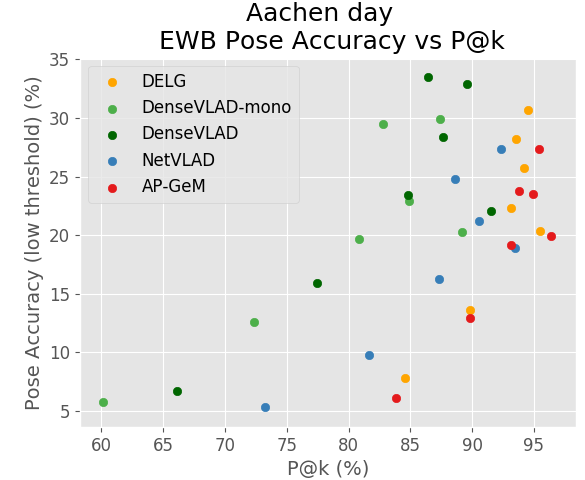}%
\includegraphics[width=0.2\textwidth]{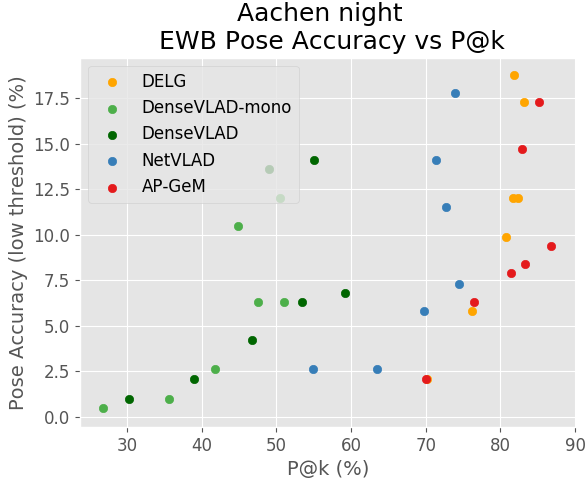}%
\includegraphics[width=0.2\textwidth]{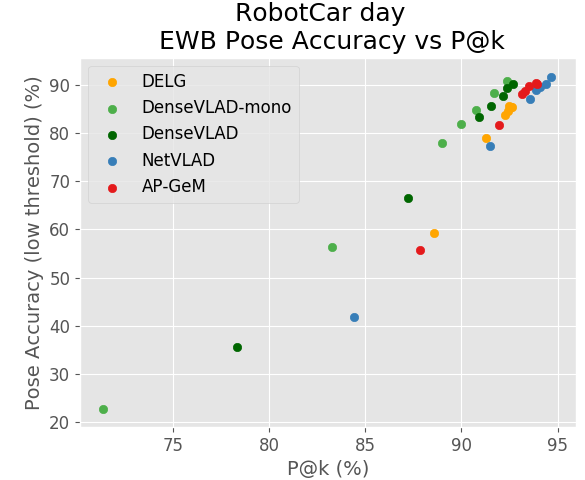}%
\includegraphics[width=0.2\textwidth]{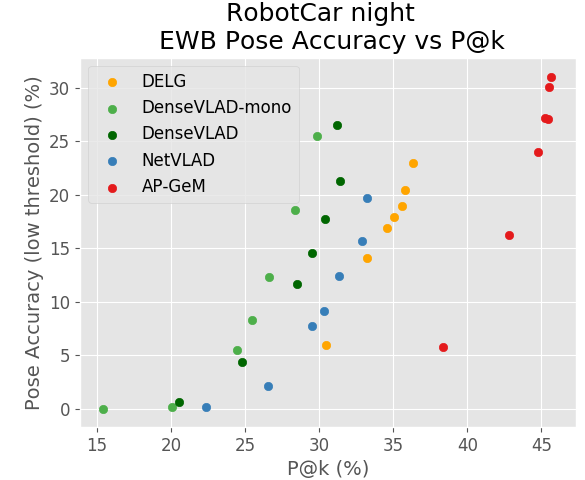}%
\includegraphics[width=0.2\textwidth]{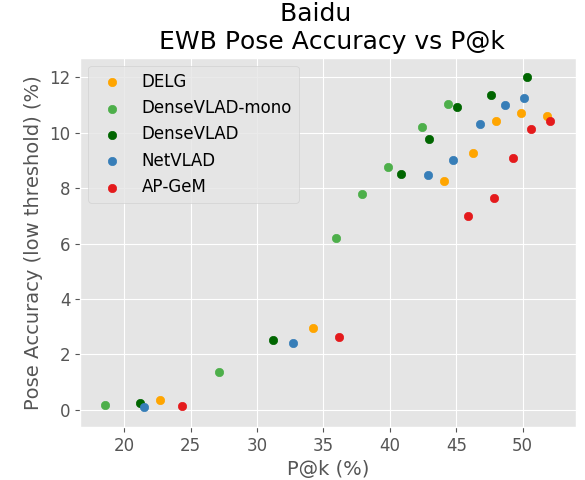}%
\\
\includegraphics[width=0.2\textwidth]{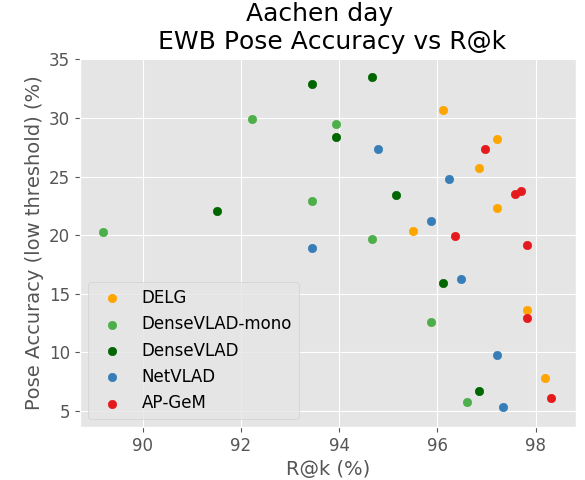}%
\includegraphics[width=0.2\textwidth]{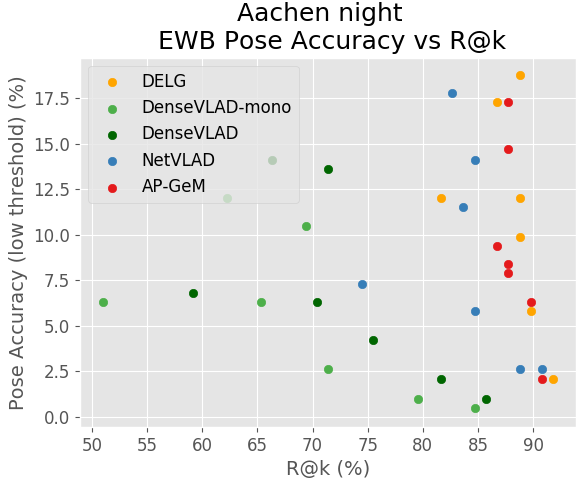}%
\includegraphics[width=0.2\textwidth]{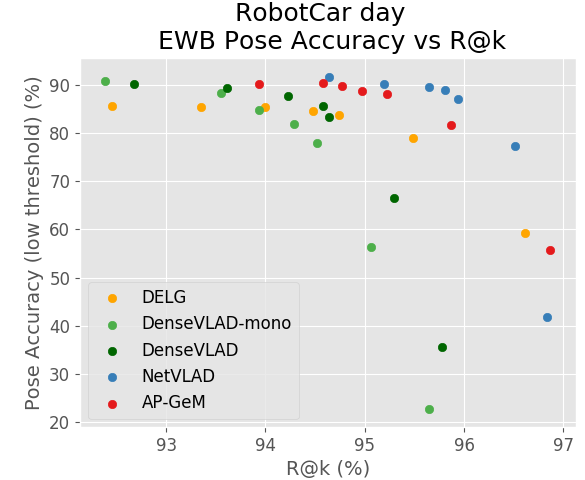}%
\includegraphics[width=0.2\textwidth]{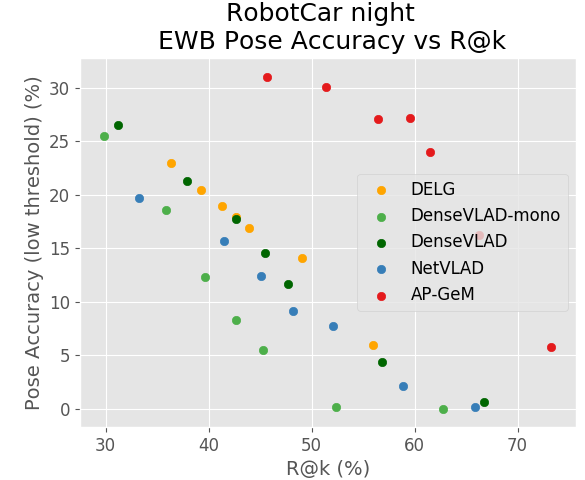}%
\includegraphics[width=0.2\textwidth]{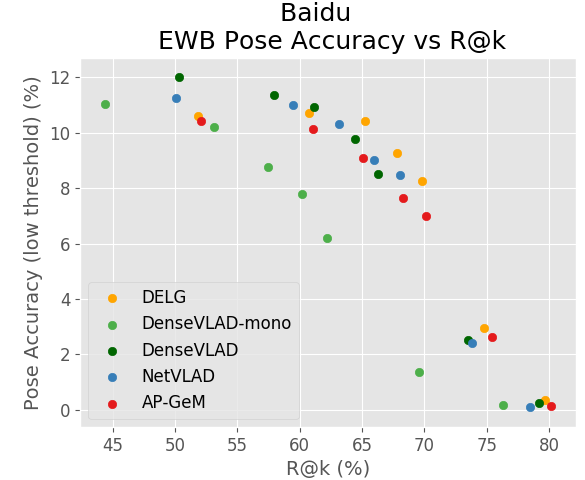}%
\end{center}
   \caption{\textbf{Task 1a (Approximate localization) vs retrieval metrics}.
  Here, correlation appears to be between pose estimation based on interpolation and P@$k$, but not with R@$k$.}
  %Pose Accuracy, while it is not so clear with the R@$k$. This is the opposite of the trends observed in \textbf{Task 2a/b}. Another interesting thing to notice is that the points associated to DenseVLAD in the P@$k$ pointclouds are consistently on the left of the other methods: for equal P@$k$, they achieve better pose accuracy, i.e. retrieve closer images to the query. This tends to confirm our hypothesis about DenseVLAD being less robust to viewpoint changes, and thus retrieving closer images.}
\label{fig:retrieval_vs_interp}
\end{figure*}

Finally, Figure~\ref{fig:retrieval_vs_interp} shows scatter plots for pose approximation via the EWB pose interpolation scheme. We observe a weak correlation between the pose estimation and the P@$k$ measures, but no correlation, or even in several cases an inverse correlation, with R@$k$.
%Pose Accuracy, while it is not so clear with the R@$k$.
This seems to be opposite to the trends observed for \textbf{Task 2a/b}.
An interesting thing to notice is that the points associated to DenseVLAD in the P@$k$ plots are consistently on the left of the other methods. For equal P@$k$, DenseVLAD achieves better pose accuracy, \ie, retrieves images closer to the query. This confirms that DenseVLAD is less robust to viewpoint changes and thus retrieves closer images.

% \gabriela{I did not integrated the comments/observations/discussions from the text below, so please check and see what we should add/integrate. }

%Similar to Fig.~\textbf{5} in the main paper, we report plots that show the correlation of Pose Accuracy at the lowest accuracy threshold with various methods against the retrieval metrics (P@$k$ and R@$k$). For \textbf{Task 1} we report the EWB method (see Fig.~\ref{fig:retrieval_vs_interp}), for \textbf{Task 2a} (see Fig.~\ref{fig:retrieval_vs_pose_local}) and \textbf{Task 2b} (see Fig.~\ref{fig:retrieval_vs_pose_global}) we only report D2-Net.
%Indeed, we saw that there was little difference between local features in the \textbf{Task 2a/b} experiments. We use the same definition of relevant images as in the main paper.

Overall, we notice some correlation between P@$k$ and pose accuracy for \textbf{Task 1}, and R@$k$ and pose accuracy for \textbf{Task 2a/b}. The correlation is clearest between R@$k$ and pose accuracy for \textbf{Task 2b}, where it appears in every dataset. These correlations are understandable: to have a good pose approximation, all of the retrieved images need to be close enough to the query. A single database image taken far away from the query, even with low weight in the interpolation scheme, can significantly affect the accuracy of the pose approximation.
For the geometry-based approaches, the correlation with R@$k$ can be explained by the fact that the localization pipeline does filtering based on the local feature matches. Thus, the system is less sensitive to wrong retrievals which will be filtered out if good local feature correspondences can be estimated. Still, at least one correctly retrieved image is necessary to facilitate pose estimation.

% Another interesting fact is that on \textbf{Task 1} (Fig.~\ref{fig:retrieval_vs_interp}), data points corresponding to DenseVLAD are consistently on the left of the scatter map. This represents better localization accuracy given similar P@$k$. This is consistent with our previous hypothesis that DenseVLAD is less robust to viewpoint changes, not having been trained for such scenario, and thus tends to retrieve geographically closer images.

%\towcolumn

{\small
\bibliographystyle{ieee}
% \bibliography{torsten}
\bibliography{bibliography,torsten}
}

\end{document}